%% file: ms.tex
\documentclass[journal]{IEEEtran}

\usepackage[pdftex]{graphicx}
\usepackage[caption=false,font=footnotesize]{subfig}
\usepackage{amsmath, amssymb, bm}
\usepackage{esvect}
\usepackage{array}
\usepackage{booktabs}
\usepackage{tabularx}
\usepackage[dvipsnames]{xcolor}
\usepackage{multirow}
\usepackage{algorithm}% http://ctan.org/pkg/algorithms
\usepackage{algpseudocode}% http://ctan.org/pkg/algorithmicx
\usepackage{url}
\usepackage{xparse}
\usepackage{dsfont}
\usepackage{standalone}
\usepackage{tcolorbox}
\usepackage[noadjust]{cite}

\usepackage{scalerel}
\usepackage{tikz}
\usetikzlibrary{svg.path}

\definecolor{orcidlogocol}{HTML}{A6CE39}
\tikzset{
    orcidlogo/.pic={
        \fill[orcidlogocol] svg{M256,128c0,70.7-57.3,128-128,128C57.3,256,0,198.7,0,128C0,57.3,57.3,0,128,0C198.7,0,256,57.3,256,128z};
        \fill[white] svg{M86.3,186.2H70.9V79.1h15.4v48.4V186.2z}
        svg{M108.9,79.1h41.6c39.6,0,57,28.3,57,53.6c0,27.5-21.5,53.6-56.8,53.6h-41.8V79.1z M124.3,172.4h24.5c34.9,0,42.9-26.5,42.9-39.7c0-21.5-13.7-39.7-43.7-39.7h-23.7V172.4z}
        svg{M88.7,56.8c0,5.5-4.5,10.1-10.1,10.1c-5.6,0-10.1-4.6-10.1-10.1c0-5.6,4.5-10.1,10.1-10.1C84.2,46.7,88.7,51.3,88.7,56.8z};
    }
}

\newcommand\orcidicon[1]{\href{https://orcid.org/#1}{\mbox{\scalerel*{
    \begin{tikzpicture}[yscale=-1,transform shape]
    \pic{orcidlogo};
    \end{tikzpicture}
}{|}}}}

\tcbset{width=\columnwidth,coltext=red,left=-6pt,fontupper=\footnotesize}

\usepackage[hidelinks]{hyperref}
\usepackage{zref-xr, zref-user}
\zexternaldocument*[supp-]{supp_material}

\input{macros.tex}
\begin{document}
% \linenumbers

\title{HAWKS: Evolving Challenging Benchmark Sets for Cluster Analysis}

% Authors
\author{
Cameron~Shand, Richard~Allmendinger,~\IEEEmembership{Member,~IEEE,} Julia~Handl, Andrew~Webb, and~John~Keane
\IEEEcompsocitemizethanks{\IEEEcompsocthanksitem C. Shand is with the University College London, London WC1E 6BT, U.K. (e-mail: c.shand@ucl.ac.uk).
\IEEEcompsocthanksitem R. Allmendinger, J. Handl and J. Keane are with the University of Manchester, Manchester M15 6PB, U.K. (e-mail: richard.allmendinger@manchester.ac.uk, julia.handl@manchester.ac.uk, john.keane@manchester.ac.uk).
\IEEEcompsocthanksitem A. Webb is with vTime, Liverpool L8 5RN, U.K. (email: andrew@awebb.info).}% <-this % stops an unwanted space
\thanks{Manuscript received February 13, 2021; revised May 17, 2021 and September 2, 2021; accepted December 7 2021.
}}

\ifCLASSOPTIONpeerreview
\markboth{IEEE Transactions on Evolutionary Computation}%
{Anonymized}
\else
\markboth{IEEE Transactions on Evolutionary Computation}%
{Shand \MakeLowercase{\textit{et al.}}: HAWKS: Evolving Challenging Benchmark Sets for Cluster Analysis}

% \IEEEpubid{}

\maketitle

% ---------------------------------------------- %
% ----------------- Abstract ------------------- %
% ---------------------------------------------- %
\begin{abstract}
Comprehensive benchmarking of clustering algorithms is rendered difficult by two key factors: (i)~the elusiveness of a unique mathematical definition of this unsupervised learning approach and (ii)~dependencies between the generating models or clustering criteria adopted by some clustering algorithms and indices for internal cluster validation. Consequently, there is no consensus regarding the best practice for rigorous benchmarking, and whether this is possible at all outside the context of a given application. Here, we argue that synthetic datasets must continue to play an important role in the evaluation of clustering algorithms, but that this necessitates constructing benchmarks that appropriately cover the diverse set of properties that impact clustering algorithm performance. Through our framework, HAWKS, we demonstrate the important role evolutionary algorithms play to support flexible generation of such benchmarks, allowing simple modification and extension. We illustrate two possible uses of our framework: (i)~the evolution of benchmark data consistent with a set of hand-derived properties and (ii)~the generation of datasets that tease out performance differences between a given pair of algorithms. Our work has implications for the design of clustering benchmarks that sufficiently challenge a broad range of algorithms, and for furthering insight into the strengths and weaknesses of specific approaches.
\end{abstract}


\IEEEpeerreviewmaketitle

% ---------------------------------------------- %
% ---------------- Introduction ---------------- %
% ---------------------------------------------- %
\section{Introduction}

\IEEEPARstart{C}{luster} analysis is an unsupervised learning approach with the high-level aim of identifying groups (clusters) of objects that are more similar to each other than to the objects in other groups. It is a fundamental approach for knowledge discovery, with a broad range of applications including bioinformatics~\cite{handl2005computational,maulik2011clustapps,xu2010clustering}, cybersecurity~\cite{ahmed2016anomaly}, medicine~\cite{ma2010review}, market segmentation~\cite{dolnicar2002market}, and social network analysis~\cite{handcock2007social}.

% The problem with clustering, and utility of a ground-truth
Due to the unsupervised nature of clustering, the process of evaluating the quality of a partition (i.e.\ a given set of clusters) is not straightforward. Attempts to formally capture the qualities intuitively associated with pronounced cluster structure (such as compactness of individual clusters and separation between clusters) have led to the mathematical definition of a range of \emph{internal} validation indices, which can be both complementary and conflicting~\cite{hennig2015true,jain1999,milligan1996,von2012clustering}. Arbelaitz et al.~\cite{arbelaitz2013validity} studied 30 such indices and concluded that the utility of such measures varied depending on the datasets considered, highlighting the limited scope of each individual index.  

\emph{External} cluster validation indices are thought to address this limitation and to provide a more objective assessment of clustering performance~\cite{handl2005computational}. However, they require knowledge of the ground truth for a given dataset, i.e.\ information about the correct cluster membership of each data point --- information that is difficult to come by in realistic unsupervised learning applications. For this reason, synthetic benchmark datasets (i.e.\ datasets with a known generating model) play an important role in the evaluation of clustering performance. A key advantage of such data is that both the ground truth and any assumptions implicit to the generating process are accessible. This allows for both an objective assessment of clustering performance and informed reasoning about the key drivers behind the observed performance.

In principle, direct control over the generating model then allows for the provision of datasets with specific and varied properties. This facilitates the testing of performance with regards to these known characteristics; the benefit is concrete: practically translatable insight regarding the strengths and weaknesses of particular algorithms~\cite{hooker1995testing}. However, existing generators for synthetic clustering benchmarks have not been designed with this level of flexibility in mind --- instead, they typically use a set of fixed (manually tuned) parameter bounds within their generating model~\cite{handl2005generator}, limiting the complexity and diversity of datasets that can be obtained and failing to fully match the range of challenges posed by real-world datasets.

Our framework, HAWKS, is designed to address this limitation through the integration of an evolutionary algorithm (EA) into the generating process. EAs lend themselves as a mechanism to directly control and adapt key aspects of a partition's generating model --- here, we demonstrate that this facilitates the design of more powerful benchmarks, exhibiting a diverse range of properties. Furthermore, the innate modularity of EAs provides flexibility in choosing all key model components, including the representation of individual clusters and the set of objectives and constraints constituting the partition-level generating model, such as constraints on inter-cluster relationships. This flexibility is key to a broader utility of the framework, and our experiments illustrate HAWKS' potential in evolving benchmark sets either to meet a predefined set of criteria, or to directly maximize performance differences between pairs of algorithms.

% Contributions
In summary, the main contributions of this paper are as follows:
\begin{enumerate}
    \item We propose an evolutionary framework for the generation of clustering benchmarks, HAWKS, that allows for flexible parameterization of its generating model. 
	\item We describe a set of measurable properties (problem features) that quantify the difficulty of a cluster structure from a range of perspectives. This set of problem features is utilized to define an instance space~\cite{smith2012instancespace}, enabling visual examination of the correlations between these properties and algorithm performance. For this purpose, we represent each benchmark set by its associated problem features, embed this representation into two dimensions, and use colour coding to highlight the top performing algorithm for each dataset. 
	\item We provide an indicative example of the use of HAWKS to generate datasets across the instance space by varying a subset of parameters. In this first optimization mode, individual problem features can be deployed as objectives and/or constraints. By varying the relative importance of each feature, a diverse collection of benchmarks can be obtained.
	\item We present a second optimization mode for HAWKS, which aims to generate benchmarks that elicit performance differences between pairs of clustering algorithms. Analysis of the problem features and cluster structures associated with the resulting datasets allows for the identification of the relative strengths and weaknesses of each algorithm.

\end{enumerate}

% Outline
The remainder of this paper proceeds as follows. Section~\ref{sec:bg} further motivates the need for synthetic cluster generators, and positions our work relative to existing generators and the literature. Section~\ref{sec:hawks} describes HAWKS, discussing the importance of each component in the framework as a whole, and the design choices we have made, to illustrate its use. Section~\ref{sec:exp_setup} describes our experimental setup, including the set of problem features we deploy to measure the complexity of a dataset, and details of existing benchmark sets we compare against. The experimental results are presented in Section~\ref{sec:exp_results}, demonstrating HAWKS' ability to evolve diverse benchmark data, and instances tailored to challenge specific algorithms. Finally, we conclude and discuss future research directions in Section~\ref{sec:conc}.

% ---------------------------------------------- %
% ----------------- Background ----------------- %
% ---------------------------------------------- %
\section{Background} \label{sec:bg}

This section reviews the relevant background to our work. We start by positioning our work relative to similar problems in the literature, most prominently the relevance of benchmarks, and related work on the algorithm selection problem. We then review the issues of cluster analysis, and discuss the implications for the development and use of synthetic benchmarks. Finally, we provide an overview of existing benchmark generators for clustering, and analyze their strengths and limitations.

\subsection{The general role of benchmarks} \label{ssec:bg_need}

Empirical comparison between techniques is a cornerstone of the scientific method. At a community-level, methods developed by independent researchers need to be compared in order to gain insights into the applicability, generalizability, and efficacy of their developments. As direct comparison is only possible on the same problem instances, subsequent research is highly likely to adopt instances used by other researchers. The importance of reproducibility further re-enforces the need for a common, accessible set of data that can be utilized to facilitate comparisons across independent studies. This feedback loop results in ``standard'' benchmarks becoming virtually required to include in experimentation~\cite{hooker1995testing}. This requirement is supported explicitly through the creation of benchmark suites --- a collection of problems collated and/or created for the purpose of widespread comparison \cite{macia2014towards,mersmann2010benchmarking}.

The issue with this feedback loop is that the community as a whole risks tuning both hyperparameters and algorithmic development to these specific problems~\cite{hooker1995testing, schmidhuber2015deep}. If these popular problems represent a broad-spectrum of real-world challenges, then this is not a negative; analyzing whether these problems adequately cover the space of encounterable problems, however, is difficult if even possible to do in its entirety \cite{macia2014towards}. 

To combat this challenge, Hooker~\cite{hooker1995testing} argues for ``controlled experimentation'' i.e.\ comparing algorithmic performance specifically on a problem characteristic that the research in question is addressing, compared to the ``competitive testing'' that is encouraged when the same subset of datasets are re-used time and again. This argument is consistent with the implications of the No-Free-Lunch (NFL) theorem for learning, which supports the intuition that no single algorithm is expected to be superior across all problems~\cite{wolpert1996lack}, and that binary statements about algorithm superiority (i.e.\ ``algorithm $A$ is better than algorithm $B$'') are only possible within particular problem classes. Controlled experimentation is inherently simpler with synthetic benchmarks, as we have control over the generating mechanism and thus (to varying extents) the properties of the instances. The use of real-world instances for this purpose is possible only if there is an appropriate measure of the problem characteristic and a controlled way to vary it.

\subsection{The algorithm selection problem} \label{ssec:bg_asp}
In the above, we have introduced the notion of problem characteristics, and the need for these to be varied in benchmark data. This is to ensure appropriate coverage across the space of possible instances, and the ability to appropriately differentiate between the challenges these instances may pose for different algorithms. 

Rice~\cite{rice1976algorithm} formalized these characteristics as ``problem features'' in the context of the algorithm selection problem (ASP), where the goal is to predict which algorithm from a portfolio is best-suited to a given problem instance based on its problem features. This is premised on the existence of an identifiable relationship between the problem features and problem difficulty for a given algorithm, typically requiring these features to be specific to the problem class. A series of papers by Smith-Miles further extended this framework~\cite{smith2009cross,smith2012instancespace,smith2014towards}, using an instance space to visualize the interaction between problem features and algorithmic performance~\cite{smith2012instancespace} and applying this approach to combinatorial optimization~\cite{smith2012measuring,smith2015generating} and supervised learning~\cite{munoz2018instance}.

The problem of selecting an appropriate algorithm for a given task is closely related to the ``algorithm configuration problem'' (i.e.\ hyperparameter optimization), which is important in both the metaheuristic and machine learning communities~\cite{birattari2009tuning,automl_book,lopez2016irace}. In this context, the identification of problem features has featured prominently in the form of exploratory landscape analysis, where quantification of different aspects of the fitness landscape informs not only algorithm configuration, but a more general understanding of the suitability of algorithm components to different problem characteristics~\cite{kerschke2017exploratory,kotthoff2015improving,mersmann2011exploratory,mersmann2010benchmarking}.

\subsection{Relevance to unsupervised learning} \label{ssec:bg_clustering}

Unsupervised learning aims to identify related groups in a dataset in the absence of information about the groups themselves. This pattern recognition task is subjective, even for humans~\cite{bowker2000sorting,everitt2011cluster}, explaining the existence of a diverse range of clustering algorithms that make a broad variety of mathematical assumptions about desirable cluster properties~\cite{von2012clustering,hennig2015true}.

The dilemma stemming from the NFL theorem is thus exacerbated in cluster analysis, with differing clustering algorithms useful for different clustering problems~\cite{ben2018clustering,hennig2015true}, due to unavoidable differences in the underlying formulation~\cite{kleinberg2003impossibility}. As the inductive bias of each clustering algorithm differs, this fundamentally governs its capabilities: for example, K-Means assumes hyper-spherical, compact clusters, and this strictly limits its ability to deal with data that violates this assumption.

In consequence, having a representative and diverse collection of benchmark datasets is particularly crucial in cluster analysis, in order to fairly test each algorithm's strengths and weaknesses, and to avoid any biases towards those methods whose inductive bias may be consistent or correlated with the assumptions of a limited benchmark set. However, compared to other optimization problems, evaluation of this is complicated by the multifarious nature of cluster quality measures itself~\cite{von2012clustering,hennig2015true,hennig2018thoughts}. Without explicit effort to first identify and quantify aspects of problem structure or difficulty that matter in cluster analysis, it is challenging to assess and ensure appropriate diversity of problem instances and, thus, a comprehensive benchmark suite. Any efforts to improve existing benchmarks must therefore involve such identification and quantification as an integral step. For a further discussion on the considerations and complexities of benchmark studies in clustering, see \cite{milligan1996,vanmechelen2018benchmarking}.

Previous works on directly tackling the ASP for clustering (in the context of ``meta-learning'') have used generic statistical properties of the data (such as the mean of the moments across the data). An alternative is the use of measures directly characterizing a given group structure~\cite{ferrari2015clustering,soares2009analysis}. The latter approach is likely to be more powerful in identifying drivers of algorithm performance, but its limitation lies in the difficulty of transferring insights on feature-performance correlations to data with an unknown group structure.

\subsection{Existing clustering benchmarks and generators} \label{ssec:bg_generators}

In the following, we consider existing clustering benchmarks and the extent to which these address the representativeness and diversity issues discussed above. This is non-exhaustive, as many prior works generate synthetic data as a part of the work, rather than the focus of it.

There are various works that have conducted similar benchmarking studies, to better compare and contrast clustering algorithms under different conditions. In early work of this kind, six type of error perturbation were applied multivariate Gaussian distributions to assess the ability of fifteen clustering algorithms to handle noise \cite{milligan1980}. Another example used Gaussian clusters (scaling the sampling interval to adjust feature-wise overlap) to assess algorithmic robustness to simple measures (cluster size, dimensionality) and sensitivity to hyperparameters \cite{rodriguez2019}.

The datasets from the UCI Machine Learning Repository~\cite{Dua2017} comprise a diverse mix of real-world datasets and are one of the most common benchmarks for the evaluation of machine learning algorithms. Recent analysis demonstrates, however, that even with real-world data from a range of application areas, diversity in complexity is not guaranteed. Macia et al.~\cite{macia2014towards} analyzed the datasets of the UCI Machine Learning Repository, finding a surprising similarity in complexity across them. They discovered a lack of diversity in both the statistical properties of the datasets and the relative classification performance of different algorithms and parameter settings. 

At the other end of the spectrum, two-dimensional toy datasets are a common approach to the evaluation of clustering methods~\cite{franti2018k,handl2006feature}. Typically, these datasets have been handcrafted to illustrate simple capabilities or properties of clustering algorithms. They remain popular due to their simplicity and because they enable intuitive visualization of results, rendering them highly effective at clearly exhibiting a particular property and consequent algorithm behaviour (see e.g.\ the ``moons'' dataset\footnote{Example: \url{https://rdrr.io/cran/clusterSim/man/shapes.two.moon.html}}). Despite these advantages, the scenarios that toy datasets depict are often too contrived.

Although synthetic data is commonly generated \emph{ad hoc} for individual work, some generators have been explicitly designed to have broader utility, with more complex properties than toy data. The generator proposed by Qiu and Joe~\cite{qiu2006generation} (abbreviated to \emph{QJ}) uses a geometric framework for cluster placement. The measure of separation used (proposed in~\cite{qiu2006separation}), provides a measure of the spatial separation between clusters. By adjusting this minimum amount of separation allowed, the user can generate datasets that have different amounts of separation between clusters, providing some loose control over the resulting difficulty of the datasets. To do this, the covariance matrices are iteratively scaled until this minimum separation is achieved. Although useful and geometrically interpretable, this provides a single perspective of cluster structure. Their generator can, however, embed additional complexity through adding noise~\cite{qiu2006generation}. Despite having several useful parameters to customize generated problem difficulty, the generator (available as an R package\footnote{\url{https://cran.r-project.org/web/packages/clusterGeneration/index.html}}) is not easily extended to incorporate other measures of cluster structure.

Handl and Knowles~\cite{handl2005generator} (abbreviated to \emph{HK}) created two generators (named ``gaussian'' and ``ellipsoidal'') that have been used extensively for the creation of synthetic clusters. At their core, these generators aim to generate clusters that are as compact as possible with either no (``gaussian'') or minimal (``ellipsoidal'') overlap. The ``gaussian'' generator uses a trial-and-error scheme where Gaussian clusters are randomly generated, and rejected if they overlap with existing clusters. The ``ellipsoidal'' generator was designed specifically for generating elongated clusters in higher dimensions, and uses an EA to optimize cluster location such that overall variance in the dataset is reduced while penalizing overlap. Despite the popularity of these generators, their design is rigid with many hand-tuned parameters, and without consideration or easy scope for extension to consider additional or alternative aspects of cluster structure.

In this paper, we therefore propose a more general evolutionary framework for generating new synthetic benchmarks for cluster analysis. Our work tackles two crucial steps for the construction of diverse datasets: (i)~the explicit identification of problem characteristics that are relevant to differentiating clustering algorithm performance and (ii)~the design of an optimizer that is sufficiently general to evolve benchmarks for a range of criteria. Specifically, we experiment with two different types of criteria: approximating a desired target value of problem characteristics, and directly maximizing performance differences between algorithms. The latter approach helps highlight the strengths and weaknesses of the contestant methods, providing direct insight into their individual inductive bias. 
To evaluate the performance of our approach, we compare datasets obtained from our framework against multiple generators/dataset collections in terms of both the spread of performance across clustering algorithms, and the variance across our proposed set of problem features. This work greatly extends an initial proof-of-concept outlined in \cite{shand2019evolving}, generalizing the framework and introducing the ability to generate datasets directly for particular clustering algorithms.
%use of optimization as a means to evolve benchmarks that meet criteria directly linked to the performance of clustering algorithm performance. 

% ---------------------------------------------- %
% -------------------- HAWKS ------------------- %
% ---------------------------------------------- %
\section{HAWKS} \label{sec:hawks}

We now proceed to introduce the details of our framework, HAWKS\footnote{The name is derived from the first letter of the surnames of the authors.}, for the generation of diverse, complex datasets. HAWKS uses an EA to evolve a population of datasets, where the objective function and constraints work together to vary properties of the resulting datasets that affect clustering algorithm performance. %A simple outline of this process is shown in Fig.~\ref{fig:hawks_summary}.
For each component of HAWKS, we discuss its general role in dataset generation, and motivate the specific design choice made for the experiments presented in this paper. An overview figure of the inputs and components can be found in Fig.~S-\zref{supp-fig:hawks_summary}. The open-source code can be found online\footnote{\url{https://github.com/sea-shunned/hawks}}.

% \begin{figure*}
% 	\centering
% 	\includegraphics[width=0.95\textwidth]{supp_imgs/hawks_summary.pdf}
% 	\caption{Overview of HAWKS. The input parameters (left) affect different components of HAWKS, which themselves and form the hierarchy of clusters, which constitute an individual dataset, which in turn form the population that the EA interacts with. A simple example and representation is shown on the right.}
% 	\label{fig:hawks_summary}
% \end{figure*}

\subsection{Representing a dataset} \label{ssec:hawks_representation}
In HAWKS, a dataset is represented by a set of clusters, which are themselves defined by the parameters of a given distribution. With this level of abstraction, we avoid handling of individual data points, and provide an intuitive manipulation of clusters through adjustment of distribution parameters. In principle, this approach allows for the inclusion of a variety of cluster generating models, given suitable initialization and variation operators for each distribution have been defined (see Sections~\ref{ssec:hawks_init} and~\ref{ssec:hawks_variation}). For example, adding clusters generated from a uniform distribution could be parameterized with $D \times 2$ lower and upper bounds, with a given range for initialization, and mutation operators that e.g. adjust the bounds or rotate the hypercube.

For the experiments reported in this paper, each cluster is represented by a multivariate Gaussian distribution, due to their relative simplicity and prominence in a machine learning context. For a dataset in $D$ dimensions, each cluster is therefore encoded by a ($\hawksmean,\hawkscov$) pair, which we will refer to as a single gene. Here, $\hawksmean$ is the $D$-dimensional mean vector and $\hawkscov$ is the symmetric $D \times D$ covariance matrix. A dataset with $K$ clusters is therefore represented as a genotype composed of $K$ genes. %, or $D+\frac{D \times (D+1)}{2}$ decision variables.
%or $2K$ (sets of) decision variables.  
In the current implementation, $K$ is set \emph{a priori} as an input parameter.

\subsection{Initializing a population of datasets} \label{ssec:hawks_init}
The first step of HAWKS is to create an initial population of datasets/individuals. The sizes of the clusters can be controlled in two ways: (i)~all clusters can be of equal size or (ii)~each cluster is of random size (with an optional minimum size to avoid clusters small enough to be considered  outliers)\footnote{Owing to its non-triviality, the method to generate random cluster sizes with a minimum size that overall sum to $N$ is described in the supplementary material (Section~\zref{supp-ssec:cluster_sizes}).}. In both cases, the user predefines the total number of data points $N$. The cluster sizes remain fixed across all individuals for the remainder of the evolution, to avoid interference with the fitness (described in Section~\ref{ssec:hawks_fitness}) and focus of the search on the distribution parameters only.

All other aspects of the cluster-level initialization are specific to the type of distribution used; here, we describe our approach for Gaussian clusters. The initial means are sampled from a $D$-dimensional uniform distribution, i.e.\ for the $i$th cluster $\hawksmeanarg{i} \sim U(0, \beta_{1})^{D}$ where $\beta_{1}$ is the upper threshold to control the initial sampling space for the means. In HAWKS, the covariance matrix is defined for the $i$th cluster as $\hawkscov[(i)] = \hawksrotationarg{i} \hawksscalingarg{i} \hawkscovaxisarg[(i)] {\hawksrotationarg{i}}^{\intercal}$, where \hawksrotationarg{k} and \hawksscalingarg{k} are the $k$th rotation and scaling matrices, respectively, and \hawkscovaxisarg[(k)]~is the $k$th axis-aligned covariance matrix (i.e.~a diagonal matrix that consists of only the variances). These variances are sampled similarly to the means, i.e.~$\hawkscovaxis = \diag(U(0, \beta_{2})^{D})$ where $\beta_{2}$ is the upper threshold to control the initial sampling space for the variances. The scaling matrix is set to the identity matrix at this stage. The covariances are then obtained via rotation using a random rotation matrix (\hawksrotation), which is drawn from the Haar distribution~\cite{stewart1980} to generate a valid covariance matrix. This method permits generation of clusters with a variety of shapes and orientations, thus ensuring the initial population has a diverse set of individuals.

This approach, while more complex than constructing a covariance matrix with random values, allows both more intuitive parameterization and the ability to separately modify individual components of the covariance matrix, which we later exploit for mutation (Section~\ref{ssec:hawks_variation}). Other approaches, such as that proposed in \cite{maitra2010sim}, may allow for initialization of clusters according to their measure of overlap, though this somewhat deviates from the distribution-agnostic approach of HAWKS.

\subsection{Computing the fitness of a dataset} \label{ssec:hawks_fitness}

The quantification of fitness plays a key role in focusing the search on those datasets that are deemed to present interesting clustering benchmarks. It is thus one of the most vital design decisions in a given generator, but is complicated by our limited understanding of the desirable properties of clustering benchmarks, as discussed in Section~\ref{ssec:bg_clustering}. Therefore, our framework is designed to allow for the interchangeable use of different fitness functions, providing scope for a variety of choices to be trialled.

%inevitably narrow (to some degree) the space of possible datasets that can be generated. The %modularity of our framework, however, allows for the design of different fitness functions which can be used interchangeably for different tasks. 
In this paper, we introduce and experiment with two different \emph{modes} of optimization which support the generation of datasets for distinct goals. The first, named \emph{Index} mode, optimizes the datasets towards a user-defined target value for a given cluster validity index. This provides control of the broad difficulty of the datasets, where the nature of the difficulty is governed by the properties of the index used. The second, named \emph{Versus} mode, directly optimizes datasets such that they maximize the performance difference between two clustering algorithms. By generating a range of datasets that are simple and difficult for specific algorithms we can ``stress-test'' them, revealing their relative strengths and weaknesses. We describe these approaches in more detail in the following sections.

\subsubsection{Index Mode}

Cluster validity indices capture the amount of structure of a partition, in terms of the underlying data distribution. Given knowledge of the generating model (i.e.\ the true partition), they can thus act as a proxy for the ease of recognizing this structure in a given dataset. 

As previously discussed, there are many different cluster validity indices, each with a slightly different perspective of how cluster structure should be quantified. Previous work, including~\cite{arbelaitz2013validity}, could not conclude superiority of any single validity index. Thus, to obtain a comprehensive understanding of the difficulty of a given dataset, many cluster validity indices could potentially be used in combination~\cite{akhanli2020comparing}. In the context of a fitness function, this could be done in the form of an aggregation of indices or through formulation as a many-objective problem. Here, we opt for a middle ground, limiting ourselves to the definition of fitness through a single cluster validity index, but allowing for the incorporation of additional considerations (which could include validity indices) through the use of constraints (discussed in Section~\ref{ssec:hawks_constraints}).

For the experiments reported in this paper, we use the silhouette width~\cite{rousseeuw1987silhouettes} as a representative example of a validity index. It is an established, widely-used method and has two particular characteristics that makes it a promising choice for use here: (i)~it is a very rich measure that provides information at multiple level of resolutions (the individual data point, the cluster, and the partition-level); and (ii)~as it is bounded in the range $[-1, 1]$, it can be easily compared across individuals and even runs (when a similar dimensionality is used), and it is readily interpretable to the user.  

The silhouette width is a combination internal validity index, as it measures a ratio of intra-cluster compactness and inter-cluster separation \cite{handl2005computational}. For a single data point, \datapoint[][i]{x}, it is defined as:
\begin{equation} \label{eq:bg_clusts_silhwidth}
	s(\datapoint[][i]{x}) = \frac{b(\datapoint[][i]{x}) - a(\datapoint[][i]{x})}{\max\{a(\datapoint[][i]{x}), b(\datapoint[][i]{x})\}}.
\end{equation}

Here, $a(\datapoint[][i]{x})$ represents the cluster compactness (with respect to \datapoint[][i]{x}) and is the average distance from \datapoint[][i]{x}~to all other data points in its cluster. The separation between clusters is represented by $b(\datapoint[][i]{x})$; for data point \datapoint[][i]{x}~this is defined as the minimum of the average distances to all data points in every other cluster. The silhouette width is calculated for all $N$ data points in dataset $X$, and an average is taken to obtain the overall silhouette width:
\begin{equation} \label{eq:bg_clusts_silhall}
	\silhall = \frac{1}{N} \sum_{i=1}^N s(\datapoint[][i]{x}).
\end{equation}

A value of 1 represents very compact and well-separated clusters, whereas a negative silhouette width value indicates that points in different clusters are not well-separated (and that their cluster membership should be changed). 

Independent of the validity index chosen, direct maximization or minimization of such an index would always lead to the evolution of datasets that are trivially separable or fully unstructured, respectively. HAWKS therefore requires input in the form of a desired target value, allowing direct modulation of the desired level of structure.

In other words, and using the example of the silhouette width, a target value (denoted \silhtarget) is specified by the user and datasets are then optimized to meet this target value. This is achieved by minimizing the absolute difference between \silhtarget~and \silhall, defined as:
\begin{equation} \label{eq:hawks_fit-func}
	\min\; f(\hawksmeanarg{1},\hawkscov[(1)],\ldots,\hawksmeanarg{K},\hawkscov[(K)]) \equiv \min\; \left | \silhtarget - \silhall \right |.
\end{equation}

% A disadvantage to the use of the silhouette width as an objective is its complexity: $O(DN^{2})$. Utilizing the fact that only a subset of pairwise distances are changed for a given individual each generation (clusters that are modified by the genetic operators), the silhouette width is partially evaluated to reduce computation.

\subsubsection{Versus Mode} \label{sssec:versus}
The \emph{Index} mode provides us with the ability to generate benchmarks that meet specific thresholds of user-defined validity criteria. However, shared assumptions between validity indices and clustering algorithms themselves can make it difficult to generate datasets with properties that specifically challenge a given algorithm. This is particularly the case when working with clustering techniques whose inductive biases are poorly understood, e.g.\ self-organizing approaches or those that utilize deep learning~\cite{xie2016}.

To perform ``controlled experimentation'', a more direct link between dataset generation and algorithm performance may be needed \cite{hooker1995testing}. Our \emph{Versus} mode tries to address this by directly optimizing the performance difference between two algorithms, thereby exploiting the strengths of one algorithm, the weaknesses of the other, or a combination of the two. 

We re-formulate the fitness function such that we maximize the difference between the scores of the `winning' algorithm ($\mathcal{A}_{w}$) and the `losing' algorithm ($\mathcal{A}_{l}$), using a scoring function $\phi$:
\begin{equation} \label{eq:versus_fitness}
	\max f(\hawksmeanarg{1},\hawkscov[(1)],\ldots,\hawksmeanarg{K},\hawkscov[(K)]) \equiv \max (\phi(\winneralg) - \phi(\loseralg)).
\end{equation}

While any scoring function can be used, we use the generating model of the clusters to determine the labels, allowing the use of external validity indices. We therefore use the Adjusted Rand Index (ARI)~\cite{hubert1985comparing}, an adjusted-for-chance version of Rand index, to compare the two partitions. The ARI measures the co-occurrences of cluster assignments between two partitions, which in our case corresponds to the output of a given clustering algorithm, and the ground truth. The upper bound of 1 represents identical assignment, whereas 0 indicates random assignment. As the ARI considers only the assignment of points, it can be used with any clustering algorithm and has no preference of structure.

\subsection{Augmenting cluster properties using constraints} \label{ssec:hawks_constraints}

As previously discussed, it is difficult to define a single fitness measure that represents all properties that may be desirable in a benchmark dataset. Therefore, our framework uses constraints to allow for the integration of additional considerations, with two main aims: (i)~to avoid the generation of trivial datasets that are e.g.\ simple for any algorithm, or too noisy to be clustered, and (ii)~to introduce additional properties that balance limitations of individual fitness functions or further enhance diversity in the datasets obtained. 
To control optimization of the fitness and constraints, we use stochastic ranking \cite{runarsson2000} to balance the mutual satisfaction of both (this is discussed further in Section~\ref{ssec:hawks_selection}). An upper or lower threshold for these constraints are given as input to HAWKS, for which the quadratic loss function is used to calculate the penalty incurred, as per the original work \cite{runarsson2000}.

In the following, we discuss two constraints introduced to meet these aims by directly accounting for local overlap between clusters and for cluster shape (as measured by cluster elongation --- for a different generating model, other choices may be plausible). These choices of constraints have to be seen in the context of the fitness functions adopted for our experiments: when using the silhouette width in \emph{Index} mode, the fitness provides a powerful overall perspective of separation between clusters. However, the reliance on averaging can lose fine-grained information about local variability, e.g.\ lack of inter-cluster separation for small clusters. Furthermore, the silhouette width does not directly capture information about cluster shape, while differences in cluster shapes are known to drive some performance differences between algorithms. Similarly, for the \emph{Versus} mode, a performance difference between the algorithms is sought with no explicit concern for the underlying cluster properties. While this exploratory approach is desired, there are situations where this can lead to cluster structures that are not meaningful (e.g.\ clusters that are completely overlapping) but one algorithm is still perceived to be favoured due to arbitrary artefacts.

% Although the silhouette width provides an overall perspective of separation between the clusters, it is just one cluster property. Other properties are introduced as constraints, as these can influence the final datasets and (from a practical perspective) can be easily parameterized and added modularly. At present, we consider two distinct constraints.

\subsubsection{Overlap}
Real-world data typically does not contain cleanly separated clusters, either due to noise or reflective of the underlying relationships between variables. Clustering algorithms differ significantly in their ability to handle some degree of cluster overlap, and control over this aspect of our benchmark is therefore important.
At the extreme ends, a benchmark set with very well-separated clusters may be too simple to detect any performance differences between algorithms; equally, a dataset where clusters fully overlap will be of no use. 

The definition of overlap is not purely objective, and can be implemented in different ways. To ensure HAWKS remains agnostic to cluster distribution, we focus on methods that calculate the overlap using the sampled data points rather than the distributions themselves. While the latter may be advantageous, there are multiple measures even for well-defined distributions, i.e. the Gaussian clusters we use here \cite{maitra2010sim,nowakowska2014tractable,qiu2006separation,sun2011}. Fr{\"a}nti and Sieranoja~\cite{franti2018k} defined a data point as overlapping if it is closer to a centroid of a different cluster than to the centroid of its own cluster. This definition, however, does not extend to highly eccentric clusters. We use a definition similar to~\cite{handl2005generator}, where a data point is considered as overlapping if its nearest neighbour belongs to a different cluster. In addition to avoiding the compactness bias introduced by using centroids, this definition has the specific benefit of countering an inherent limitation of the silhouette width, where a high average silhouette width value can be driven by having large clusters that are very well-separated and fails to sufficiently reflect the presence of very small but highly overlapping clusters.

We calculate the overlap as the percentage of data points whose nearest neighbour is in a different cluster:
\begin{equation} \label{eq:hawks_overlap}
	\overlap = 1 - \frac{1}{N}\sum_{\datapoint{x} \in X} \mathds{1}_{\cluster{k}}(n_{\datapoint{x}})
\end{equation}
where $\cluster{k}$ is the cluster that data point $\datapoint{x}$ belongs to, $n_{\datapoint{x}}$ is the nearest neighbour of $\datapoint{x}$, and $\mathds{1}(\cdot)$ is the indicator function that is 1 if $n_{\datapoint{x}} \in \cluster{k}$ and 0 if $n_{\datapoint{x}} \notin \cluster{k}$.

\subsubsection{Elongation}
The elongation (or, more specifically, the eccentricity) of clusters can pose problems for compactness-based algorithms such as K-Means, and thus presents an additional aspect that we are keen to control. Although our initialization parameters can adjust initial cluster eccentricity, by using an explicit constraint we can encourage (or penalize) this further during the evolution.

Our definition of this constraint is specific to the generating model used, and different definitions are possible. As previously mentioned, each full covariance matrix (\hawkscov) in our cluster representation is separated into the axis-aligned variances ($\hawkscovaxis$) and a rotation matrix. As the variances on the diagonal of \hawkscovaxis~are the eigenvalues of the full covariance matrix, i.e.~$\hawkscovaxis = \diag(\lambda_{1},\ldots,\lambda_{D})$, the ratio of the maximum and minimum of these eigenvalues gives us a measure of the cluster eccentricity. As even a single eccentric cluster can pose challenges for compactness-based algorithms, we take the minimum of these ratios across all $K$ clusters, i.e.:
\begin{equation} \label{eq:hawks_ratio}
	\eigratio = \max_{\forall\: k \in \{1,\ldots,K\}} \frac{| \lambda_{\text{max}} (\hawkscov[(k)]) |}{| \lambda_{\text{min}} (\hawkscov[(k)]) |},
\end{equation}
where $\lambda_{\text{max}}(\hawkscov[(k)])$ and $\lambda_{\text{min}}(\hawkscov[(k)])$ are the maximum and minimum eigenvalues of $\hawkscov[(k)]$, respectively.

\subsection{Perturbing a dataset} \label{ssec:hawks_variation}

\begin{figure}
	\centering
	\subfloat[Crossover\label{fig:hawks_crossover}]{\includegraphics[width=0.75\columnwidth]{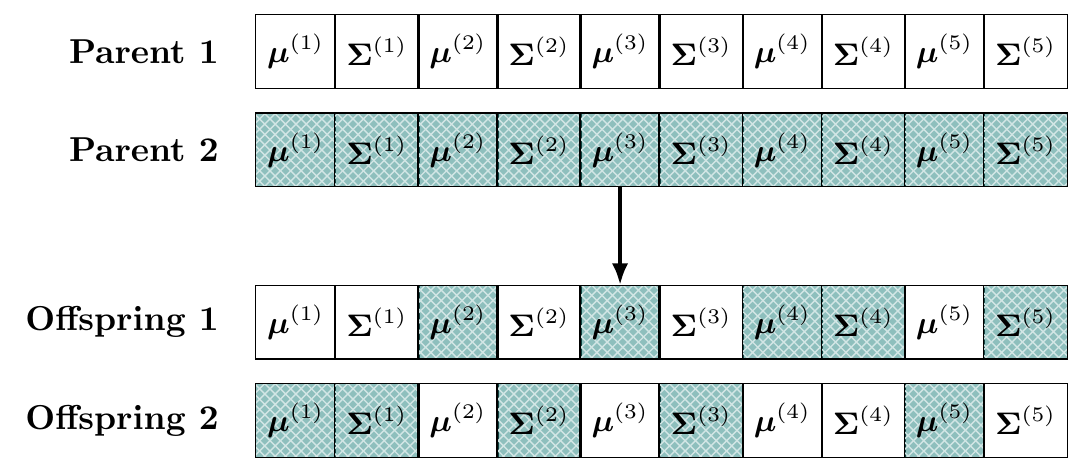}}\hfill
	\subfloat[Mutation\label{fig:hawks_mutation}]{\includegraphics[width=0.75\columnwidth]{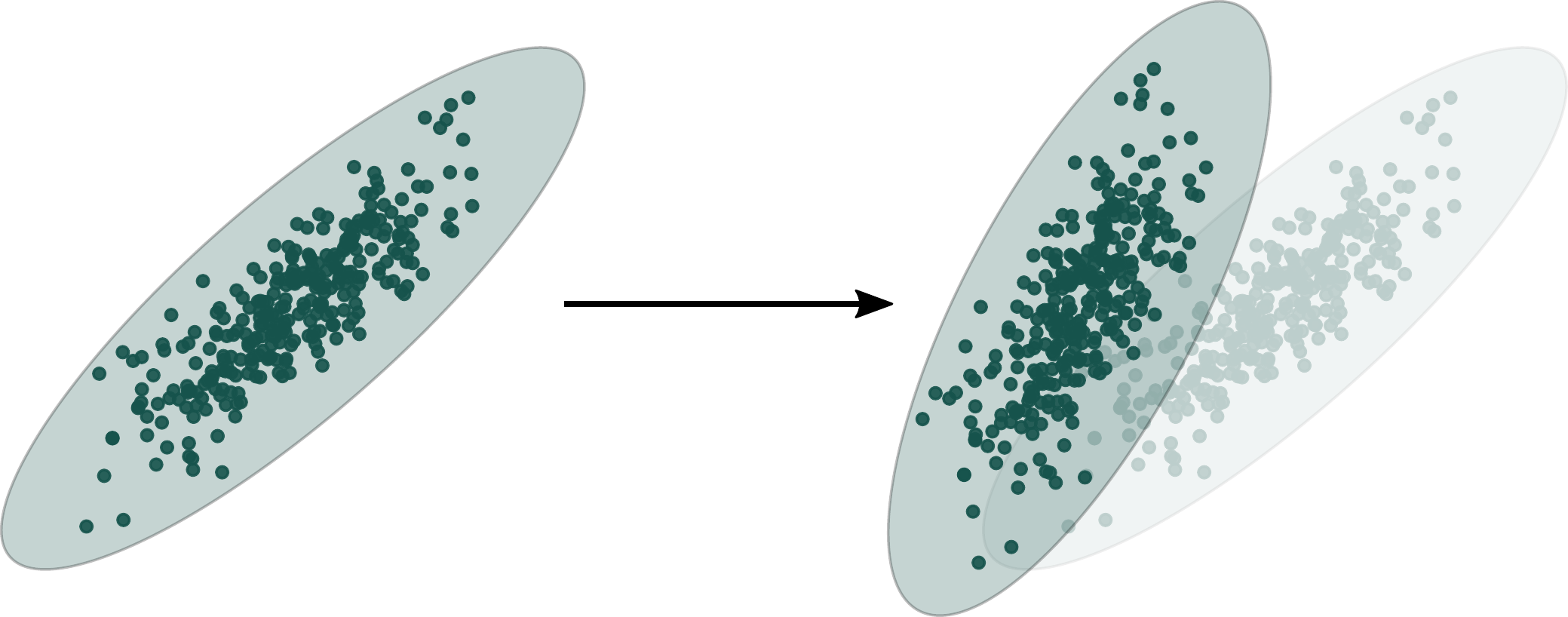}}\hfill
	\caption{Illustrations of the HAWKS genetic operators. Uniform crossover (a) is used to independently swap the means and covariances between two individuals. In (b), a single cluster is mutated where both the location (mean) and shape (covariance) have been randomly perturbed (the original cluster is shown again faded to illustrate the relative differences).}
	\label{fig:hawks_operators}
\end{figure}

As is typical in an EA, the variation operators in HAWKS provide a further opportunity to integrate domain knowledge and focus the search. Specifically, a close alignment between the generating model used (i.e.\ the representation of an individual cluster and partition) and the operators is crucial to optimization performance and the types of datasets that can be obtained. In the following, we describe the variation operators designed for the multivariate Gaussian distributions utilized as the generating model in our experiments. 

\subsubsection{Crossover}
For recombination between individuals, HAWKS uses a high-level uniform crossover scheme where the components defining each cluster distribution (in our Gaussian case, $\hawksmean$ and $\hawkscov$) can be swapped separately, as illustrated in Fig.\ \ref{fig:hawks_crossover}.

\subsubsection{Mutation}
Meaningful mutations of a cluster require careful design of a geometrically relevant operator that is relevant for the distribution being used. In our Gaussian case, we use a separate operator for the mean and covariance terms.

A key issue identified in~\cite{shand2019evolving} was the drift of the mean operator in higher dimensions. The original operator shifts the mean to a random nearby point; at higher dimensionality there are an increasing number of directions that point away from other clusters, and a random walk is thereby likely to increase the silhouette width at most steps. In the supplementary material (Section~\zref{supp-sec:mutating}), we test multiple new mutation operators (drawing inspiration from operators used in particle swarm optimization and differential evolution) to address this issue by directly considering the location of other clusters.

The resulting operator selected and used throughout this paper utilizes concepts from particle swarm optimization \cite{eberhart1995particle}, where a combination of other centroids and a global representative is used to embed direction either towards or away from existing clusters, directly affecting the fitness. The new mean, \hawksmeanargnew{i}, is obtained as follows:
\begin{equation} \label{eq:hawks_mean_mut}
	\small
    \hawksmeanargnew{i} = 
    \begin{cases}
        \hawksmeanarg{i} + [w_{1} (\hawksmeanarg{n} - \hawksmeanarg{i}) + w_{2} (\hawksmeanglobal - \hawksmeanarg{i})] & \text{if } p \leq 0.5 \\[0.2cm]
        \hawksmeanarg{i} - [w_{1} (\hawksmeanarg{n} - \hawksmeanarg{i}) + w_{2} (\hawksmeanglobal - \hawksmeanarg{i})] & \text{if } p > 0.5, \\
	\end{cases}
\end{equation}
where \hawksmeanarg{i} is the current cluster mean, \hawksmeanarg{n} is the mean of another randomly selected cluster, \hawksmeanglobal~is the global mean across all data points, $w_{1}$ and $w_{2}$ are random weighting coefficients in the range $[0,1]$, and $p$ is a random coin-flip to decide whether to move away from or towards this weighted combination of an existing cluster mean and the global mean.

The covariance governs the shape of the cluster, affecting both the overlap between clusters and the capabilities of compactness-based algorithms (such as K-Means). To mutate the covariance, we rotate the cluster and scale its eigenvalues, effectively changing the shape of the cluster. Similar to the initialization, the rotation matrix is drawn from the Haar distribution~\cite{stewart1980}, though here it is raised to a fractional power to avoid complete reorientation of the cluster. The scaling matrix, $S$, is drawn from a Dirichlet distribution in order to ensure that the resulting determinant is unchanged i.e.~$\det(\hawkscovaxis) = \det(S \cdot \hawkscovaxis)$\footnote{This is described in further detail in Section~\zref{supp-ssec:scaling_matrix}.}. Using an analogy, this has the same effect as rotating a balloon and applying pressure to the principal semi-axes, thereby changing the shape while maintaining the volume. The combined effect of both mutation operators is illustrated in Fig.~\ref{fig:hawks_mutation}.

\subsection{Selecting a dataset} \label{ssec:hawks_selection}

As previously discussed, we use stochastic ranking~\cite{runarsson2000} to balance the satisfaction of the objective and constraint(s), which may be complementary or impossible to both fully satisfy. In brief, stochastic ranking using a bubble-sort-like procedure to rank the population, where individuals are compared on their fitness with probability $\stochranking$, and are otherwise compared on their constraint violation. By adjusting $\stochranking$, we can effectively weigh the satisfaction of the objective or the constraints, providing another way of controlling the properties of the resulting datasets. For example, if using only the overlap constraint in \emph{Index} mode, setting $\stochranking$ to a higher value will add selection pressure to datasets with a silhouette width closer to $\silhtarget$, potentially at the cost of a higher degree of overlap. Thus, unlike the traditional use of stochastic ranking that uses a narrow range of values for $\stochranking$ \cite{li2016stochastic, runarsson2000} to avoid too much weighting towards the infeasible or feasible regions, we can use the entire range (as datasets that heavily violate the constraints may still be useful).

Similar to the original work~\cite{runarsson2000}, for environmental selection we use stochastic ranking to select the top $|\mathcal{P}|$ individuals from the sorted pool of parents and offspring. For parental selection, we use standard binary tournament where the rank is used to determine the winner (in lieu of using only the fitness) to ensure a continued selection pressure towards individuals that best satisfy the fitness and constraints as weighted by $\stochranking$.

% ---------------------------------------------- %
% -------------------- Setup ------------------- %
% ---------------------------------------------- %
\section{Experimental Setup} \label{sec:exp_setup}

In order to assess the capabilities of HAWKS, we have three primary aims in our experiments: (i)~compare the diversity of performance of multiple, distinct clustering algorithms on datasets from HAWKS against that of other popular generators or dataset collections; (ii)~compare the diversity across a distinct set of problem features, to see if datasets from HAWKS cover a wider space of properties than other datasets; and, (iii)~gain insights into the algorithms themselves, utilizing the \emph{Versus} mode of HAWKS to directly challenge clustering algorithms.

The remainder of this section describes our proposed problem features (Section \ref{ssec:problem_features}), the generators and dataset collections we compare against (Section \ref{ssec:other_generators}), and the relevant parameters HAWKS used in each experiment (Section \ref{ssec:hawks_configs}).

\subsection{Problem features} \label{ssec:problem_features}
In order to measure dataset diversity (with respect to their inherent properties), we need to define a set of problem features describing properties relevant to algorithm performance. This is central to the algorithm selection problem (ASP), as these features are used to predict which algorithm is best for a given problem instance. These problem features are also vital for the creation of an instance space that visualizes the datasets, allowing identification of areas in the space where particular algorithms are most suitable.

In previously discussed work on the ASP for clustering, the problem features (also referred to as ``meta-features'') used are statistical or information-theoretic, and not specific to clustering~\cite{ferrari2015clustering,soares2009analysis}. The problem features previously used in \cite{shand2019evolving} were also too simplistic, and coincided with parameters available in HAWKS, reducing the utility of the instance space.

% Broadly, the problem features should measure characteristics about the dataset that enable successful prediction of which algorithm is best for a given dataset based on these features.
As there are many properties that influence performance for a given clustering algorithm, and for most algorithms this aspect is not fully understood, a full complementary set of problem features is arguably impossible~\cite{arbelaitz2013validity,hennig2015true,hennig2018thoughts}. Nonetheless, going beyond simple statistical measures of the data by incorporating measures specific to clustering (especially those that make use of the generating model) should improve the discriminative power of the problem features.\footnote{As discussed earlier, use of the ground truth in these measures comes at the cost of limiting applicability to datasets where such knowledge is unavailable; this is a major issue for ASP applications but is not our intended focus here.} The remainder of this section describes our proposed set of problem features for this task.

To this end, we use the following problem features: average cluster eccentricity, connectivity, dimensionality, entropy of cluster sizes, number of clusters, average silhouette width, and the standard deviation of the silhouette width. For a full explanation of these problem features (both their formulation and the motivation behind their use), see Section~\zref{supp-ssec:problem_features}.

\subsection{Other generators and datasets} \label{ssec:other_generators}

To provide context for the diversity in performance and problem features for the datasets produced by HAWKS, we compare against multiple generators or collections of popular datasets that have been previously used to evaluate clustering algorithms. We briefly describe these below, though more details about the parameters used for these datasets, and how they compare (in terms of their size, dimensionality etc.) can be found in the supplementary material (Table~S-\zref{supp-tab:dataset_params}) and their respective papers.

\subsubsection{HK}
This is a collection of 350 datasets generated using the ``ellipsoidal'' generator proposed in~\cite{handl2005generator}, and used as a benchmark in~\cite{Garza-Fabre2017}. For 35 unique combinations of the number of clusters and dimensionality, 10 datasets are generated.
%$K \in \{10, 20, 40, 60, 80, 100, 120\}$ and $D \in \{20, 50, 100, 150, 200\}$.

\subsubsection{QJ}
We use the set of 243 datasets generated using the parameters proposed by Qiu and Joe~\cite{qiu2006generation}. The authors calculated three target separation values using their measure of separation: ``close structure'', ``separated'', and ``well-separated''. Further complexity to these datasets is added by specifying varying proportions of noisy variables, alongside small variation in the number of dimensions.

\subsubsection{SIPU}
Fr{\"a}nti and Sieranoja~\cite{franti2018k} introduced a benchmark consisting of multiple sets of clustering datasets, of which we use the `S-sets', `A-sets', and `G2 sets'. These sets were originally intended to stress-test compactness-based algorithms, of which K-Means exhibited a wide range of performance. The `S-sets' are all 2D data where $N=5000$ and $K=15$, but have different degrees of overlap between the clusters (determined by the aforementioned method of the closest centroid). The `A-sets' are also 2D data with varying numbers of (equally-sized) clusters. The `G2 sets' of datasets consist of two Gaussians with varying degrees of overlap, constant size ($N=2048$) and varying dimensionality (chosen across the range $2^{1},2^{2},\ldots,2^{10}$). The variance of overlap and high number of dimensions in this benchmark presents a variety of challenges for clustering algorithms.

\subsubsection{UCI}
The UCI Machine Learning Repository \cite{Dua2017} is a popular source of datasets used for machine learning. As noted in \cite{von2012clustering,franti2018k} the class labels do not necessarily translate to meaningful cluster labels, and thus these datasets may not be the most suitable for clustering. Nonetheless, they have seen extensive use in the clustering literature, and we include them for completeness. Specifically, we use the subset of 20 datasets used by Arbelaitz et al.~\cite{arbelaitz2013validity}.

\subsubsection{UKC}
These 8 real-world datasets, curated in~\cite{Garza-Fabre2017}, are the (anonymized) locations of different crimes. Alongside the \emph{UCI} datasets, these will help provide insight into whether there are significant differences between the real-world and synthetic data (in terms of their problem features or clustering algorithm performance).

\subsection{HAWKS configurations} \label{ssec:hawks_configs}

In this section we highlight the key parameters for HAWKS, separating those that are common across all experiments and those adjusted for different modes. In particular, while the core EA parameters (population size, mutation probability etc.) remain the same across modes, in our experiment for the \emph{Index} mode we vary the objective function target and constraint parameters to generate different datasets. In contrast, the constraint parameters remain fixed for the \emph{Versus} mode experiment as the variation comes from selecting different pairs of clustering algorithms. The full configurations for both sets of experiments can be found on GitHub\footnote{\url{https://github.com/sea-shunned/hawks_configs}}.

\subsubsection{Common parameters}
In both \emph{Index} and \emph{Versus} mode experiments, the following core EA parameters are used: $G_{\text{max}}=100$ generations, $\mathcal{P}=10$ individuals, $p_{c}=0.7$ crossover probability, and $p_{m} = 1/K$ mutation probability for the mean and covariance mutation operators. The low population size was previously found to be sufficient for both diversity and convergence \cite{shand2019evolving}. {A sensitivity analysis for increasing both the number of generations and population size can be found in Section~\zref{supp-ssec:sensitivity}.

\subsubsection{Solution selection}
At the end of a run, a single dataset needs to be selected. Here, we simply select the individual with the highest fitness. Different choices are possible, e.g., due to our use of stochastic ranking, the individual with the highest (sorted) rank could be selected instead (though this had little effect on the experiments in this paper).

\subsubsection{Analyzing benchmark dataset diversity (Index mode)}
To test the ability of HAWKS to produce a variety of datasets, we vary several parameters to encourage coverage of our feature space:
\begin{itemize}
    \item A poor and a high silhouette width ($\silhtarget \in \{0.45,0.9\}$).
    \item Two upper thresholds for the \overlap~constraint ($\overlap \leq \{0,0.1\}$) to penalize any overlap, or penalize if more than 10\% of the data points overlap.
    \item Two lower thresholds for the eccentricity constraint ($\eigratio \geq \{1,50\}$) to allow for any amount of eccentricity, or encourage all clusters have some eccentricity. 
    \item Two levels of dimensionality ($D \in \{2,50\}$), dataset size ($N \in \{500,2500\}$), and number of clusters ($K \in \{5,30\}$).
\end{itemize}
Of note is the use of $\stochranking=0.5$, which weights the fitness and constraints equally. By varying $\silhtarget$, we directly attempt to generate datasets that either have poor cluster separation or are well-separated. %While in isolation these two values may not produce a large amount of variety, 
Different values of the overlap and eccentricity constraints help further modulate the level of separation and the minimum cluster eccentricity. HAWKS is run 7 times for each of the 64 unique combinations of parameters listed above, resulting in 448 datasets. 
%The small number of runs per configuration is to avoid flooding the instance space with HAWKS-generated datasets, which (as PCA is used) could bias towards improved coverage across the space. The number of datasets is still more than those contained in the other collections, however. In order to be well-spread across the space, variation in the problem features is needed (and is not dependent upon the number of datasets.)

As we aim to measure diversity in clustering performance, we need an objective way to measure this. We therefore use the ARI (see Section~\ref{sssec:versus}) to compare the ground truth (which we know for all datasets used here) with the assignments from each algorithm. For the clustering algorithms themselves, we select four well-established algorithms with distinct properties and inductive biases, allowing us to assess the diversity of challenges that the datasets pose. These are:
\begin{itemize}
	\item Average-linkage --- A hierarchical clustering method that uses the average distance between all points in a cluster when deciding which are the closest (and thus should be merged).
	\item Gaussian mixture models (GMM) --- Probabilistic models which try to represent subpopulations (of the data) through a number of Gaussian distributions. To obtain a crisp clustering, each point is assigned to the cluster with the highest probability.
	\item \kmeanplus~--- Proposed by Arthur and Vassilvitskii~\cite{arthur2007k} to improve K-Means, the initialization scheme probabilistically selects cluster centres such that points that are further away from existing cluster centres are more likely to be selected.
	\item Single-linkage --- In contrast to average-linkage, single-linkage considers the distance between two clusters as the shortest distance from \emph{any} member of one cluster to any member of another.
\end{itemize}
\emph{Scikit-learn}~\cite{scikit-learn} was used for all clustering algorithms, using Euclidean distance and default parameters. To assess the maximum potential of each algorithm, we provide the true number of clusters for each dataset. Owing to the propensity of the linkage algorithms (particularly single-linkage) to be side-tracked by singleton clusters, we also run average- and single-linkage with double the true number of clusters~\cite{hubert1974approximate}.

\subsubsection{Challenging specific algorithms (Versus mode)}
To ascertain the capabilities of HAWKS' \emph{Versus} mode, we run each of the four clustering algorithms against the others in a head-to-head, with either as the `winner' and `loser', respectively.

Some pairings of algorithms (e.g.~\kmeanplus~vs.~GMM\footnote{Referring to Eq.~\ref{eq:versus_fitness}, we use the format \algversus{\winneralg}{\loseralg}.}) are likely to be more competitive, due to shared capabilities and inductive biases. We expect this to translate into a reduction of the performance differences that can be observed (and thus a weaker fitness gradient). While the constraints are still important in such a scenario, we wish to avoid situations where the optimization is being mainly driven by reducing the constraint penalty. Rather than removing the constraints, we increase $\stochranking$ to emphasize a difference in algorithmic performance. This provides a useful lever for weighting the maximization of the performance difference against producing datasets that do not sacrifice cluster structure (e.g.\ through a high overlap). As such, we use $\stochranking=0.75$ in all head-to-head runs. Further adjustment of this parameter is encouraged for complex algorithm pairings.

Finally, all experiments presented for the \emph{Versus} mode are run using $K=5$ clusters and $N=2000$ data points. The dimensionality is consistently set to $D=2$, ensuring a straightforward visualization of the datasets and enabling us to observe the properties of these datasets without concerns about information loss through projection. %Although this restricts one avenue where differences can be found, it is necessary to assess the efficacy of this approach.

% ---------------------------------------------- %
% ------------------ Results ------------------- %
% ---------------------------------------------- %
\section{Experimental Results} \label{sec:exp_results}

In this section, we explore HAWKS' ability to produce datasets that exhibit a broad range of properties and pose challenges to different clustering algorithms. Separate results for the \emph{Index} and \emph{Versus} modes highlight the general flexibility of our framework.

% ------------------ Benchmark ----------------- %
\subsection{Analyzing benchmark dataset diversity (Index mode)} \label{ssec:benchmark}

This section presents the instance space constructed from our set of 7 problem features applied to the 1,176 datasets from 6 popular collections/generators and our generator HAWKS. We assess diversity by considering (i)~the variation observed across problem features and (ii)~the variation observed in the performance of 4 clustering algorithms. 

First, we use the instance space to understand how the datasets are spread across the space, and if there are distinct patterns either in the sources of the datasets or in the performance of the clustering algorithms. The two principal components of the instance space shown in Fig.~\ref{fig:instance_space} account for $57.56\%$ of the variance, and so there is some information loss in the projection. Nevertheless, it is evident that the components capture a sufficient proportion of the variance to highlight some key differences between datasets.

The instance space in Fig.~\ref{fig:instance_source} uses colour and marker coding to flag up which collection each dataset comes from. Visualizations of how each problem feature varies across the space can be found in Fig.~S-\zref{supp-fig:benchmark_features}. The distinct spread of HAWKS datasets across the central part of the space is encouraging and highlights an appropriate level of diversity in terms of the problem features. Furthermore, the interpretation of the principal components (in terms of the underlying problem features), provides clear guidance on additional experiments that could be conducted to expand coverage in various directions. In contrast, the \emph{QJ} datasets expand across a narrow band, indicating a lack of variance across the problem features. The \emph{SIPU} datasets show a strong banding of instances, indicating that there is little variance among the datasets from each configuration (the separated datasets at the bottom of the space are the `G2 sets', which have a much higher dimensionality than other datasets). The \emph{UCI} datasets are spread across the upper-half of the space, though this is unfortunately due to a lack of structure (which we later explore when looking at clustering algorithm performance), as their higher connectivity indicates that the labels do not line up with a spatial perspective of clustering (shown in Fig.~S-\zref{supp-fig:benchmark_connectivity}). The \emph{UKC} datasets do not seem to represent anything extraordinary with regards to the problem features we use here, leading to the conclusion that either the synthetic datasets used here are not too dissimilar to real-world data or our set of problem features does not capture some aspect of complexity that they uniquely exhibit. Notably, the \emph{HK} datasets are distinct from every other collection, indicating that they have unique characteristics. As shown in Fig.~S-\zref{supp-fig:benchmark_eccen}, the main difference is the average eccentricity of the clusters, which is higher than in any other collection. As these datasets originate from the ``ellipsoidal'' generator, which was designed exclusively to create eccentric clusters in higher dimensions, this is unsurprising. Of interest, however, is whether this leads to a distinct difference in the relative performance of clustering algorithms on that benchmark suite.

\begin{figure}
	\centering
	\subfloat[Dataset source\label{fig:instance_source}]{\includegraphics[width=\columnwidth]{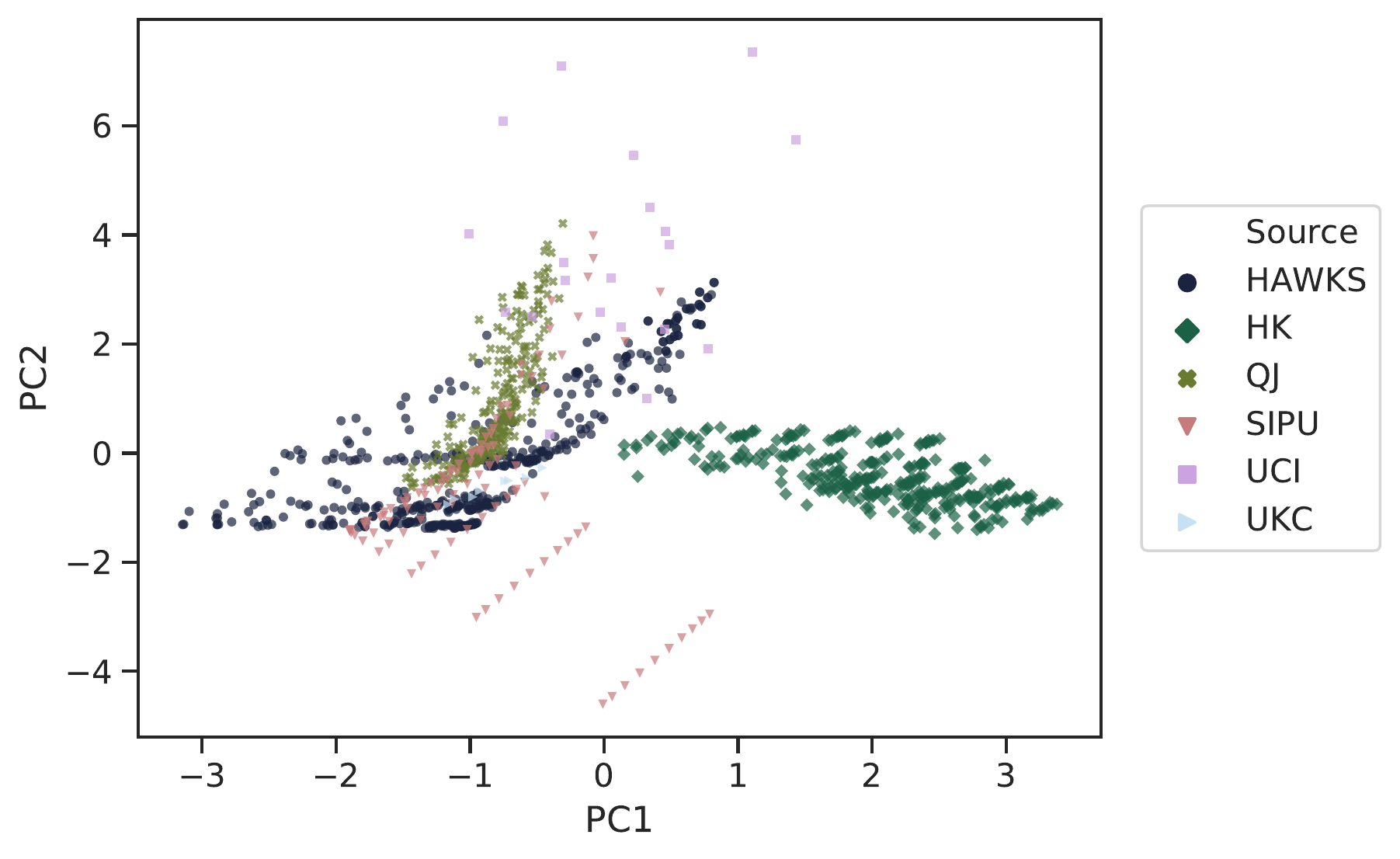}}\hfill
	\subfloat[Algorithm with highest ARI\label{fig:instance_source-alg}]{\includegraphics[width=\columnwidth]{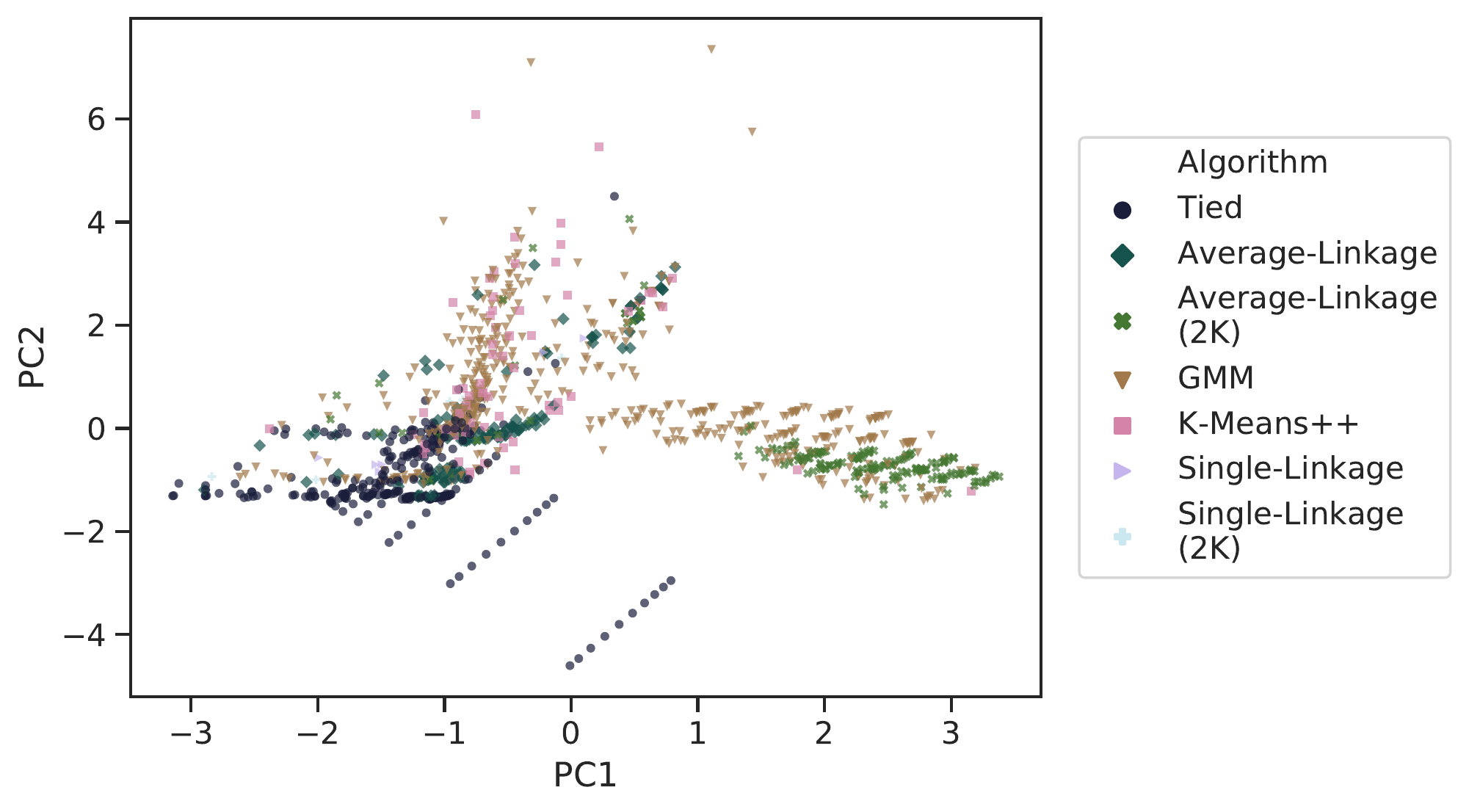}}
	\caption{Two instance spaces, with the source of the dataset (a) or clustering algorithm with the highest ARI (b) highlighted. The 7 problem features are projected down to 2D using PCA. The contribution of each problem feature to the two principal components is $\begin{bmatrix} -0.029 & 0.208 & 0.519 & 0.308 & 0.506 & -0.256 & 0.520\protect\\ 0.675 & -0.355 & -0.206 & -0.007 & -0.120 & -0.555 & 0.231 \protect\end{bmatrix}$ for the connectivity, dimensionality, average eccentricity, entropy, number of clusters, silhouette width (average), and silhouette width (standard deviation) respectively.}
	\label{fig:instance_space}
\end{figure}

To consider this aspect, Fig.~\ref{fig:instance_source-alg} shows the instance space again with the clustering algorithm achieving the highest ARI for a given dataset highlighted. The `tied' category is used when at least two algorithms were able to achieve the same ARI, which in every case was 1 and thus the dataset was trivial to cluster. This visualization permits identification of potential footprints, though the full information is tabulated in Table~\ref{tab:results_best-alg} for completeness. It is evident that there are some areas of the space that correlate with a high performance of a particular algorithm, and this happens for the \emph{HK} benchmark suite in particular where the highly ellipsoidal clusters consistently favour GMM or average-link with the higher setting of $K$.

In the more central part of the instance space, however, such footprints are unclear, indicating a complex performance landscape. This could be a consequence of the projection (which may lose too much information to fully distinguish the datasets), or may point to the role of other features that have not yet been captured here, but distinctly impact on performance. Notably, variation in the identity of the best performing algorithm is the most pronounced for the HAWKS benchmark (see Table~\ref{tab:results_best-alg}), which is the only collection of datasets where every algorithm was best for a particular dataset. As this was achieved through just a few parameter settings, this is encouraging for the potential of HAWKS to generate diverse datasets.

\input{tables/table1.tex}

\begin{figure}
	\centering
	\includegraphics[width=\columnwidth]{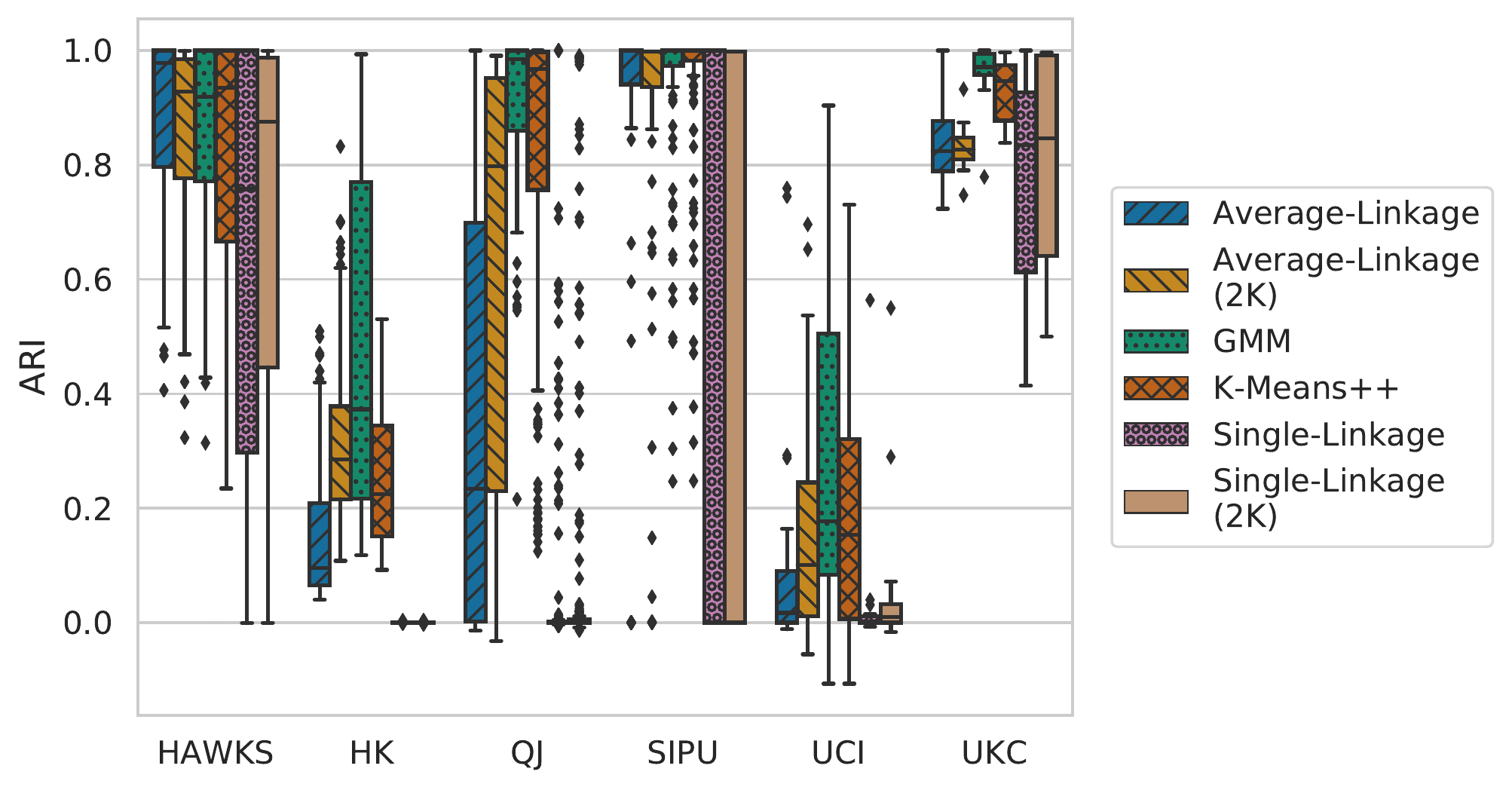}
	\caption{Clustering performance for each algorithm for each set of datasets.}
	\label{fig:suite_bplot} 
\end{figure}

To further examine the performance variation of the clustering algorithms across these datasets, Fig.~\ref{fig:suite_bplot} shows boxplots for the aggregated performance of each algorithm for each group of datasets. Here, larger boxplots indicate a greater variety of performance for that clustering algorithm, which is preferred. 
For HAWKS, the boxplots indicate that its datasets elicit a broad range of performance across all algorithms, though the high median ARIs for all algorithms indicate that in general the datasets were not that difficult. As we discourage overlap and half of the datasets were optimized to have a high silhouette width ($\silhtarget=0.9$), this is somewhat expected and could be addressed by revised parameter choices.

The near-perfect average performance for all algorithms but single-linkage on the \emph{SIPU} datasets indicate their relative simplicity. Similarly, the poor average performance across the \emph{UCI} datasets is consistent with the low connectivity and silhouette width values previously observed in the instance space, and points to weakly defined structure of the ground truth. The \emph{HK} datasets show a reasonable diversity of performance, and the significantly lower mean ARI indicates that these are much harder datasets. This may be in part due to the much higher eccentricity of these clusters, but also potentially due to a greater variance in the silhouette width (Fig.~S-\zref{supp-fig:benchmark_sw-std}), indicating that many data points on the edges of clusters are closer to the points in other clusters than to points in their own. The high performance of all algorithms on the \emph{UKC} datasets indicates that the clusters in this real-world dataset are generally well-defined. This is consistent with the instance space which provided no evidence of unusual complexity.

To include considerations of significance into our analysis, we follow the approach outlined by Dem{\v{s}}ar~\cite{demvsar2006statistical} to compare multiple methods across multiple datasets. In brief, a Friedman test is used to rank each competing algorithm for each dataset, where the null hypothesis is that all algorithms have equal ranks. As rejection indicates at least one algorithm is significantly different, the (two-tailed) Nemenyi test~\cite{nemenyi1962distribution} is used as a post-hoc test to ascertain \emph{which} algorithm is different by calculating the critical difference (CD), which is the minimum that two average ranks must differ by to be significantly ($p<0.05$) different. We illustrate the results using CD diagrams, which show the average rank of each algorithm across all datasets, with solid lines connecting algorithms whose difference in rank is less than the CD. Well-ordered rankings of algorithms indicate a lack of variance in performance (as an algorithm is consistently bad or good), whereas if all algorithms had an average rank of 3.5 (and thus clustered in the middle of the CD diagram), this would show that the datasets have an equal spread of difficulty for these clustering algorithms.

\begin{figure}
	\centering
	\subfloat[HAWKS\label{fig:cd_hawks}]{\includegraphics[width=\columnwidth]{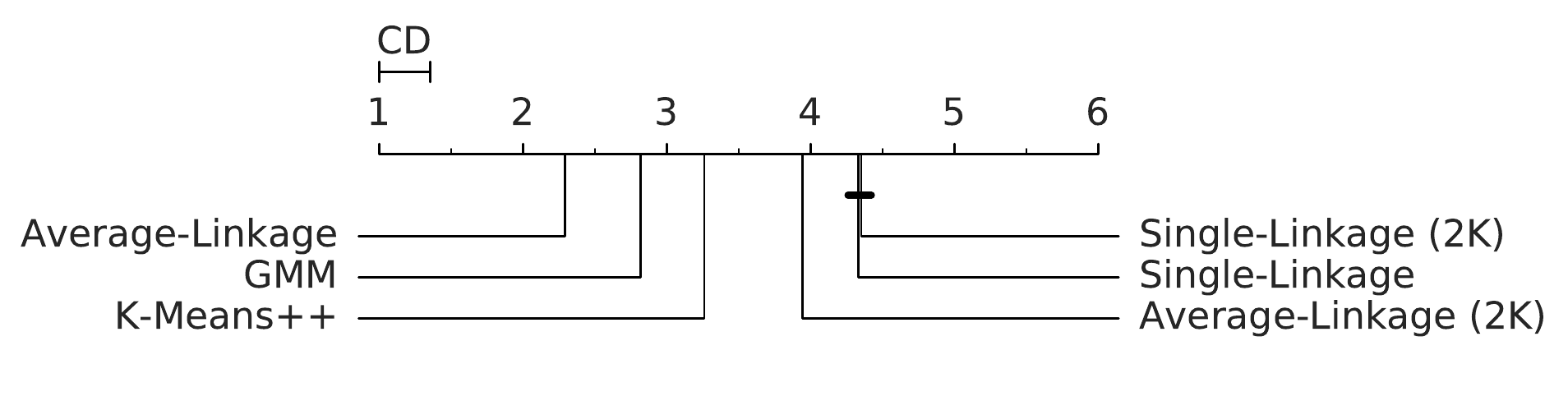}}\hfill
	\subfloat[\emph{HK}\label{fig:cd_hk}]{\includegraphics[width=\columnwidth]{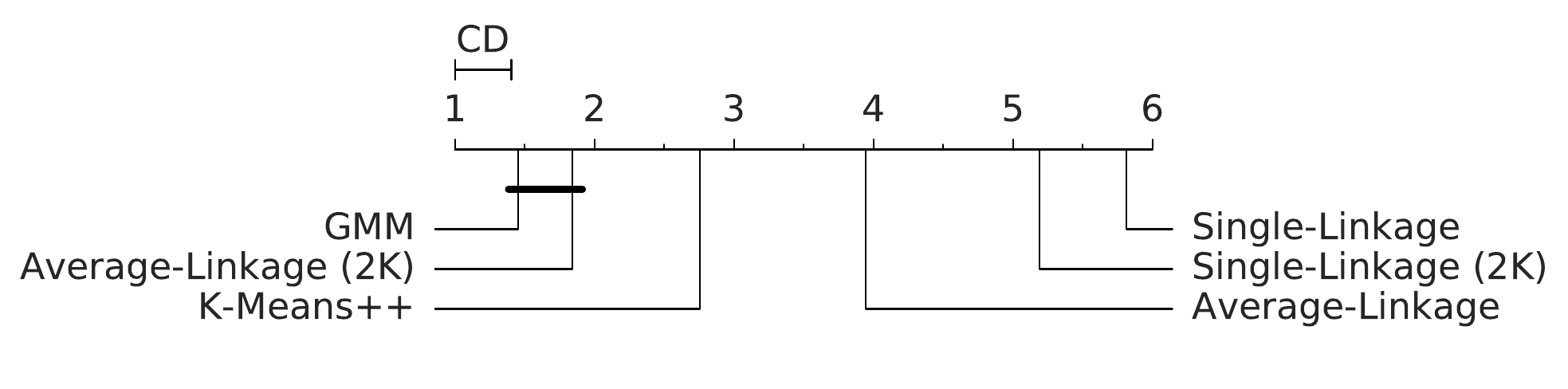}}
	\caption{Critical difference (CD) diagrams, showing the mean rank (in terms of ARI) for the HAWKS and \emph{HK} datasets, for each algorithm. Algorithms connected by solid lines are not significantly different according to a two-tailed Nemenyi test. CD diagrams for the other dataset sources can be found in Fig.~S-\zref{supp-fig:cd_extras}.}
	\label{fig:cd}
  \end{figure}

We show the CD diagrams for the HAWKS and \emph{HK} generators in Fig.~\ref{fig:cd} (as these two generators showed the greatest diversity in the boxplots)\footnote{The remaining CD diagrams can be found in Section~\zref{supp-ssec:cd_diagrams} of the supplementary material.}. As evident from the CD diagram, the ranks of the algorithms are more similar for HAWKS, whereas there is a clearer superiority for a subset of algorithms with the \emph{HK} datasets. The best-performing algorithm for HAWKS was average-linkage, which highlights the variety of cluster structures that can be generated (as the representation uses purely Gaussian clusters, one might have expected that GMM would on average perform the best). The eccentricity of the generated clusters is reflected in the higher average rank of GMM over \kmeanplus. Furthermore, the poor performance of single-linkage (using both $K$ and $2K$) indicates that there is (on average) insufficient cluster separation to avoid the `chaining' effect~\cite{hubert1974approximate}, where there is at least one pair of points in different clusters that are closer than a pair of points within a cluster, thereby forming a `bridge' between clusters. 

As shown in Table~\ref{tab:results_best-alg}, all of the generators struggle to produce datasets that single-linkage is uniquely better at, suggesting a possible lower utility of this algorithm, in general. To further investigate this, and to create a fully comprehensive benchmark, it would be of interest to more directly generate datasets that favour single-linkage. In contrast to existing generators, the \emph{Versus} mode in HAWKS facilitates this direct generation, and the results from this mode are discussed in the next section.

% ----------------- Performance ------------------ %
\subsection{Generating datasets that challenge specific algorithms (Versus mode)} \label{ssec:versus}

In this section, we present the results of running HAWKS in \emph{Versus} mode, such that we evolve datasets to directly maximize the performance difference between pairs of algorithms. As seen in Section~\ref{ssec:benchmark}, most current generators have some bias towards a particular algorithm, and no generator (except HAWKS, though not consistently so) was able to produce datasets that were uniquely suited to single-linkage.

First, we need to look at the broad capability of the \emph{Versus} mode in producing a performance differential in the various head-to-heads. In Fig.~\ref{fig:versus_grid}, we can see a grid of plots showing the performance (ARI) of every algorithm against every other. Each line (in the off-diagonal plots) represents the best dataset from a single run, showing the ARI for `winning' (left) and `losing' (right) algorithms. Here, the angle of the lines indicate the magnitude of the performance difference, and the spread shows the consistency of HAWKS across runs. The plots on the diagonal aggregate the performance for that particular algorithm, e.g.~the bottom-right plot indicates that HAWKS was able to produce datasets that single-linkage performed both very well and very poorly on, dependent on whether it was designated as the winner or loser of the head-to-head.

\begin{figure}[th]
	\centering
	\includegraphics[width=\columnwidth]{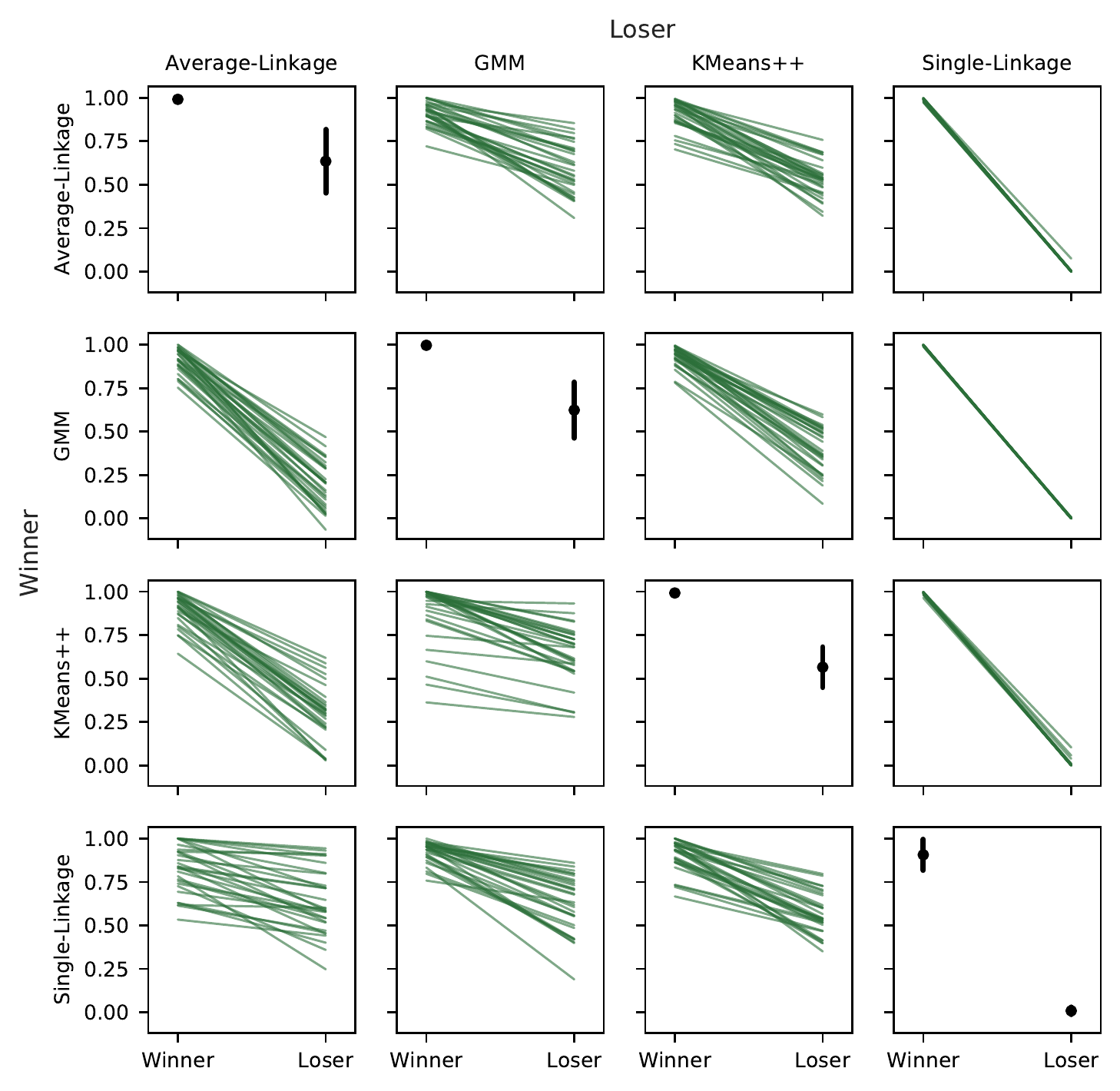}
	\caption{Performance (ARI) for each algorithm as the winner (row) and loser (column). Each line represents the best individual from a single run, connecting the ARI obtained for \winneralg~and \loseralg. On the diagonal, the average (and standard deviation) performance is aggregated for that individual as the winner and loser, indicating the overall capability of HAWKS to produce datasets that are simple and difficult for that algorithm.}
	\label{fig:versus_grid}
\end{figure}

% \begin{figure*}[th]
% 	\centering
% 	\subfloat[\label{fig:versus_grid_1}]{\includegraphics[width=0.49\textwidth]{imgs/grid_parallel_plot.pdf}}\hfill
% 	\subfloat[\label{fig:versus_grid_2}]{\includegraphics[width=0.49\textwidth]{imgs/grid_parallel_plot_newseeds.pdf}}
% 	\caption{Performance (ARI) for each algorithm as the winner (row) and loser (column). Each line represents the best individual from a single run, connecting the ARI obtained for \winneralg~and \loseralg. On the diagonal, the average (and standard deviation) performance is aggregated for that individual as the winner and loser, indicating the overall capability of HAWKS to produce datasets that are simple and difficult for that algorithm. The stochastic algorithms (GMM and \kmeanplus) are re-run with a different initialization (as indicated by the solid lines, with the original results as dashed lines) in \ref{sub@fig:versus_grid_2} to further measure robustness.}
% 	\label{fig:versus_grid}
% \end{figure*}

There are some clear differences between certain pairings of algorithms. As hinted in Section~\ref{ssec:benchmark}, HAWKS is consistently able to generate maximal (i.e.~$\phi(\winneralg) - \phi(\loseralg) \approx 1$) performance difference when single-linkage is \loseralg, but the performance difference appears minimal for \algversus{single-linkage}{average-linkage}, indicating that any weaknesses specific to average-link are not weaknesses that single-link can exploit. The average ARI for average-linkage, GMM and \kmeanplus~when set as the `loser' (\loseralg) indicates a difficulty in consistently generating datasets that these algorithms perform very poorly on. This is not unexpected, given the generating model used. As long as clusters are not fully overlapping (which we discourage with our $\overlap$ constraint) we expect these algorithms to identify some elements of the clusters, making an ARI close to 0 unlikely.

To evaluate the influence of algorithm initialization, the stochastic algorithms (GMM and \kmeanplus) are re-run with a different initialization (Fig.~S-\zref{supp-fig:versus_grid_2}). While \kmeanplus~was largely unchanged, there was an average ARI increase (as \loseralg) of 0.13 for GMM, highlighting that in some cases the lower performance was due to a poor initialization. The average ARI did, however, slightly decrease (as \winneralg), highlighting the potential disadvantage of these stochastic algorithms over linkage-based algorithms as they converge to local minima. This suggests another use case for HAWKS. When given an infinite budget of initializations and picking the best, we expect these algorithms to perform better. The robustness of the initialization method can be assessed, however, by investigating how often the algorithm is still able to achieve good performance with a limited budget. 

In order to investigate the ability of the \emph{Versus} mode to grant algorithmic insight, we need to inspect the structures HAWKS discovered for different algorithm combinations. For this, it is important to identify where HAWKS struggles to generate datasets that favour one algorithm over another. We can then try to establish whether this is due to the superiority of one algorithm over another, or the inability of HAWKS to generate structures with properties that \emph{would} differentiate them. For brevity, the following sections discuss some interesting examples for a subset of the scenarios (shown in Fig.~\ref{fig:versus_all}). Further examples of the algorithm pairings can be found in the supplementary material (Section~\zref{supp-sec:versus}).

\begin{figure}
	\centering
	\subfloat[\algversus{GMM}{single-linkage}\label{fig:versus_gmm-slink}]{\includegraphics[width=\columnwidth]{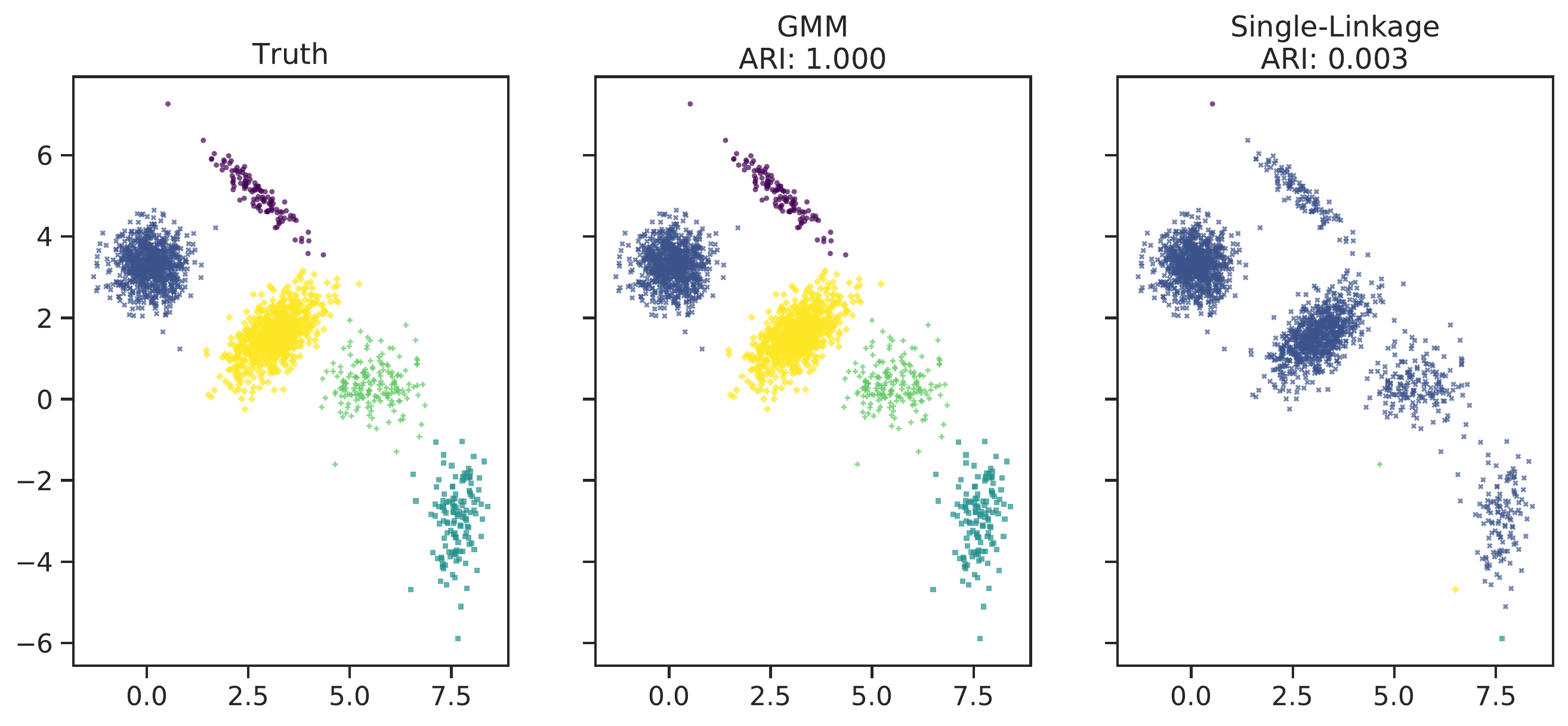}}\hfill
	\subfloat[\algversus{Single-linkage}{GMM}\label{fig:versus_slink-gmm}]{\includegraphics[width=\columnwidth]{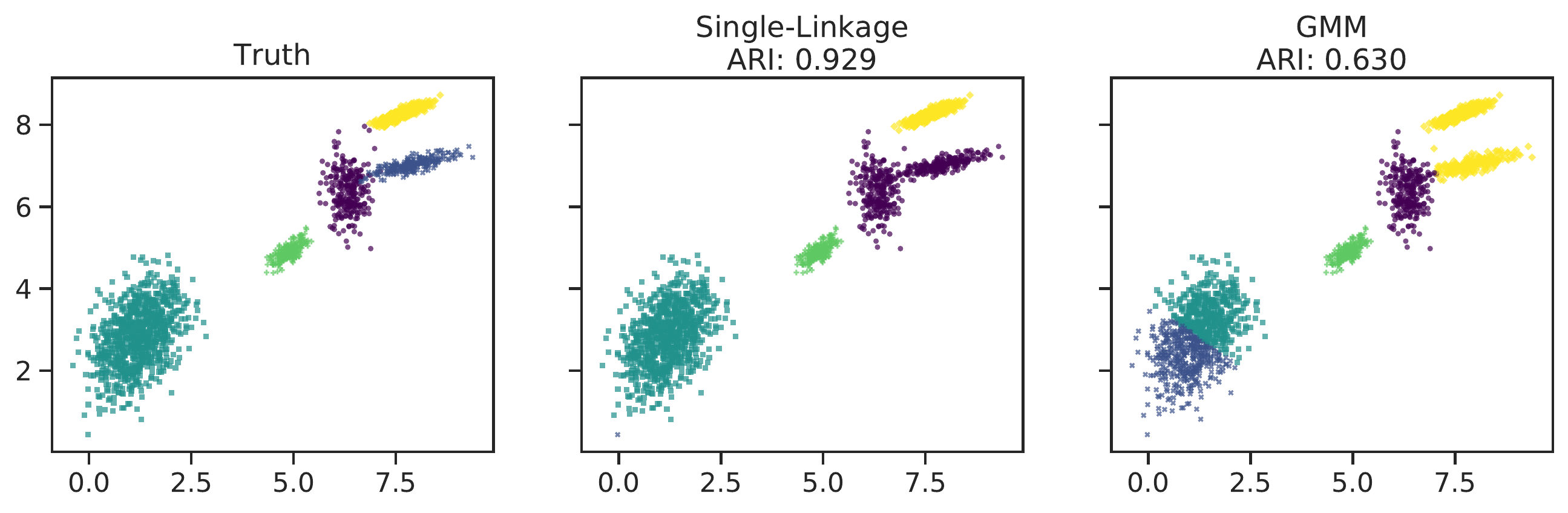}}\hfill
	\subfloat[\algversus{GMM}{\kmeanplus}\label{fig:versus_gmm-kmeans}]{\includegraphics[width=\columnwidth]{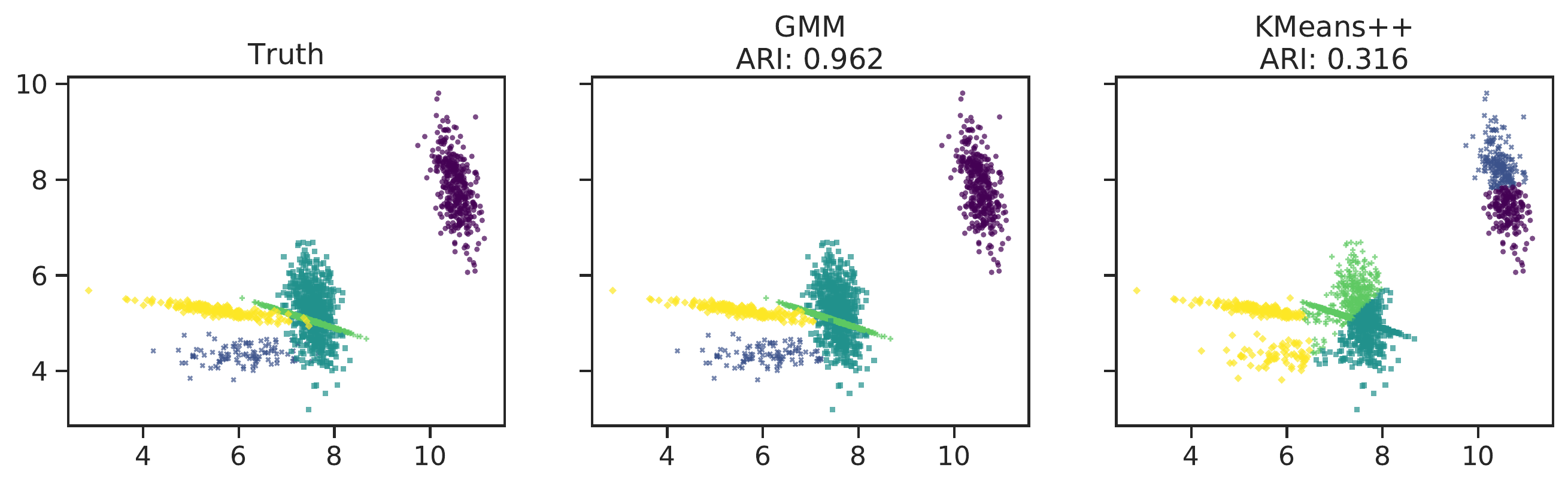}}\hfill
	\subfloat[\algversus{\kmeanplus}{GMM}\label{fig:versus_kmeans-gmm}]{\includegraphics[width=\columnwidth]{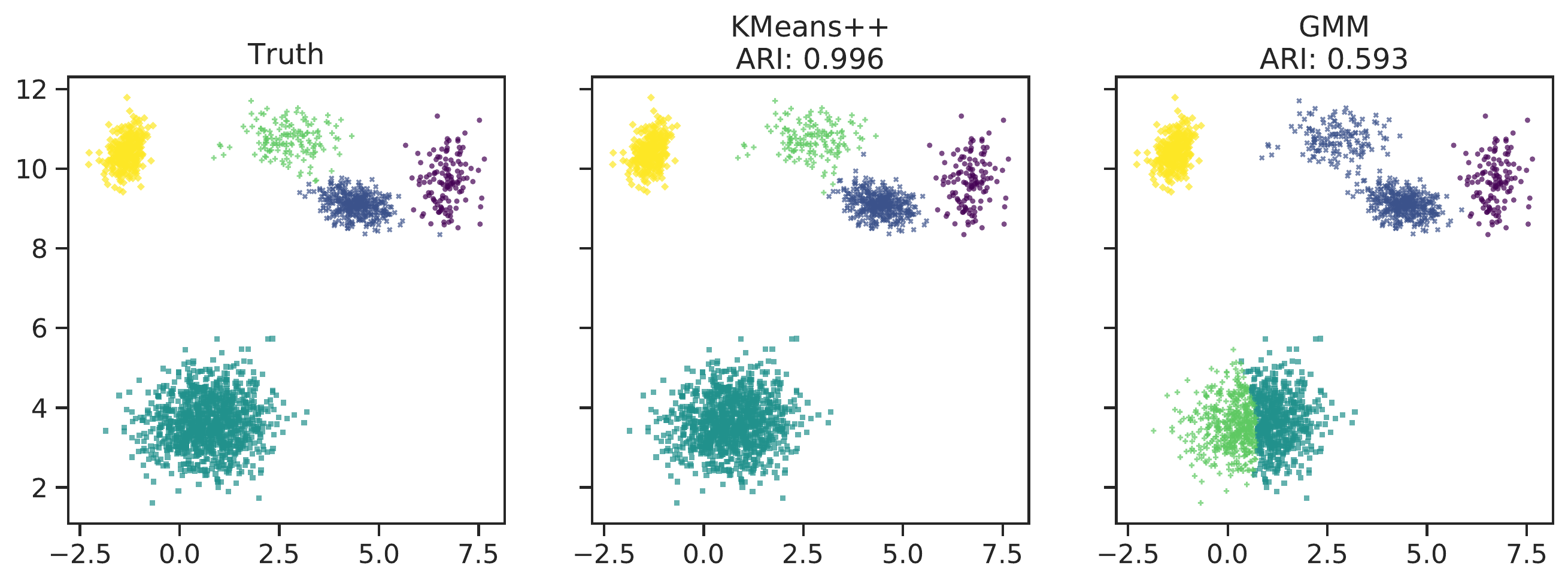}}
	\caption{Examples of datasets for the listed head-to-heads. Each figure shows the ground truth (left column), and the cluster assignment for each algorithm with the associated ARI (middle and right columns).}
	\label{fig:versus_all}
\end{figure}

\subsubsection{\algversus{GMM}{single-linkage}}
Owing to the Gaussian representation that HAWKS uses, and the known issues of single-linkage, this scenario is expected to provide a large performance difference. Fig.~\ref{fig:versus_gmm-slink} shows that HAWKS achieves this performance difference by exploiting the aforementioned `chaining' effect of single-linkage~\cite{hubert1974approximate}. Discovering this requires iterative movement of the clusters in order for the data points to be close enough to induce this effect, supporting the utility of our mutation operators' nuanced ability to adjust the location of clusters.

\subsubsection{\algversus{Single-linkage}{GMM}}
The reverse scenario should be much harder for HAWKS, as GMM is largely insensitive to eccentricity and naturally fits our cluster representation. As exemplified in Fig.~\ref{fig:versus_slink-gmm} (and observed in the other datasets produced), HAWKS tends to place large clusters far away from several smaller compact clusters in order to increase the chance of a poor initialization from GMM, exploiting the stochasticity of this method.\footnote{Each fitness evaluation uses a different initialization for GMM and \kmeanplus, otherwise HAWKS tacitly exploits this knowledge by moving the clusters into the static initial centroid locations. This will maximize performance difference, but represents a technological artefact rather than a generic property.}

\subsubsection{\algversus{GMM}{\kmeanplus}}
In Fig.~\ref{fig:versus_gmm-kmeans} we can see that a key exploit, as discovered by HAWKS, is the inability of \kmeanplus~to handle eccentric clusters. 
%Although the overlap for this dataset is low ($4\%$), it is of limited general use and likely unrealistic. Despite this, the properties of the generated datasets are of interest in identifying relative strengths and weaknesses of algorithms 
The example highlights significant differences between the use of mixture models and an algorithm relying on assignment to the closest centroid. A clear performance differential is found on this dataset, despite basic similarities in the inductive biases of the two algorithms.

\subsubsection{\algversus{\kmeanplus}{GMM}}
As both algorithms are well-suited for compact clusters, eccentricity is not a characteristic that can be used for differentiating performance in this scenario. Here, HAWKS exploits GMM's previously-mentioned weakness (of sub-dividing a single large cluster) during its initialization stage. \kmeanplus~is less sensitive to this problem due to its improved initialization routine. This shows the utility of HAWKS in identifying relative strengths and weaknesses of specific algorithms, which could aid in algorithmic development (e.g.\ when empirically comparing initialization schemes).

%\subsection{Instance space visualization}
The datasets in the \emph{Versus} mode are optimized towards a performance differential between the algorithms, rather than towards a cluster structure specified by the cluster validity index (as done in the \emph{Index} mode). It is therefore of interest to investigate how these datasets compare in terms of problem feature diversity. For this, we add the 30 datasets from each of the 12 head-to-heads to the instance space created in Section~\ref{ssec:benchmark} (using the same principal components).

In Fig.~\ref{fig:versus_instance}, we have highlighted the datasets from the \emph{Index} and \emph{Versus} modes (with the other datasets in grey for reference). Clearly, we are able to generate datasets that are notably different in terms of their problem features (and thus properties). In particular, there are many datasets in the region where previously only \emph{HK} produced datasets, further highlighting the flexibility of our generating mechanism in covering additional regions, as objective function, clustering algorithms, and other parameters are varied. The problem feature values are tabulated for all dataset sources, including those from \emph{Index} and \emph{Versus} modes, in Table~S-\zref{supp-tab:problem_features}.

\begin{figure}
	\centering
	\includegraphics[width=\columnwidth]{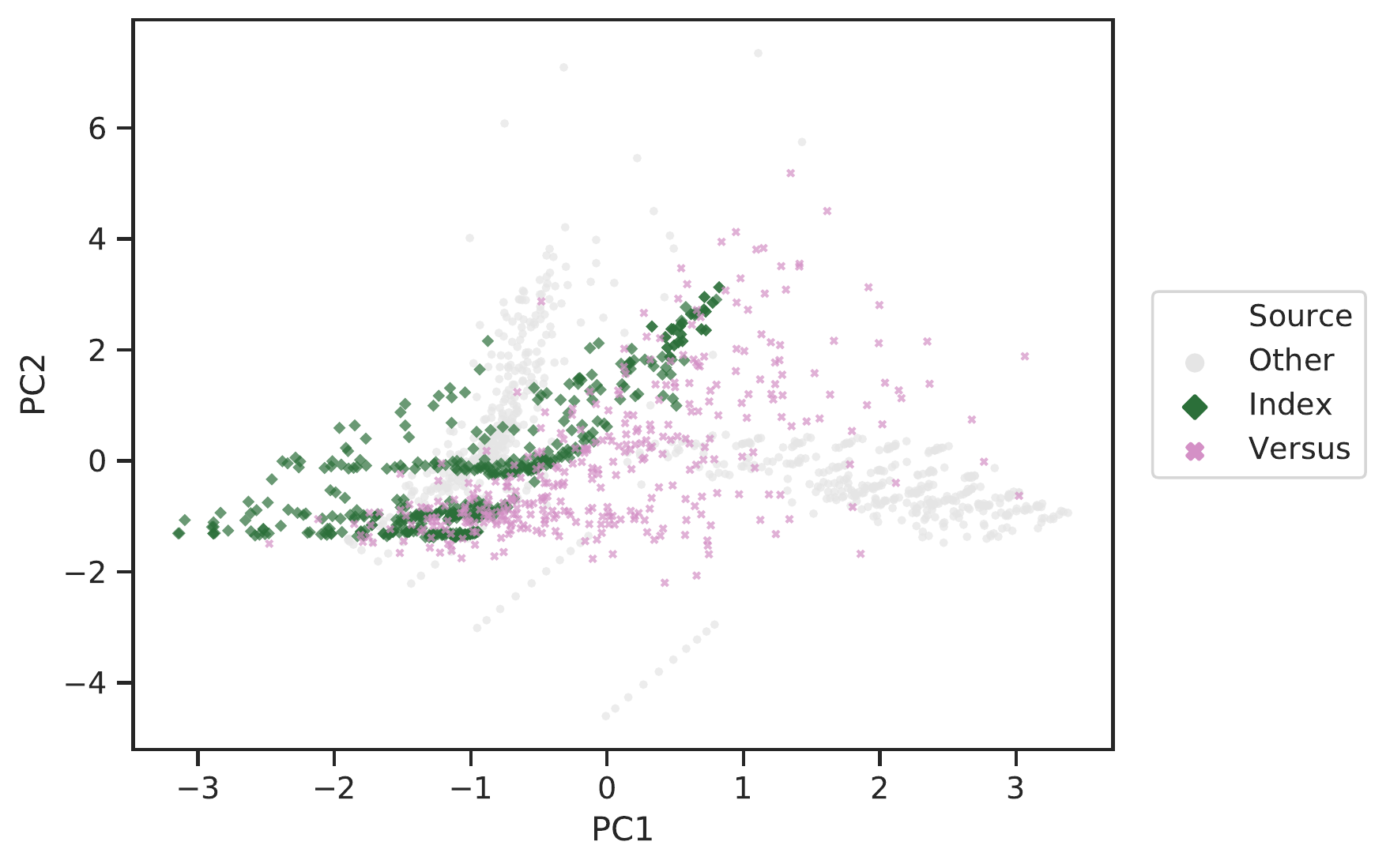}
    \caption{Instance space created in Section~\ref{ssec:benchmark} with the datasets produced by the two modes of HAWKS highlighted (the `Other' points are the other datasets).}
	\label{fig:versus_instance}
\end{figure}

% ---------------------------------------------- %
% ------------------ Conclusion ---------------- %
% ---------------------------------------------- %
\section{Conclusion} \label{sec:conc}

Clustering is a vital tool for pattern discovery, but it is often unclear which clustering algorithm is the most appropriate for a given dataset. An optimal choice requires an accurate understanding of the data properties as well as the strengths and weaknesses of candidate algorithms. Both types of information are difficult to come by in typical real-world settings.

Synthetic benchmark datasets play an important role in improving our understanding of the former, i.e.\ to examine the specific strengths and capabilities of a given clustering method. Their specific advantage is the availability of a known generating model, which allows researchers to relate aspects of the true cluster structure to algorithm performance. Unfortunately, available synthetic benchmarks for cluster analysis cover a limited variety of structural aspects, and there are no existing generators that have been designed with a wider flexibility in mind.

Our framework HAWKS employs the power of an EA to better meet the challenges highlighted above, and to generate more diverse collections of benchmarks, in particular. When compared to existing clustering benchmarks, HAWKS is found to generate datasets exhibiting more feature diversity and eliciting more variation in algorithm performance. The modular nature of HAWKS provides much opportunity for extension; additional distributions can be added to increase dataset diversity, and extension to temporal/streaming data would greatly benefit a currently underserved application. Further work to improve the ability of the instance space to distinguish algorithmic footprints, by further enriching the set of problem features and improving the projection methodology, is also needed.

Finally, HAWKS can be modified to directly generate datasets that are either simple or difficult for a given algorithm, facilitating a deeper understanding of existing algorithms and potentially informing algorithm development. This provides a new avenue for ``controlled experimentation'' with clustering algorithms, which does not rely on the use of over-simplified toy datasets. Investigations into the use of this \emph{Versus} mode in conjunction with more complex clustering algorithms, could further test HAWKS' capabilities and may provide novel insights into the inductive biases and performance of these algorithms.

\bibliographystyle{IEEEtran}
\bibliography{IEEEabrv,refs}

% \begin{comment}
\vspace{-1cm}
\begin{IEEEbiography}[{\includegraphics[width=1in,height=1.25in,clip,keepaspectratio]{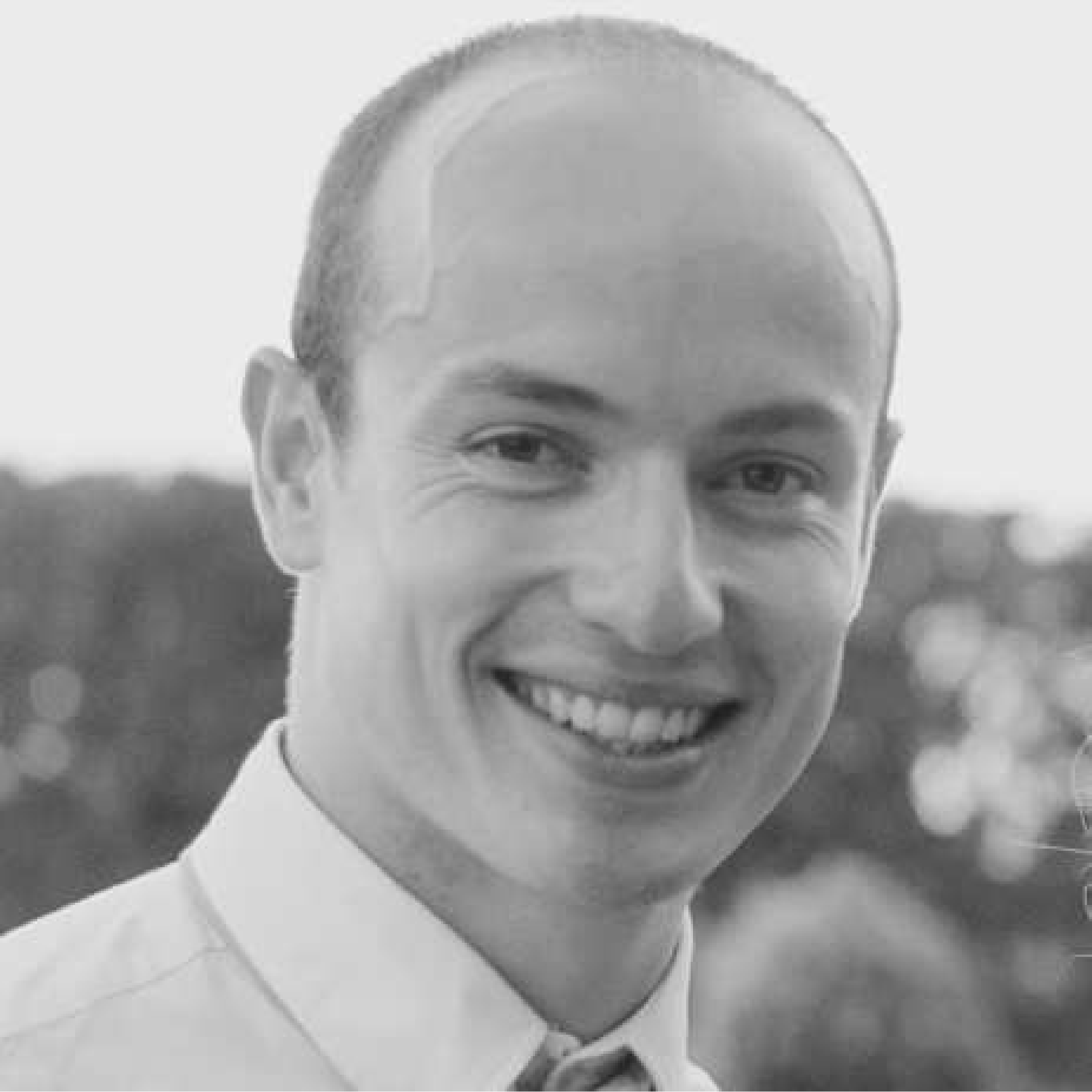}}]{Cameron Shand}
(memberreceived the M.Eng. degree in biochemical engineering from UCL (London, UK) in 2015, and the Ph.D. degree in computer science from the University of Manchester (Manchester, UK) in 2020.

Since 2020, he is currently a Research Fellow in Disease Progression Modelling and Machine Learning for Clinical Trials with the Centre for Medical Image Computing at UCL. His research interests include the development of unsupervised machine learning techniques and their clinical application, and the role of synthetic data in understanding and improving clustering methods.
\end{IEEEbiography}

\vspace{-1cm}
\begin{IEEEbiography}[{\includegraphics[width=1in,height=1.25in,clip,keepaspectratio]{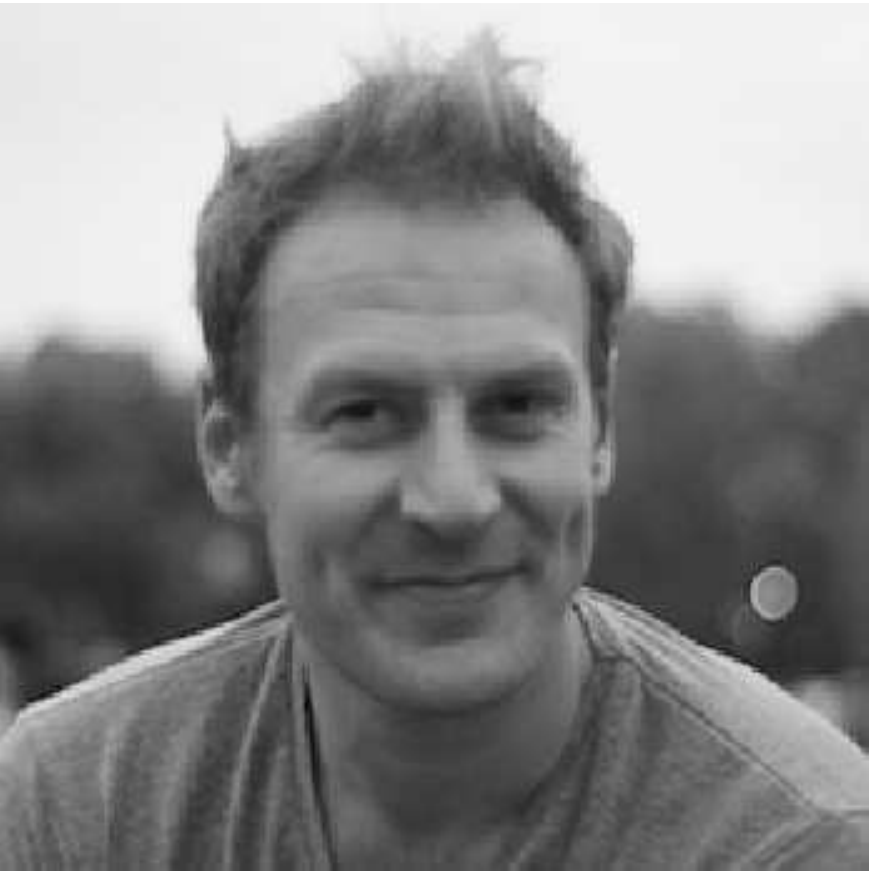}}]{Richard Allmendinger} (Member, IEEE) received a Diplom in business engineering from Karlsruhe Institute of Technology, and the Ph.D. degree in computer science from the University of Manchester (UoM), UK.

He is currently a Senior Lecturer (Associate Professor) in Decision Sciences at UoM and an Alan Turing Fellow, having worked previously as a postdoc at the biochemical engineering department, University College London. His research interests include multi-objective, dynamic, safe, and expensive optimization and machine learning. He is a Vice-Chair of the IEEE CIS Bioinformatics and Bioengineering (BB) Technical Committee, Co-Founder of the IEEE CIS Task Force on Optimization Methods in BB, and on the Editorial Board of several international journals.
\end{IEEEbiography}

\vspace{-1cm}
\begin{IEEEbiography}[{\includegraphics[width=1in,height=1.25in,clip,keepaspectratio]{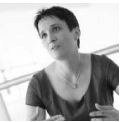}}]{Julia Handl}
received the B.Sc. (Hons.) degree in computer science from Monash University, Melbourne, VIC, Australia, in 2001, the M.Sc. degree in computer science from the University of Erlangen-Nuremberg, Erlangen, Germany, in 2003, and the Ph.D. degree in bioinformatics from the University of Manchester, Manchester, U.K., in 2006.

From 2007 to 2011, she held an MRC Special Training Fellowship with the University of Manchester, and is currently an Alan Turing Fellow and Professor in the Management Sciences Group, Alliance Manchester Business School, Manchester. Her research interests include theoretical and empirical work related to the development and use of machine learning and optimization approaches in a variety of application areas.
\end{IEEEbiography}

\vspace{-1cm}
\begin{IEEEbiography}[{\includegraphics[width=1in,height=1.25in,clip,keepaspectratio]{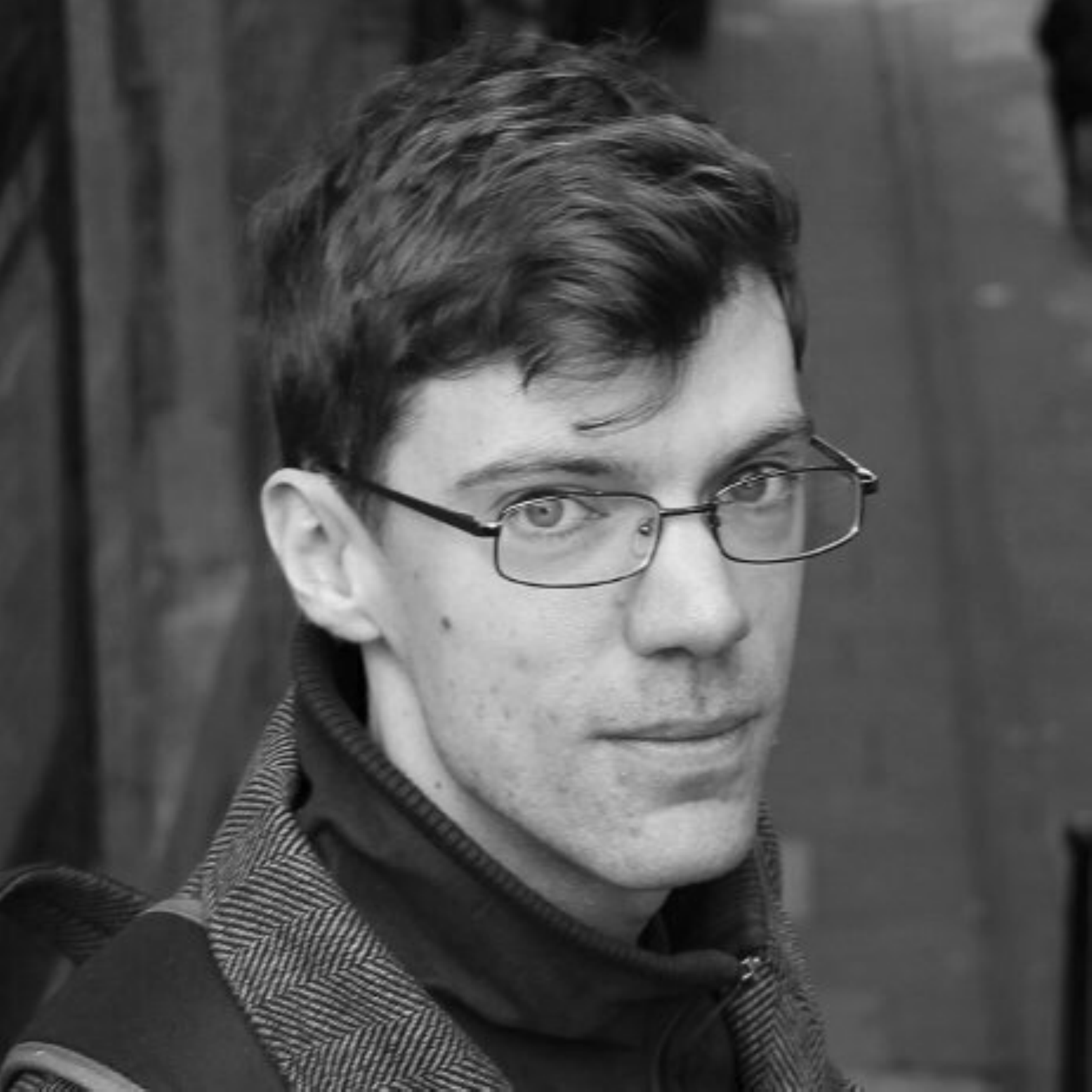}}]{Andrew Webb} received the B.Sc. (Hons.) degree in computer science from the University of Manchester (Manchester, UK) in 2010, and the Ph.D. in computer science from the University of Manchester in 2016.

From 2017 to 2019 he was a Research Associate at the University of Manchester, with research interests in deep learning and ensemble methods in machine learning. He is currently a Machine Learning Research and Development Engineer at vTime Limited, Liverpool, UK.
\end{IEEEbiography}

\vspace{-1cm}
\begin{IEEEbiographynophoto}{John Keane} holds the MG Singh Chair of Data Engineering in the Department of Computer Science, School of Engineering, University of Manchester and is an Honorary Professor of Data Analytics at the Alliance Manchester Business School. He has previous commercial experience with Phillips, Fujitsu, and the TSB Bank. His research focuses on data analytics and decision science.
\end{IEEEbiographynophoto}
% \end{comment}

\end{document}

% \linenumbers

\title{HAWKS: Evolving Challenging Benchmark Sets for Cluster Analysis\\-- SUPPLEMENTARY MATERIAL --}

% Authors

% \author{
% 	Cameron~Shand$^{\textsuperscript{\orcidicon{0000-0002-1299-890X}}}$\,,~\IEEEmembership{??,~IEEE,} %
% 	Richard Allmendinger$^{\textsuperscript{\orcidicon{0000-0003-1236-3143}}}$\,,~\IEEEmembership{Member,~IEEE,} %
% 	Julia~Handl$^{\textsuperscript{\orcidicon{0000-0002-4338-1806}}}$\,,~\IEEEmembership{??,~IEEE,} %
% 	Andrew~Webb$^{\textsuperscript{\orcidicon{??}}}$\,,~\IEEEmembership{??,~IEEE,} %
% 	and~John~Keane$^{\textsuperscript{\orcidicon{0000-0001-9022-4339}}}$\,,~\IEEEmembership{??,~IEEE}%
% }

\author{
Cameron~Shand, Richard~Allmendinger,~\IEEEmembership{Member,~IEEE,} Julia~Handl, Andrew~Webb, and~John~Keane
}

\ifCLASSOPTIONpeerreview
\markboth{IEEE Transactions on Evolutionary Computation}%
{Anonymized}
\else
\markboth{IEEE Transactions on Evolutionary Computation}%
{Authors}

\maketitle

% ---------------------------------------------- %
% --------------- HAWKS Details ---------------- %
% ---------------------------------------------- %
\section{HAWKS Details} \label{sec:hawks_details}

\subsection{HAWKS Overview} \label{ssec:hawks_summary_fig}
Fig.~\ref{fig:hawks_summary} provides an overview of the inputs (left) to the different conceptual levels of HAWKS, and the interaction between them. A simple example is shown on the right.

\begin{figure*}
	\centering
	\includegraphics[width=0.95\textwidth]{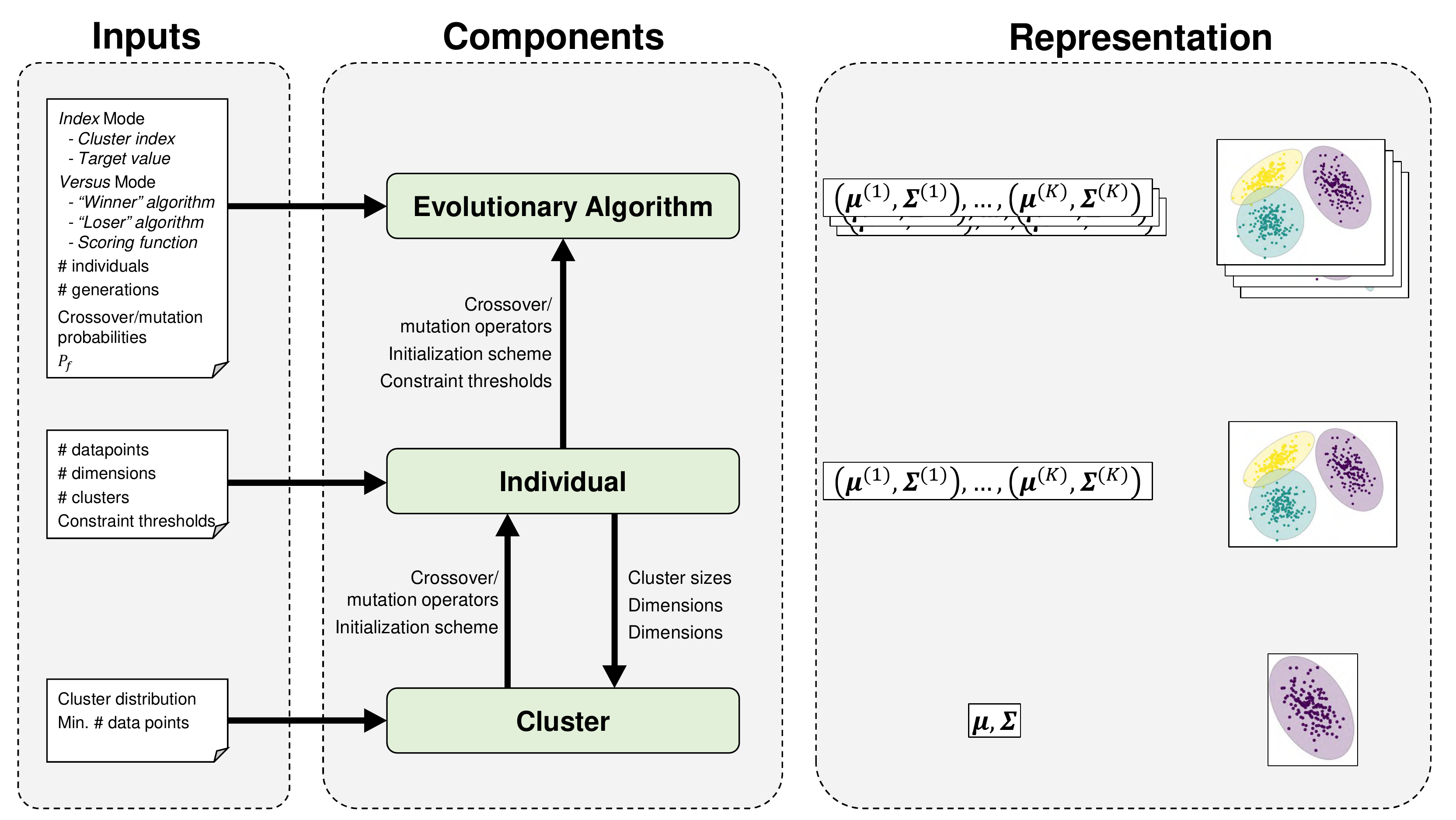}
	\caption{Overview of HAWKS. The input parameters (left) affect different components of HAWKS, which themselves and form the hierarchy of clusters, which constitute an individual dataset, which in turn form the population that the EA interacts with. A simple example and representation is shown on the right.}
	\label{fig:hawks_summary}
\end{figure*}

\subsection{Generating random cluster sizes to a pre-defined limit} \label{ssec:cluster_sizes}

To generate randomly sized clusters that sum to a given value with a minimum size, the following approach is used:
\begin{equation*}
    \mathbf{w} = [w_{1},\ldots,w_{K}]
\end{equation*}
where $K$ is the number of clusters, $w_{i}\sim \text{Dir}(\alpha)$, and $\alpha = (1,1,\ldots,1)$. For a given cluster $i$, the size is:
\begin{equation*}
    |C_{i}| = \mathcal{C}_\text{min} + w_{i}(N - (K \times \mathcal{C}_\text{min}))
\end{equation*}
where $\mathcal{C}_\text{min}$ is the minimum size a cluster can be.

This ensures a uniform distribution of cluster sizes \emph{after} scaling such that they sum to $N$, which is not guaranteed when simply sampling random numbers and scaling these to $N$ (as this no longer guarantees a uniform distribution).

\subsection{Scaling the covariance matrix} \label{ssec:scaling_matrix}

We provide further details about the covariance mutation operator described in Section~\zref{main-ssec:hawks_variation}. To mutate the covariance matrix in HAWKS, two separate matrices are generated for the perturbation. The rotation matrix randomly rotates the cluster, whereas the scaling matrix modifies the magnitude of the principal semi-axes, modifying the area of the space that is covered by this distribution.

To avoid issues of converging towards either spherical or highly eccentric clusters upon multiple applications of the mutation operator, we ensure that the determinant of the covariance matrix remains the same after applying the scaling matrix. For this, let $x_{i}$ be the elements of a vector sampled from a Dirichlet distribution. Then
\begin{equation*}
    \begin{gathered}
    \sum_{i}^{D} x_{i} = 1\\
    \Rightarrow \sum_{i}^{D} \left( x_{i} - \frac{1}{D} \right) = 0\\
    \Rightarrow \exp \left( \sum_{i}^{D} \left( x_{i} - \frac{1}{D} \right) \right) = 1\\
    \Rightarrow \prod_{i}^{D} \exp \left( x_{i} - \frac{1}{D} \right) = 1.
    \end{gathered}
\end{equation*}

The determinant of a diagonal matrix is the product of the values on the diagonal, thus the scaling matrix ($\exp (x_{i} - \frac{1}{D})$) has determinant 1.

\subsection{Problem features} \label{ssec:problem_features}

\subsubsection{Average eccentricity}
A high amount of cluster eccentricity can cause problems for compactness-based algorithms. To measure this, we use a similar method to its calculation as a constraint in HAWKS, with some key differences. For both the synthetic and real-world datasets used in this paper, we have the true labels. Using these, the eccentricity is calculated as before --- calculating the ratio of the maximum to minimum eigenvalues for each cluster, except that the average (rather than minimum) value is used as the problem feature. The eigenvalues are obtained via singular value decomposition, but (in contrast to the constraint) a subset of the eigenvalues that accounts for 95\% of the total sum is used. This is functionally identical to using the principal components (obtained via principal component analysis) that account for 95\% of the variance. This reduces the sensitivity to outliers and subspace clusters (which result in zero, or arbitrarily close to zero, eigenvalues).

\subsubsection{Connectivity}
The connectivity, defined by Handl and Knowles~\cite{Handl2007} (which is a modified version of what was proposed in~\cite{ding2004k}), measures the extent to which neighbouring data points are assigned to the same cluster. We normalize this measure by the number of data points to get a measure that is comparable across datasets. This provides a more nuanced picture than the overlap constraint used by HAWKS, which looks at the single nearest neighbour rather than the set of $L$ nearest neighbours, where $L$ is a parameter of the connectivity. Here, we use $L=10$ in line with previous work~\cite{Handl2007}.

\subsubsection{Dimensionality}
As different measures of distance or similarity are affected differently by high-dimensionality, this feature simply describes the number of dimensions in the data. This is our only feature that is ground-truth agnostic, and is shared with previous work on the ASP for clustering~\cite{ferrari2015clustering,soares2009analysis}.

\subsubsection{Entropy of cluster sizes}
The density or relative sizes of the clusters can affect the performance of different algorithms. For example, the ability of K-Means to discover clusters is diminished when a small subset of clusters contain the majority of data points. We calculate the entropy of cluster sizes as follows:
\begin{equation}
	H(\mathcal{C}) = - \sum_{k=1}^{K} \frac{|\cluster{k}|}{N} \log_{K} \left( \frac{|\cluster{k}|}{N} \right),
\end{equation}
where $\mathcal{C}$ is the set of $K$ clusters, and $|\cluster{k}|$ is the cardinality of cluster $C_{k}$, which is normalized by the size of the dataset ($N$). We use $\log_{K}$ to compare across datasets such that, for any $K$, a perfectly equal distribution of cluster sizes results in $H(\mathcal{C}) = 1.0$, and $H(\mathcal{C}) \rightarrow 0$ when one cluster has $N-K+1$ data points.

\subsubsection{Number of clusters}
The number of clusters can inherently affect algorithms where initialization of cluster location is key to convergence, particularly if clusters are well-separated.

\subsubsection{Silhouette width (average)}
The average silhouette width measures how similar the data points are within their own clusters, and how well-separated these are to the other clusters, indicating the potential difficulty for clustering algorithms in discovering these clusters.

\subsubsection{Silhouette width (standard deviation)}
Averaging the silhouette width can obscure the presence of a small number of very well-separated clusters, and ill-defined overlapping clusters. Thus, the standard deviation of the silhouette width (calculated across all data points) indicates whether all points are similarly separated. Higher values indicate the presence of overlapping clusters (from a different perspective than the connectivity measure), which clustering algorithms have different capabilities in handling.

% ---------------------------------------------- %
% ------------------ Mutating ------------------ %
% ---------------------------------------------- %
\section{Mutating clusters in higher dimensions} \label{sec:mutating}

Shand et al.~\cite{shand2019evolving} noted that the original mutation operator for the cluster means ($\hawksmean$) is decreasingly useful as the dimensionality increases due to stochastic movement of the mean, as there are an increasing number of directions to move away from other clusters. This results in a bias towards increasing the silhouette width as the clusters drift apart. In this section, we propose and compare several different operators that try to explicitly incorporate directionality into the random movement to avoid this bias.

Incorporating explicit directionality into the operator can directly guide whether the movement of clusters is either away from or towards existing clusters, increasing and decreasing the silhouette width respectively. For this, the operator would need to incorporate the position of at least one other cluster in the random perturbation.

The previous~\cite{shand2019evolving} mutation operator was defined as:
\begin{equation}
	\hawksmeanargnew{i} \sim \mathcal{N}(\hawksmeanarg{i}, s)
\end{equation}
where \hawksmeanargnew{i} is the new mean, \hawksmeanarg{i} is the current mean, and $s$ is the width (variance) of the multivariate normal distribution. In other words, the new mean is sampled from a normal distribution around the current mean. Next, we define each of the proposed operators. %By default, each dimension is sampled separately from a univariate distribution and has a $\frac{1}{D}$ chance of occurring (to avoid inflated mutation rates).

\subsection{Mutation Operator Descriptions}
\subsubsection{``Rails'' operator}
As a simple baseline to examine the utility of including directionality, this operator randomly selects a cluster, and the current cluster moves either away from or towards that cluster with a random weighting ($0 \leq w_{1} \leq 1$).

To select a random cluster that is different from the current one, let $n$ be a random integer from the set $\{1, \ldots, K\} \setminus \{i\}$. The mean of the current cluster, $\hawksmeanarg{i}$, can be mutated as follows:
\begin{equation} \label{eq:operator_rails}
	\hawksmeanargnew{i} = 
	\begin{cases}
		\hawksmeanarg{i} + [w_{1} (\hawksmeanarg{n} - \hawksmeanarg{i})] \quad\text{if } p \leq 0.5 \\
		\hawksmeanarg{i} - [w_{1} (\hawksmeanarg{n} - \hawksmeanarg{i})] \quad\text{if } p > 0.5.    \\
	\end{cases}
\end{equation}
where $\hawksmeanarg{n}$ is the mean of the random cluster $n$, $w_{1} \sim U(0, 1)$ is a random weight, $\hawksmeanarg{n} - \hawksmeanarg{i}$ is the difference between the cluster means, and $p \sim U(0, 1)$ is a random uniform probability to determine whether we mutate towards or away from the other cluster.

\subsubsection{PSO-inspired mutation with random directionality}
For a mutation that covers more of the space, rather than along the vectors between centroids, we take inspiration from another EA paradigm: particle swarm optimization (PSO). In the original PSO mutation operator, each particle is updated by incorporating their current position, the best position that particle has ever had, and the best position ever found by any particle. These latter two best positions are weighted by random coefficients in order to create a random movement of the particles. For further details, see \cite{eberhart1995particle,kennedy2010particle}. We use this as inspiration by viewing the clusters as particles, and try to mutate the location of the cluster based on the position of another cluster and some global representative. As our fitness is derived from a combination of all particles, the original notions of personal and global best are not applicable here, but the concept of independently weighting both a single point and an aggregated one is.

By updating the mean using a randomly weighted combination of an existing cluster and the global mean ($\hawksmeanglobal$) of all clusters, we can create a random movement of the cluster that still takes into account the position of the existing clusters. By incorporating the global mean, we can avoid generating well-separated groups of clusters that can deceive the silhouette width. To ensure that the location of the global mean is calculated in a way meaningful to the fitness, we calculate the mean across all \emph{data points} (not across cluster means), as it is the former that is used in the calculation of the silhouette width (and thus fitness). As a result, relative cluster sizes are incorporated into the mutation. The mean of the current cluster, $\hawksmeanarg{i}$, is thus mutated using the follow operator (referred to as PSO-random):
\begin{equation}
	\small
    \hawksmeanargnew{i} = 
    \begin{cases}
        \hawksmeanarg{i} + [w_{1} (\hawksmeanarg{n} - \hawksmeanarg{i}) + w_{2} (\hawksmeanglobal - \hawksmeanarg{i})] & \text{if } p \leq 0.5 \\[0.2cm]
        \hawksmeanarg{i} - [w_{1} (\hawksmeanarg{n} - \hawksmeanarg{i}) + w_{2} (\hawksmeanglobal - \hawksmeanarg{i})] & \text{if } p > 0.5. \\
	\end{cases}
\end{equation}

\subsubsection{PSO-inspired mutation with informed directionality}
By using a coin-flip to determine whether the cluster should move towards or away from existing clusters, there arises obvious scenarios where the existing clusters may be very far apart and a closer structure is desired. An unfavourable coin-flip moves this cluster further away, wasting a perturbation and thus evaluation. By using the sign of the difference between the current silhouette width of the individual and the target ($\silhall - \silhtarget$), we can move the cluster centre in the direction that will have the best chance of improving the fitness.

The mean of the current cluster, $\hawksmeanarg{i}$, can be mutated as follows:
\begin{equation}
    \small
	\hawksmeanargnew{i} = 
	\begin{cases}
		\hawksmeanarg{i} + [w_{1} (\hawksmeanarg{n} - \hawksmeanarg{i}) + w_{2} (\hawksmeanglobal - \hawksmeanarg{i})] & \text{if } s_{\text{all}} > \silhtarget \\[0.2cm]
		\hawksmeanarg{i} - [w_{1} (\hawksmeanarg{n} - \hawksmeanarg{i}) + w_{2} (\hawksmeanglobal - \hawksmeanarg{i})] & \text{if } s_{\text{all}} \leq \silhtarget.
	\end{cases}
\end{equation}

The addition of this information will improve convergence in situations where $\silhall - \silhtarget$ is large, as the fitness will be improved more quickly, but this adds in a bias towards the fitness (as opposed to the constraints), which is typically controlled through the stochastic ranking ($P_{f}$). This operator is referred to as PSO-informed.
% This is illustrated in \autoref{fig:mohawks_mut-psoinf}, where we can see that if the individual's silhouette width is higher than the target, the mean is more likely to mutate towards the other clusters.

\subsubsection{DE-inspired mutation}
Taking inspiration from yet another EA paradigm, differential evolution (DE), we view the existing clusters as individual vectors which can help in the creation of a ``donor vector'' (the new mean) from a ``target vector'' (the current mean). The classical DE mutation operator combines three existing individuals in order to create a new individual~\cite{storn1997differential,fleetwood2004introduction,leon2014investigation}. We adapt this to generate a new cluster mean from existing ones as follows:
\begin{equation}
	\hawksmeanargnew{i} = \hawksmeanarg{i} + F(\hawksmeanarg{r_{1}} - \hawksmeanarg{r_{2}})
\end{equation}
where $i$, $r_{1}$, and $r_{2}$ are distinct indices of a cluster, and $F$ is a constant factor in the range $[0,2]$. Unlike with the previous operators, the randomness occurs only in terms of which other means are selected, such that the movement vector for the current mean is a fixed multiple ($F$) of the vector between the randomly selected means. As a result, the number of possible locations that a cluster can mutate to is finite and related to the number of clusters. This lack of flexibility may either slow convergence when larger jumps are needed, or nuance when small movements to the cluster are needed in the final generations (which is especially useful for adjusting the overlap between clusters).
% This operator is illustrated in \autoref{fig:mohawks_mut-de}, where we can see distinct points that the mean can mutate to. This is illustrated by, for the two means randomly selected, the two new means corresponding to $F=1.0$ and $F=2.0$. 

\subsection{Experimental Setup}
To investigate the convergence properties of these operators, and whether they show a bias towards separating or bringing together clusters, we set up the following experiments. We set a low (0.2) and high (0.9) \silhtarget~to optimize towards poorly- and well-separated clusters respectively, as well as initializing the clusters either together or far apart. These four scenarios test the general ability of the operators to move cluster centroids, and are tested using $D \in \{2, 50\}$ dimensions to check robustness at a higher dimensionality. For each scenario, HAWKS is run for 30 times to assess the robustness over different initializations, and for 100 generations to assess the stability of the operators. For the DE-inspired operator, we use $F=1$ to strike a balance between convergence speed and potential oscillation when clusters are close together.

\subsection{Results}
Fig.~S-\ref{fig:mut_conv_2d} shows convergence plots for the four different scenarios in $2D$, showing the average silhouette width (and not the fitness, to make it clear when \silhall~is above or below the target) across the generations. The target, \silhtarget, is shown by the dashed horizontal line.

\begin{figure*}[ptb]
    \centering
    \subfloat[Clusters initialized overlapping, $\silhtarget=0.2$\label{fig:mut_conv_2d_olap_low}]{\includegraphics[width=0.42\textwidth]{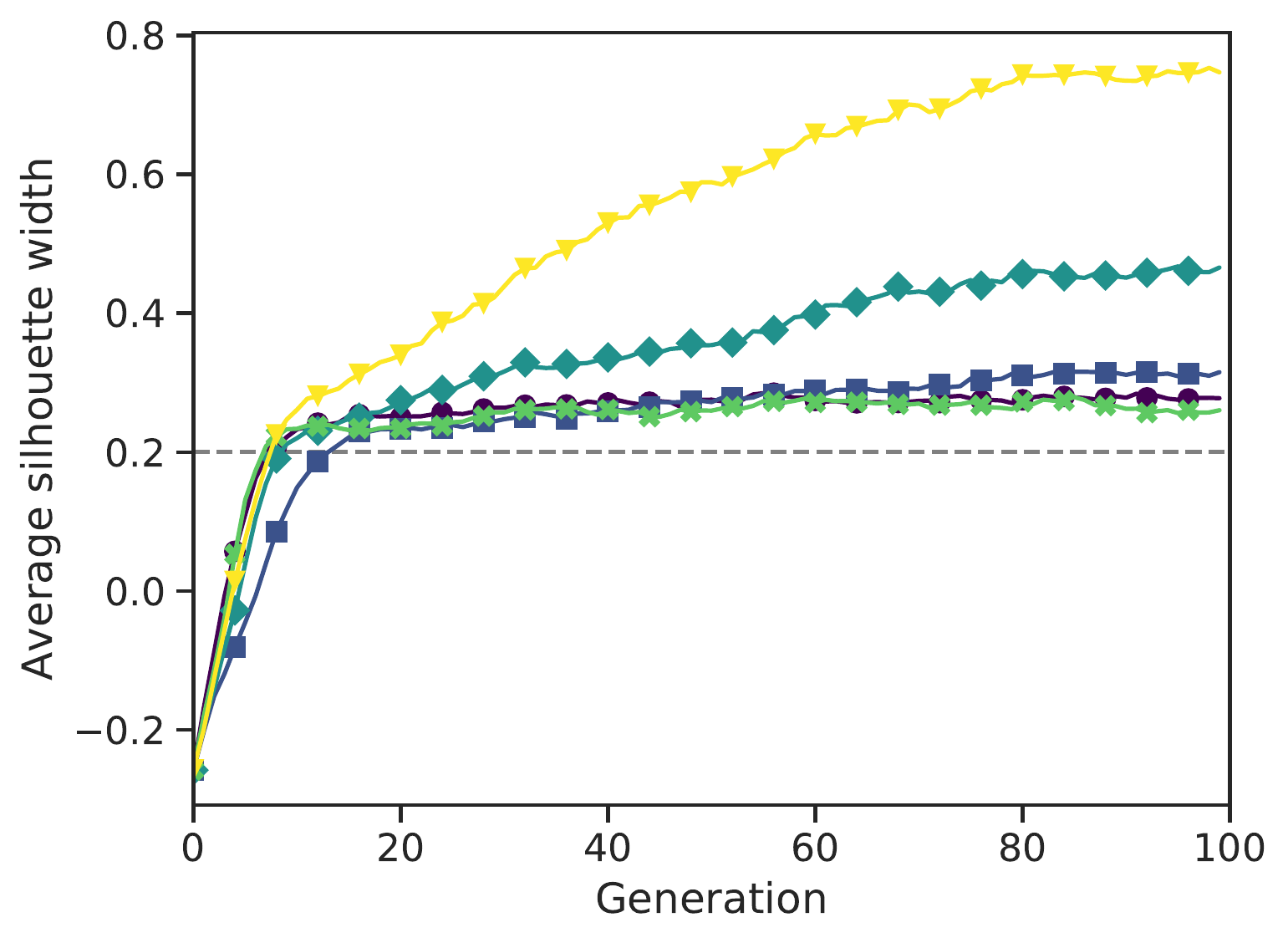}}
    \hfill
    \subfloat[Clusters initialized apart, $\silhtarget=0.2$\label{fig:mut_conv_2d_apart_low}]{\includegraphics[width=0.545\textwidth]{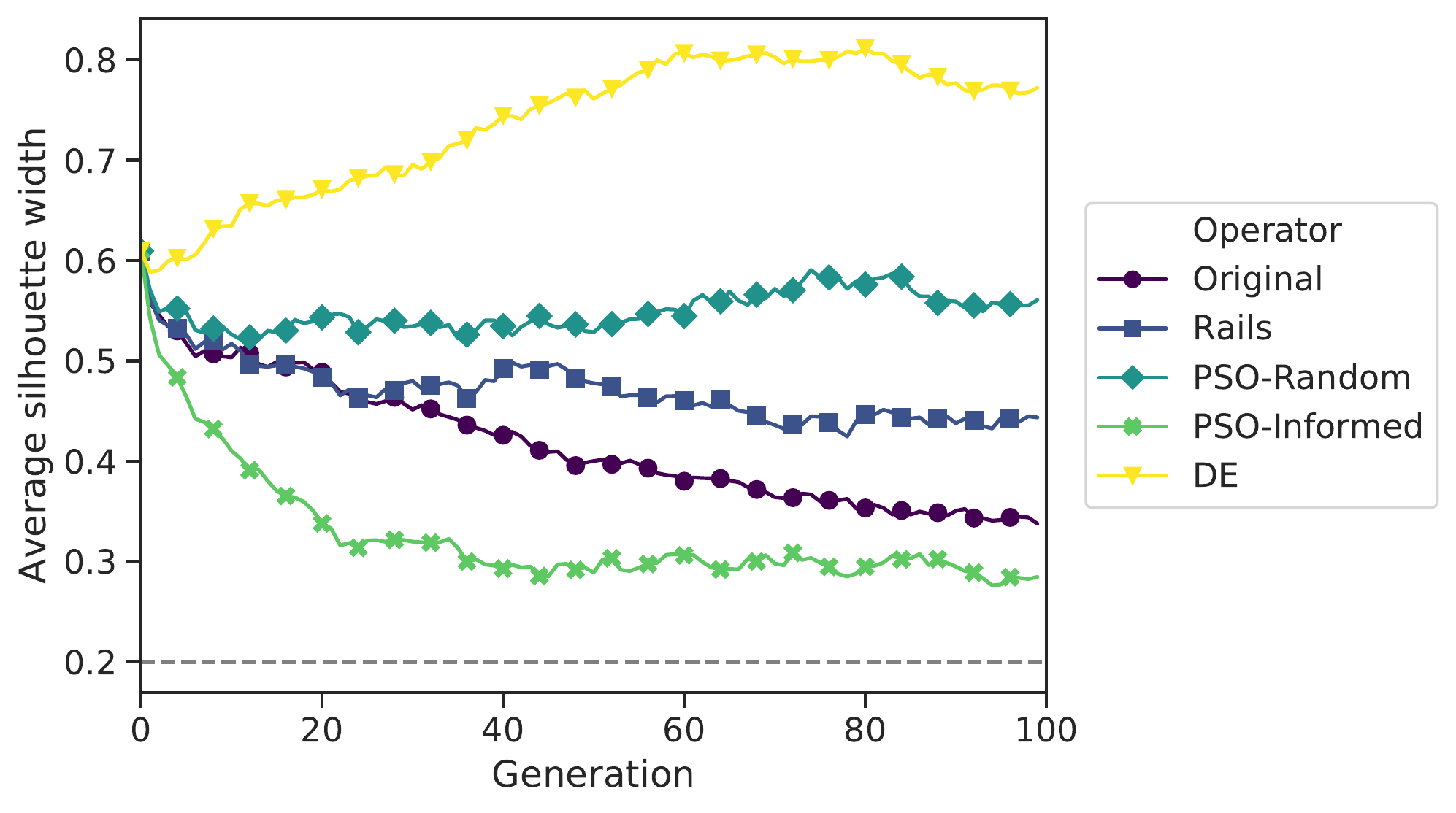}}
    \hfill
    \subfloat[Clusters initialized overlapping, $\silhtarget=0.9$\label{fig:mut_conv_2d_olap_high}]{\includegraphics[width=0.42\textwidth]{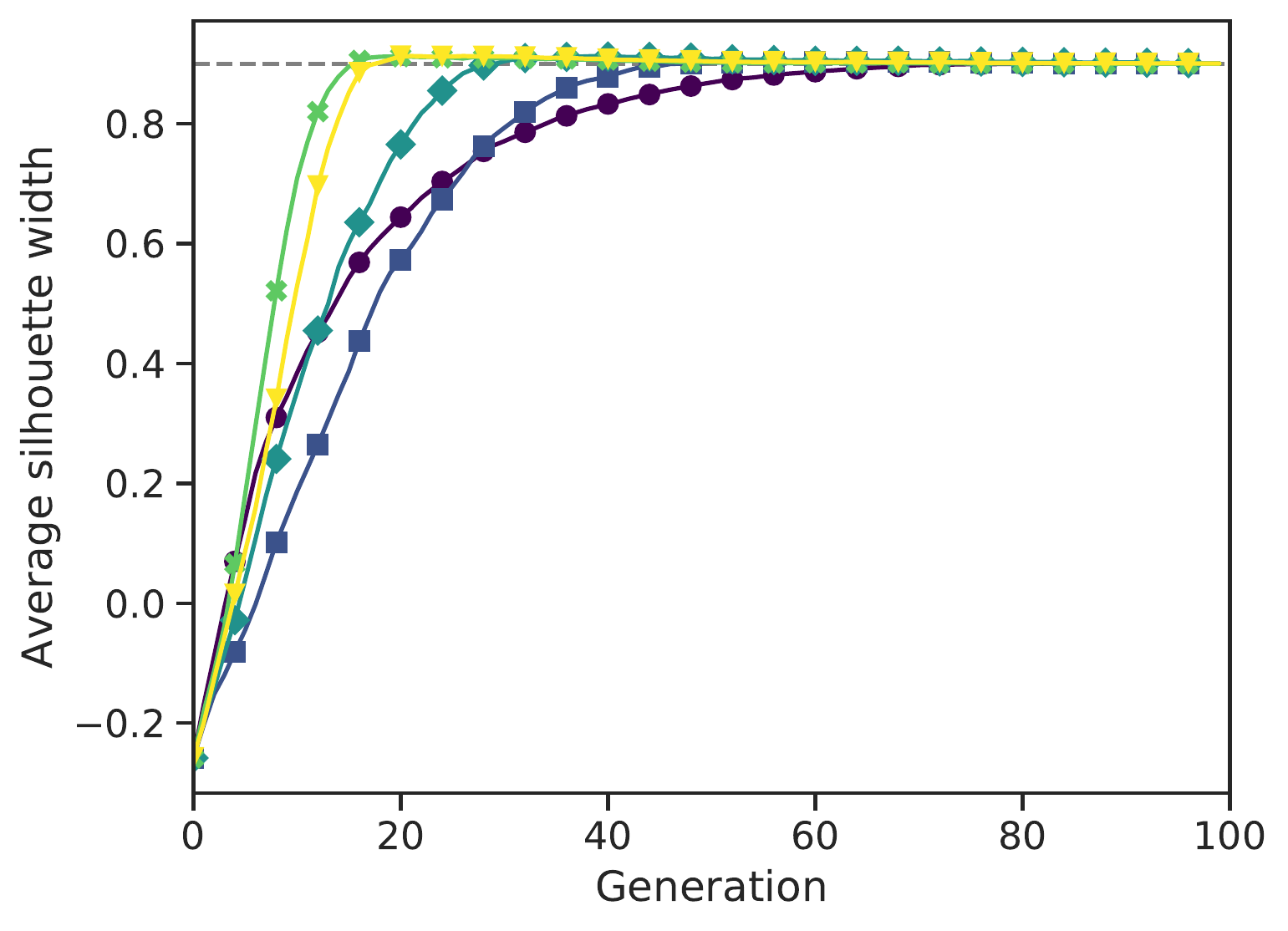}}
    \hfill
    \subfloat[Clusters initialized apart, $\silhtarget=0.9$\label{fig:mut_conv_2d_apart_high}]{\includegraphics[width=0.545\textwidth]{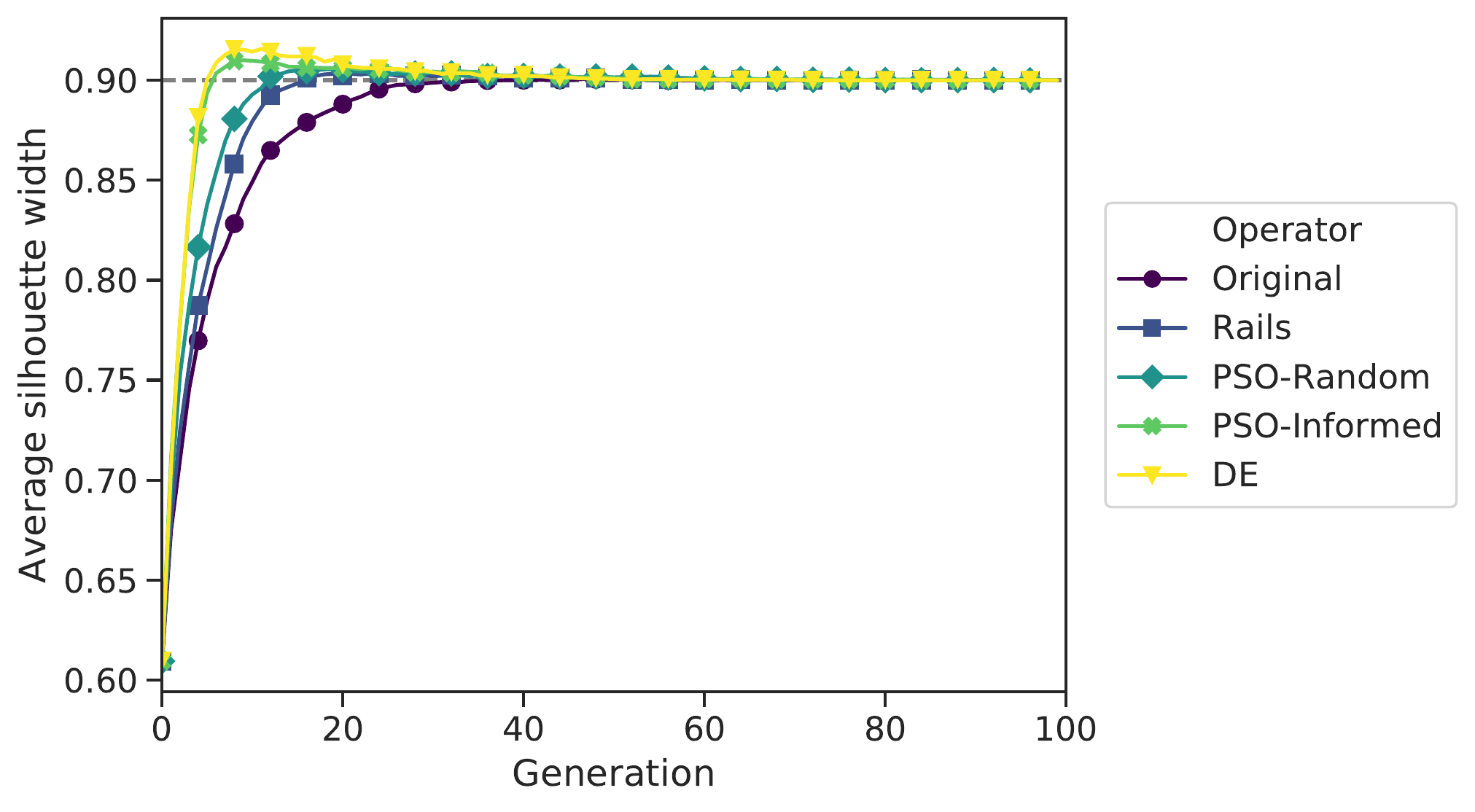}}
    \caption{Convergence plot showing the average (across all 30 runs) silhouette width across the generations when generating 2-dimensional datasets for our four scenarios. The dashed line shows the target silhouette width (\silhtarget).}
    \label{fig:mut_conv_2d}
\end{figure*}

In Fig.~S-\ref{fig:mut_conv_2d_olap_low}, all mutation operators are quickly able to move the clusters apart to increase the silhouette width, but we see a difference in the stability after this as some operators (predominantly DE and PSO-random) further increase the silhouette width above $\silhtarget$, highlighting a mechanistic bias. Fig.~S-\ref{fig:mut_conv_2d_apart_low} has the same low \silhtarget, but the initial clusters are further apart (requiring the optimization to bring clusters together). The DE-inspired operator is unable to decrease the silhouette width, and the optimization is driven by minimizing the \emph{overlap} constraint (Fig.~S-\ref{fig:mut_olap_2d_apart_low}). While the PSO-random operator does not converge, the quick decrease in the silhouette width for the PSO-informed operator shows that this is not a mechanistic issue, but one of directionality. Mutually exclusive satisfaction of the fitness (\silhtarget) and the \emph{overlap} constraint prevent any operator from fully reaching $\silhtarget=0.2$. The slow silhouette width decrease with the original operator highlights the issue with a static step size.

In Fig.~S-\ref{fig:mut_conv_2d}c--d we can see that there is no significant difference beyond the speed of convergence between the operators when optimizing for $s_{t} = 0.9$, independent of the initialization used.
No such difficulties with any of the operators can be seen for either initialization in Fig.~S-\ref{fig:mut_conv_2d}c--d, where the target silhouette width is high. None of the operators drift once converged, as this target also satisfies the \emph{overlap} constraint.

\begin{figure*}[ptb]
    \centering
    \subfloat[Clusters initialized overlapping, $\silhtarget=0.2$\label{fig:mut_olap_2d_olap_low}]{\includegraphics[width=0.42\textwidth]{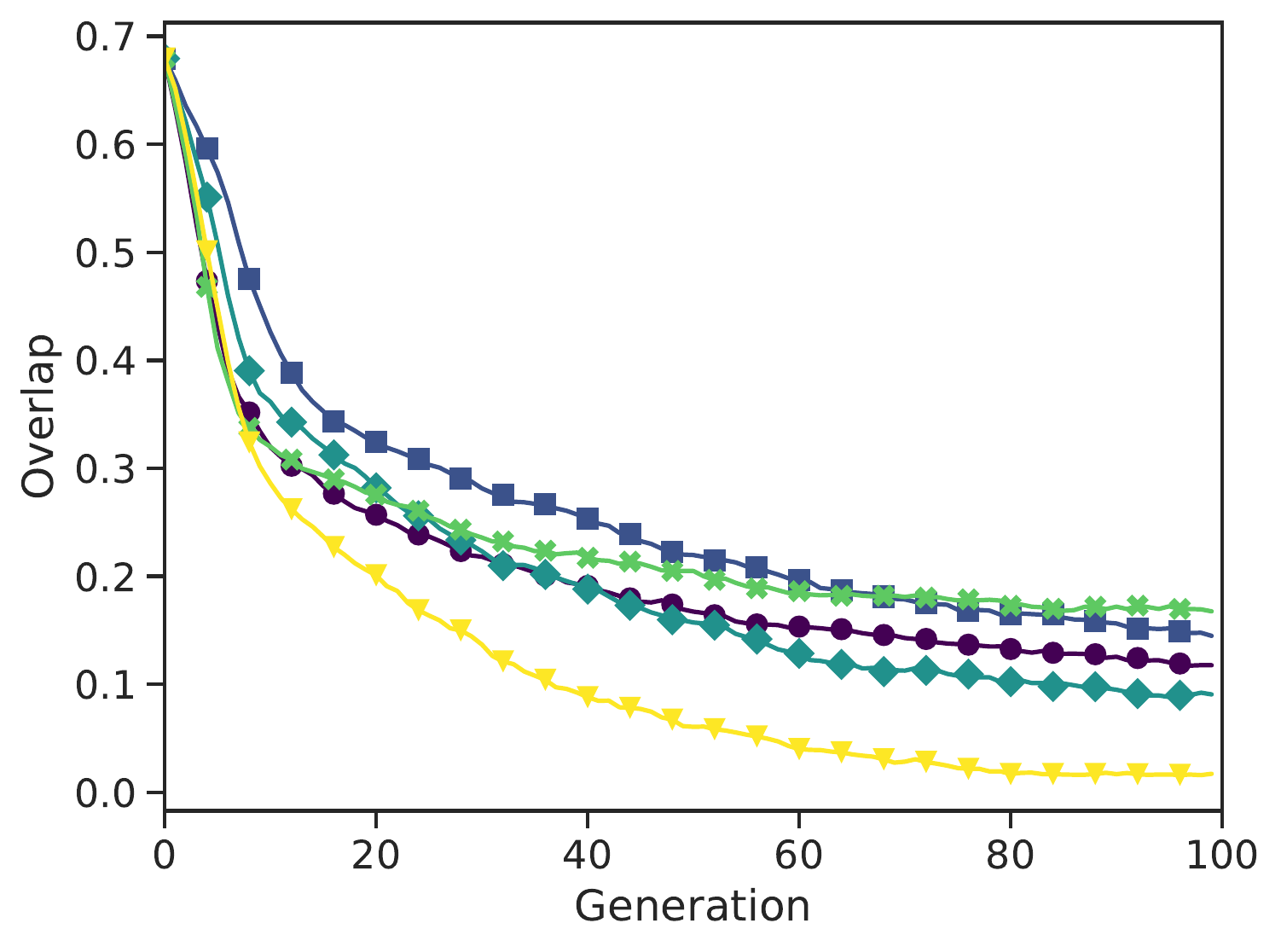}}
    \hfill
    \subfloat[Clusters initialized apart, $\silhtarget=0.2$\label{fig:mut_olap_2d_apart_low}]{\includegraphics[width=0.545\textwidth]{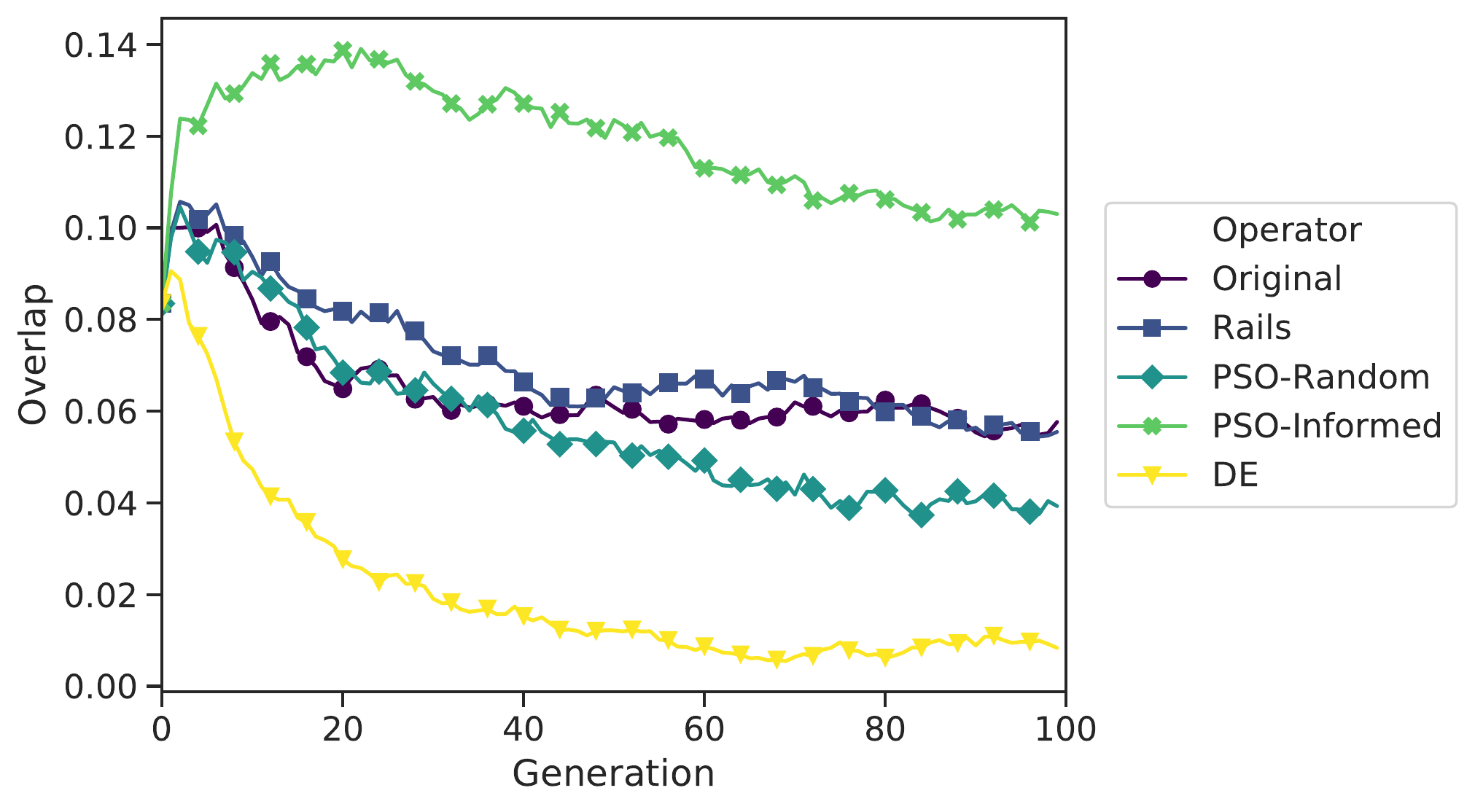}}
    \caption{Convergence plot showing the average (across all 30 runs) of the \emph{overlap} constraint across the generations when generating 2-dimensional datasets.}
    \label{fig:mut_olap_2d}
\end{figure*}

In Fig.~S-\ref{fig:mut_olap_2d_olap_low}, we can see the minimization of the \emph{overlap} constraint, which is naturally greater for the operators that increased the silhouette width. This indicates a difference in the relative ease of satisfying $\silhtarget$ and minimizing overlap between the operators, particularly with the lack of pressure (afforded by $P_{f}=0.5$) towards either one. With such a low $\silhtarget$, there is a strong inverse correlation between the overlap and fitness. The PSO-informed operator has an explicit drive to avoid a drift of the silhouette width away from the target, but the vast difference in drift for the DE-inspired operator illustrates a clear behavioural difference. In Fig.~S-\ref{fig:mut_olap_2d_apart_low}, when the clusters are initialized apart (and thus the task is to bring them closer together), we can see that the DE-inspired does consistently minimize the \emph{overlap} (as it erroneously increases the silhouette width as the clusters drift apart). The other operators also decrease the \emph{overlap}, apart from PSO-informed where the embedded preference towards the fitness disturbs the minimization of the \emph{overlap}.

This preference is further illustrated when we look at Fig.~S-\ref{fig:mut_sw-vs-olap}, where the silhouette width and \emph{overlap} of the best individuals from each run are plotted when the target silhouette width is low ($\silhtarget = 0.2$). The PSO-informed operator creates far less diversity in terms of the \emph{overlap}, thus generating less diverse datasets. To further investigate these operators, we now look at the same scenarios in $50D$.

\begin{figure}
    \centering
    \includegraphics[width=\columnwidth]{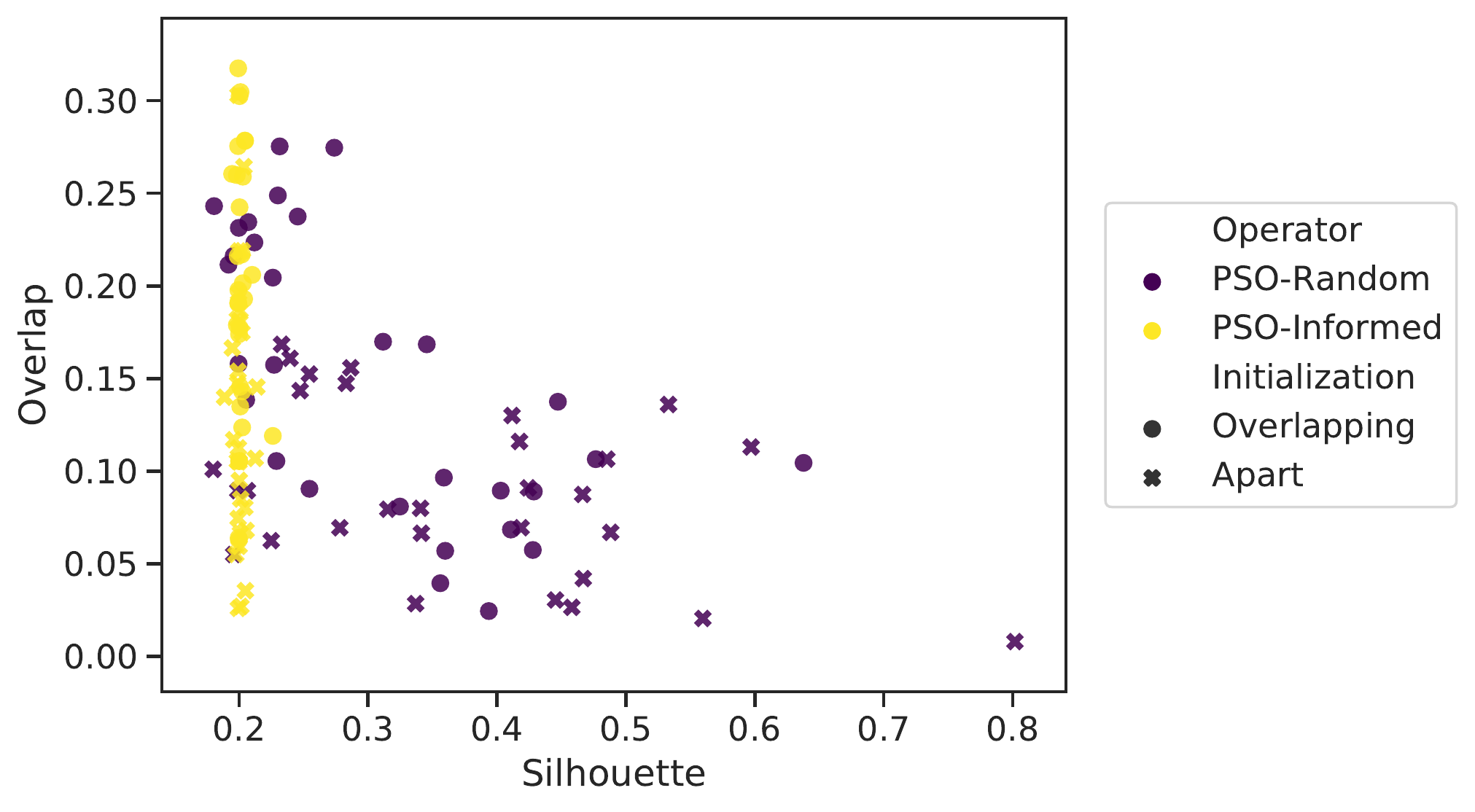}
    \caption{Scatter plot of the silhouette width and \emph{overlap} constraint for each of the best individuals from every run for the low silhouette width target.}
    \label{fig:mut_sw-vs-olap}
\end{figure}

\begin{figure*}[bt]
    \centering
    \subfloat[Clusters initialized overlapping, $\silhtarget=0.2$\label{fig:mut_conv_50d_olap_low}]{\includegraphics[width=0.42\textwidth]{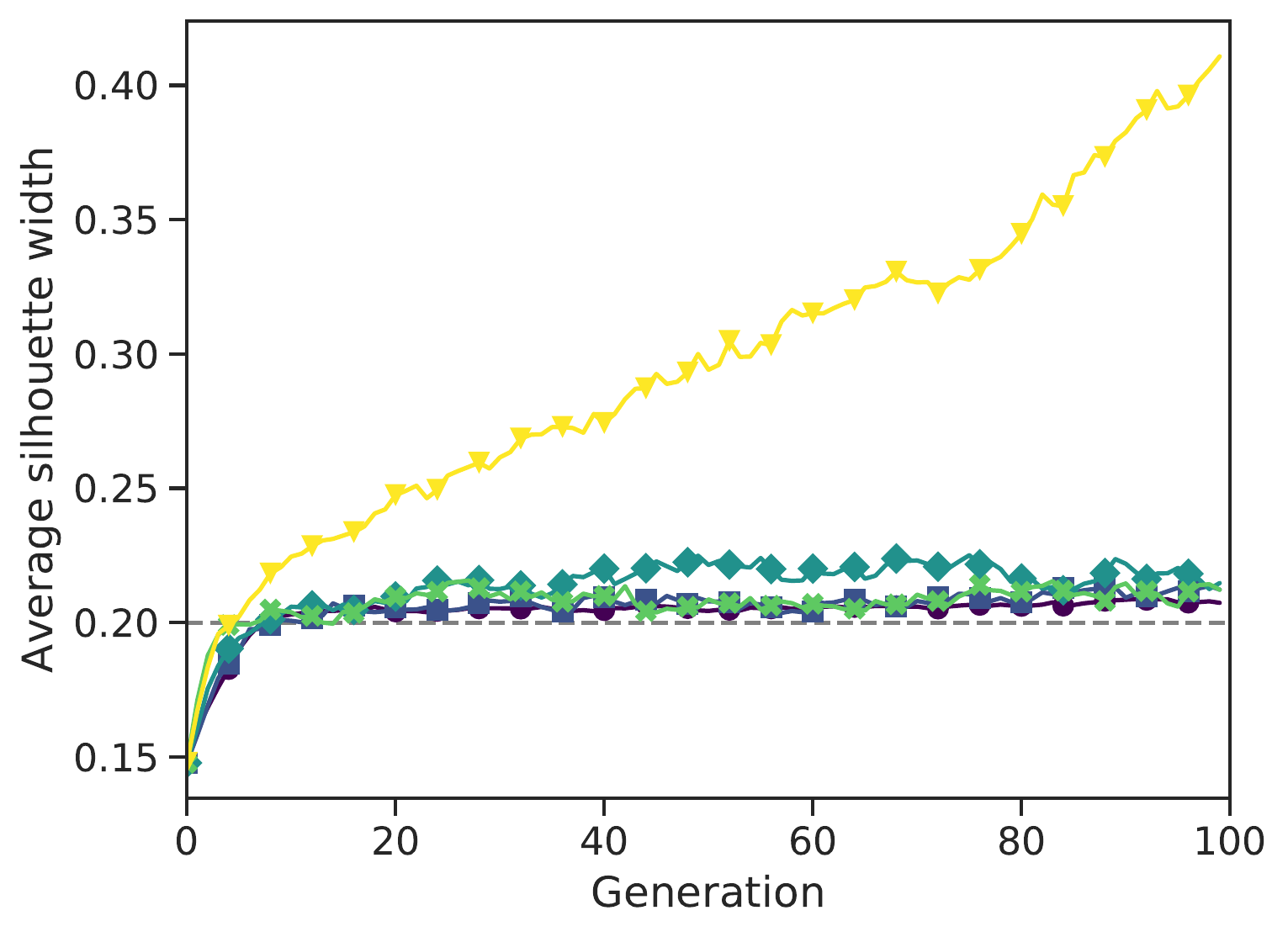}}
    \hfill
    \subfloat[Clusters initialized apart, $\silhtarget=0.2$\label{fig:mut_conv_50d_apart_low}]{\includegraphics[width=0.545\textwidth]{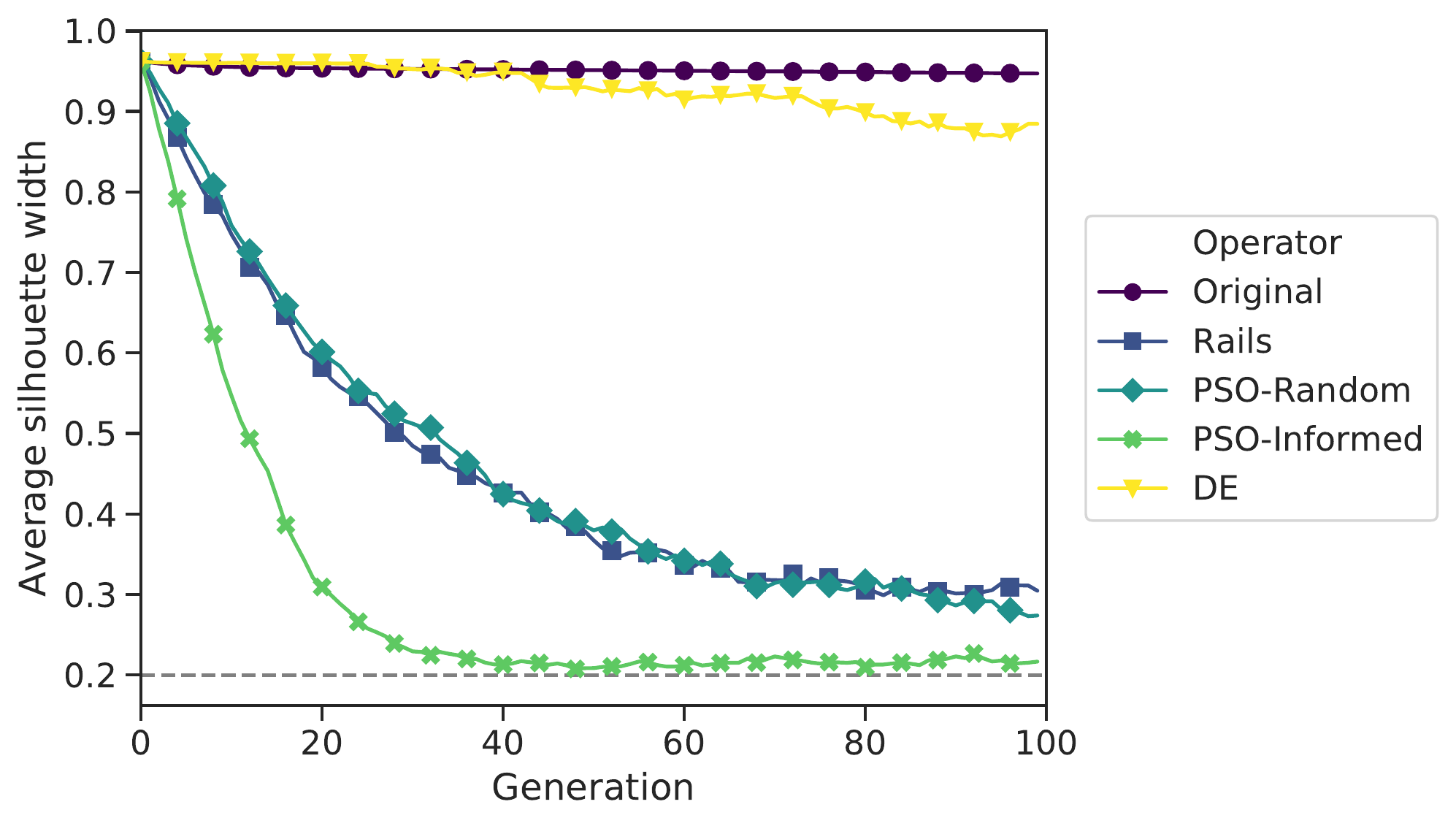}}
    \hfill
    \subfloat[Clusters initialized overlapping, $\silhtarget=0.9$\label{fig:mut_conv_50d_olap_high}]{\includegraphics[width=0.42\textwidth]{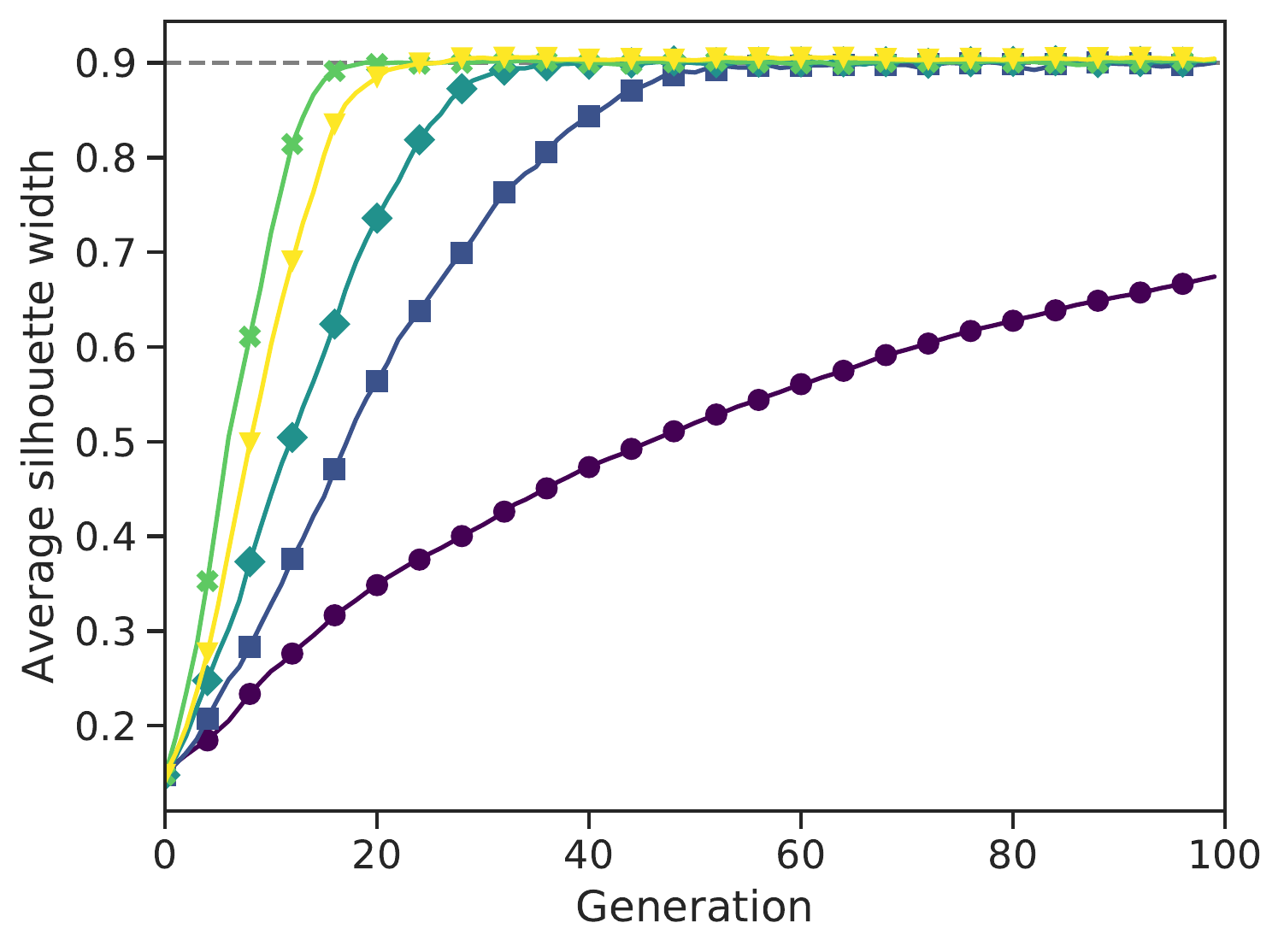}}
    \hfill
    \subfloat[Clusters initialized apart, $\silhtarget=0.9$\label{fig:mut_conv_50d_apart_high}]{\includegraphics[width=0.545\textwidth]{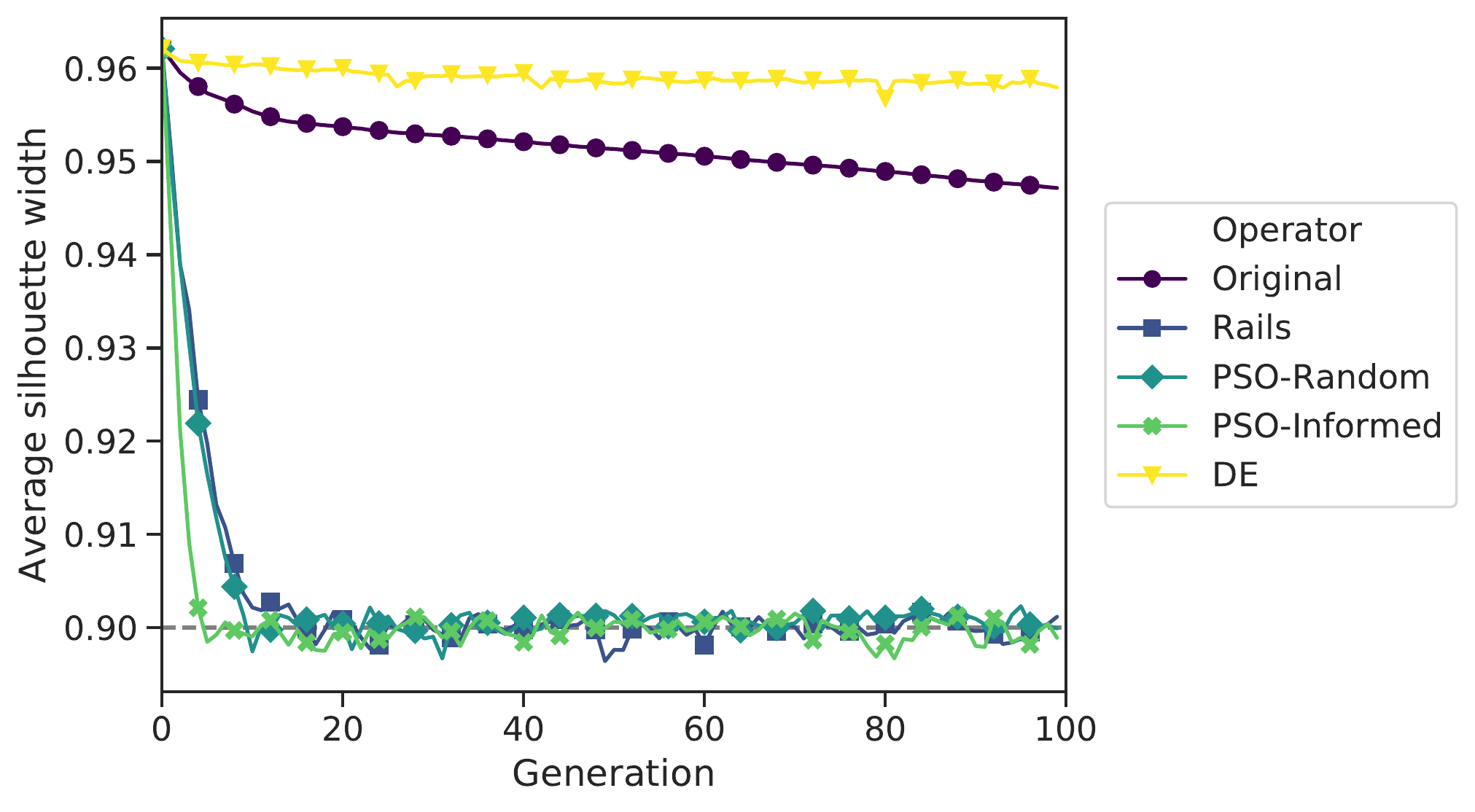}}
    \caption{Convergence plot showing the average (across all 30 runs) silhouette width across the generations when generating 50-dimensional datasets for our four scenarios.}
    \label{fig:mut_conv_50d}
\end{figure*}

Fig.~S-\ref{fig:mut_conv_50d} shows convergence plots for the four different scenarios in $50D$, which was previously where issues with the original operator were first identified. In Fig.~S-\ref{fig:mut_conv_50d_olap_low}, when the scenario is to keep clusters close together, the drift of the DE-inspired operator is more pronounced relative to the other operators, which remain mostly stable. The bias of the DE-inspired operator towards moving clusters further apart is further shown in Fig.~S-\ref{fig:mut_conv_50d_apart_low}, where the original operator is also incapable of bringing clusters closer together. As the DE-inspired operator uses the direction and magnitude of the vector between two random clusters, it does not necessarily move the cluster being mutated in the direction of other clusters, an issue which is more prevalent in higher dimensions. As expected, the PSO-informed operator converges the fastest, whereas the ``rails'' and PSO-random operators increasingly slow to decrease the silhouette width (largely due to trying to minimize the overlap).

The ability of our original operator to move clusters away from each other in higher dimensions is highlighted in Fig.~S-\ref{fig:mut_conv_50d_olap_high}, which is clearly contrasted with the proposed operators that are all able to converge significantly faster. Once again, utilizing the individual's current silhouette width allows the PSO-informed operator to consistently move the clusters apart, converging rapidly. The speed of convergence for the DE-inspired operator is likely due to its step size being based on the distance between two random clusters, thus enabling it to rapidly move clusters apart. In Fig.~S-\ref{fig:mut_conv_50d_apart_high}, when the clusters are initialized further apart, similar behaviour to the low \silhtarget~is seen. The initial silhouette width is very close to the target, yet our original and the DE-inspired operators are still unable to significantly improve the fitness as they begin \emph{above} the target. It is likely that in this scenario, the step size is too large for the DE-operator to be useful, and too small and undirected for our original operator.

As such, in our experiments we select to use the PSO-random operator, as it provides a more nuanced mechanism for moving cluster means, but avoids introducing additional bias which would otherwise be controlled by stochastic ranking. For scenarios where the fitness is prioritized, PSO-informed may be the more useful operator, but for general use the PSO-random operator is preferred.

\section{Analyzing benchmark dataset diversity (Index mode)} \label{sec:benchmark}
In Section~\zref{main-ssec:benchmark}, we compare several collections of datasets to a set of datasets produced by HAWKS. Here, we provide some further information on the different properties of these datasets, and further supporting results.

\subsection{Dataset collection details} \label{ssec:benchmark_datasets}
Table~S-\ref{tab:dataset_params} shows some basic parameters about the different collections of datasets. Where possible, values are condensed into a range (i.e.~``6--11''), but are otherwise listed. Owing to the different nature of these datasets (some real-world, some from a generator), some values are specific to a single dataset, and others represent multiple instantiations from a generator.

\input{supp_tables/supp-table1.tex}

\subsection{EA parameter sensitivity analysis} \label{ssec:sensitivity}
To ensure that the number of generations and population size were sufficient, we re-ran HAWKS with the same parameters aside from using $G_{\text{max}}=200$ and (separately) $\mathcal{P}=100$. Fig.~S-\ref{fig:sensit_bplots} shows the problem features (Fig.~\ref{fig:sensit_features}) and clustering algorithm performance (Fig.~\ref{fig:sensit_clust}) for the additional parameter configurations, showing no significant difference in either.

\begin{figure*}
	\centering
	\subfloat[Problem features\label{fig:sensit_features}]{\includegraphics[width=0.49\textwidth]{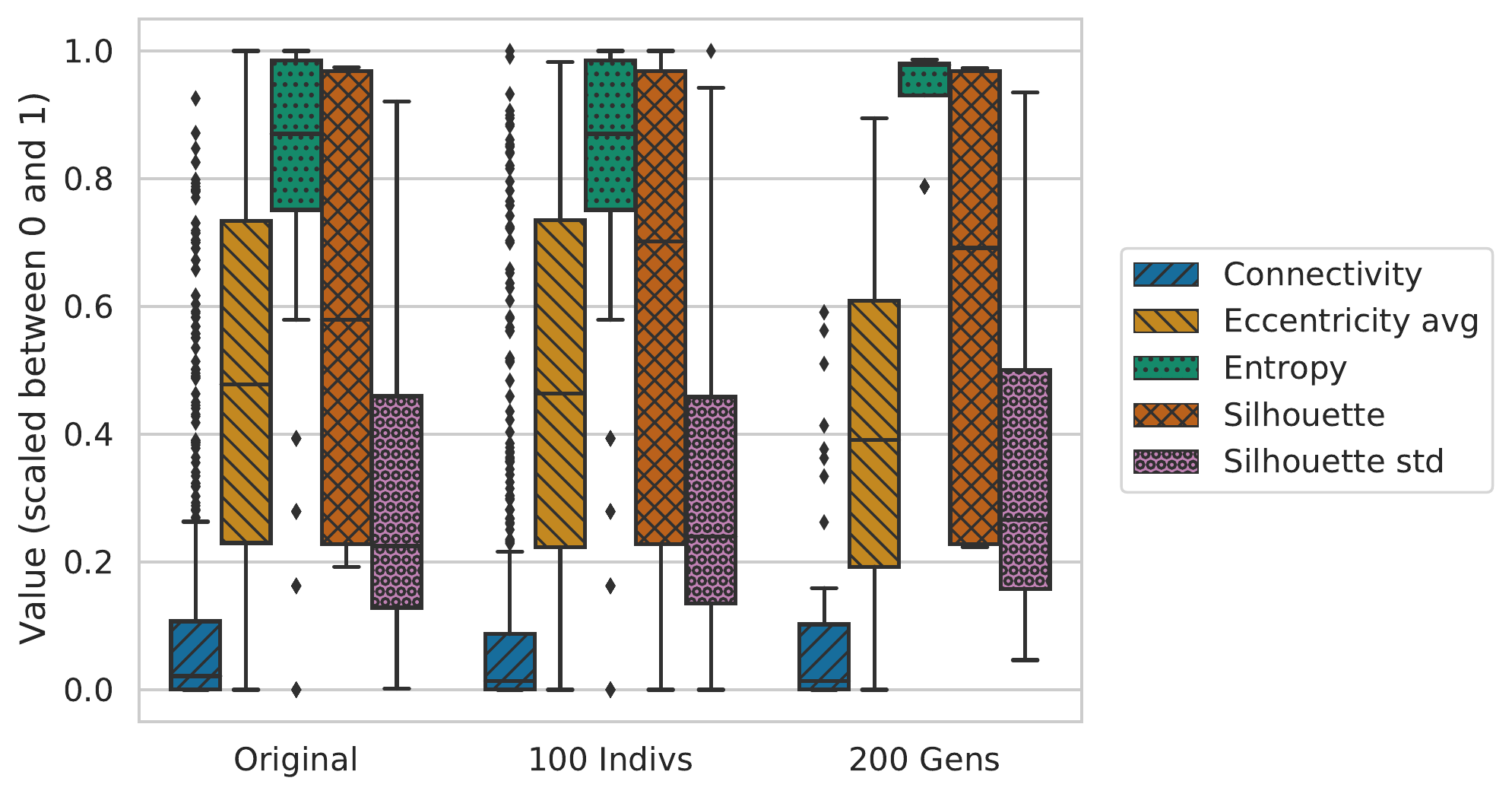}}\hfill
	\subfloat[Clustering algorithm performance\label{fig:sensit_clust}]{\includegraphics[width=0.49\textwidth]{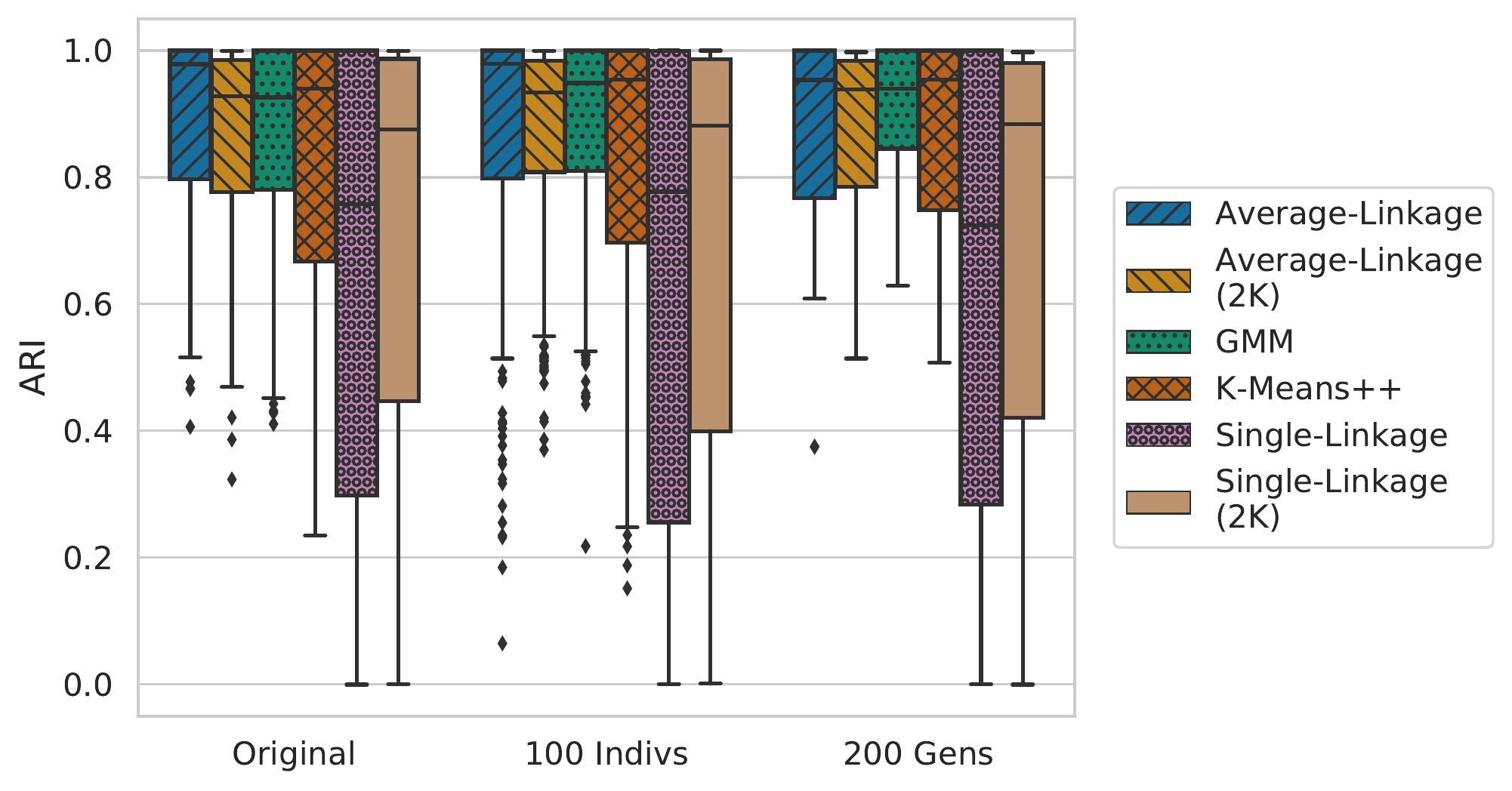}}
	\caption{Sensitivity analysis for increasing the number of generations ($G_{\text{max}}=200$) and individuals ($\mathcal{P}=100$). The clustering algorithm performance across all datasets is compared against the datasets generated for the \emph{Index} mode experiment in Section~\zref{main-ssec:benchmark}.}
	\label{fig:sensit_bplots}
  \end{figure*}

\subsection{Problem feature values across the instance space}
Fig.~\zref{main-fig:instance_space} showed the instance space created for these datasets using our set of problem features. Fig.~S-\ref{fig:benchmark_features} shows this instance space, but with the best ARI achieved by anything algorithm and the values for each of the problem features highlighted.

\begin{figure*}[tbp]
    \centering
    \subfloat[ARI\label{fig:benchmark_ARI}]{\includegraphics[width=0.46\textwidth]{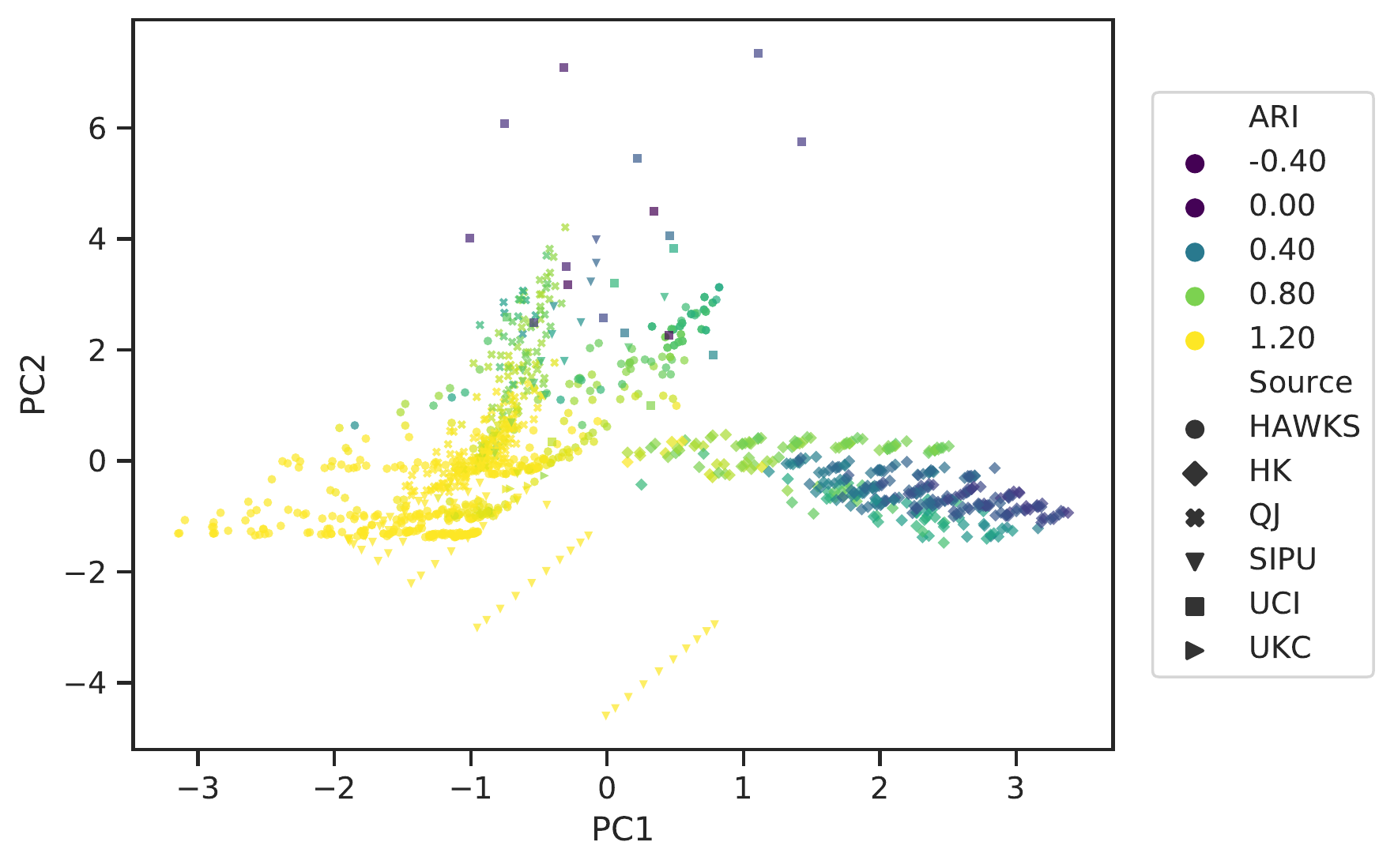}}
    \hfill
    \subfloat[Connectivity\label{fig:benchmark_connectivity}]{\includegraphics[width=0.46\textwidth]{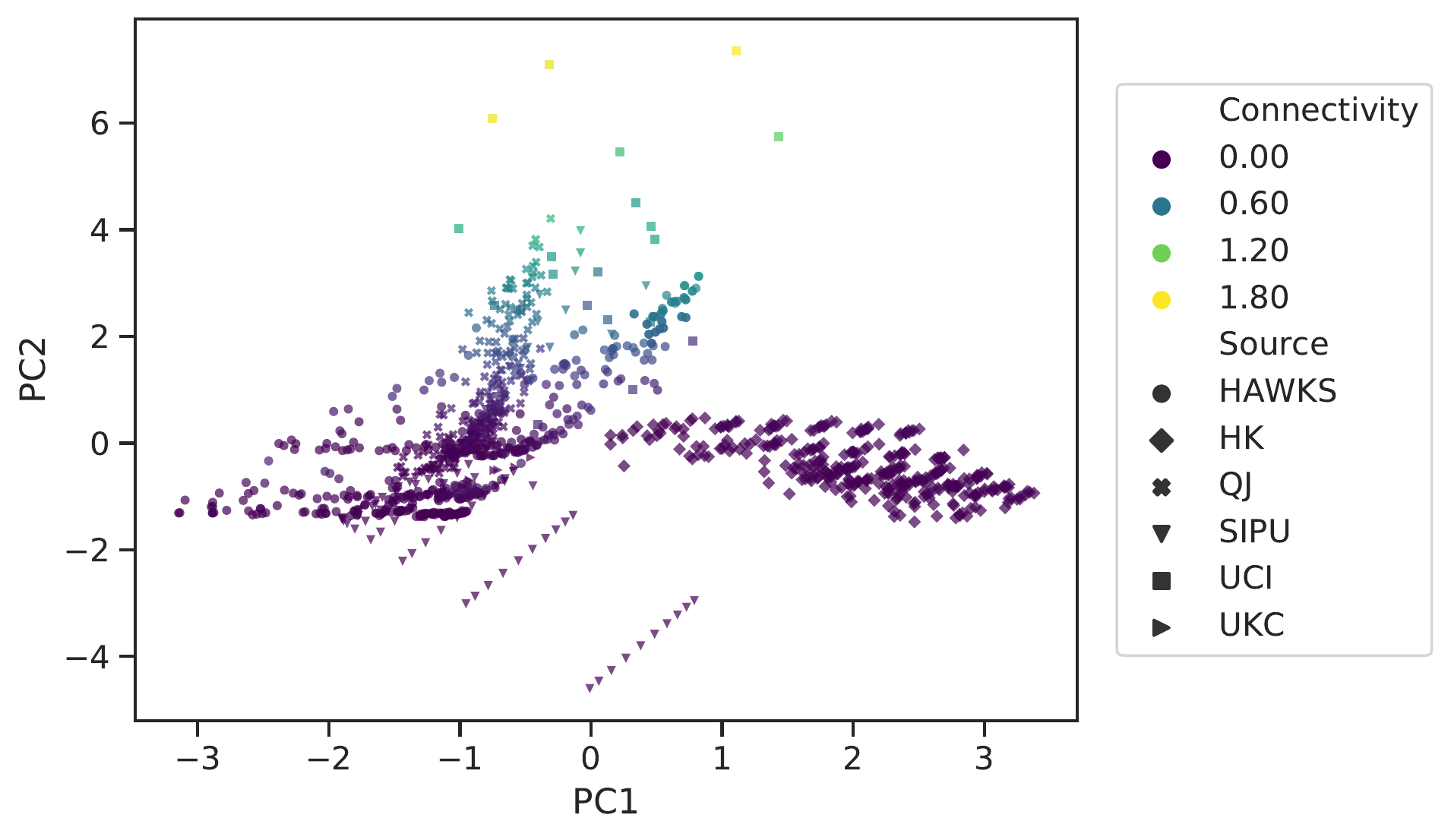}}
    \bigskip

    \subfloat[Number of dimensions\label{fig:benchmark_dims}]{\includegraphics[width=0.46\textwidth]{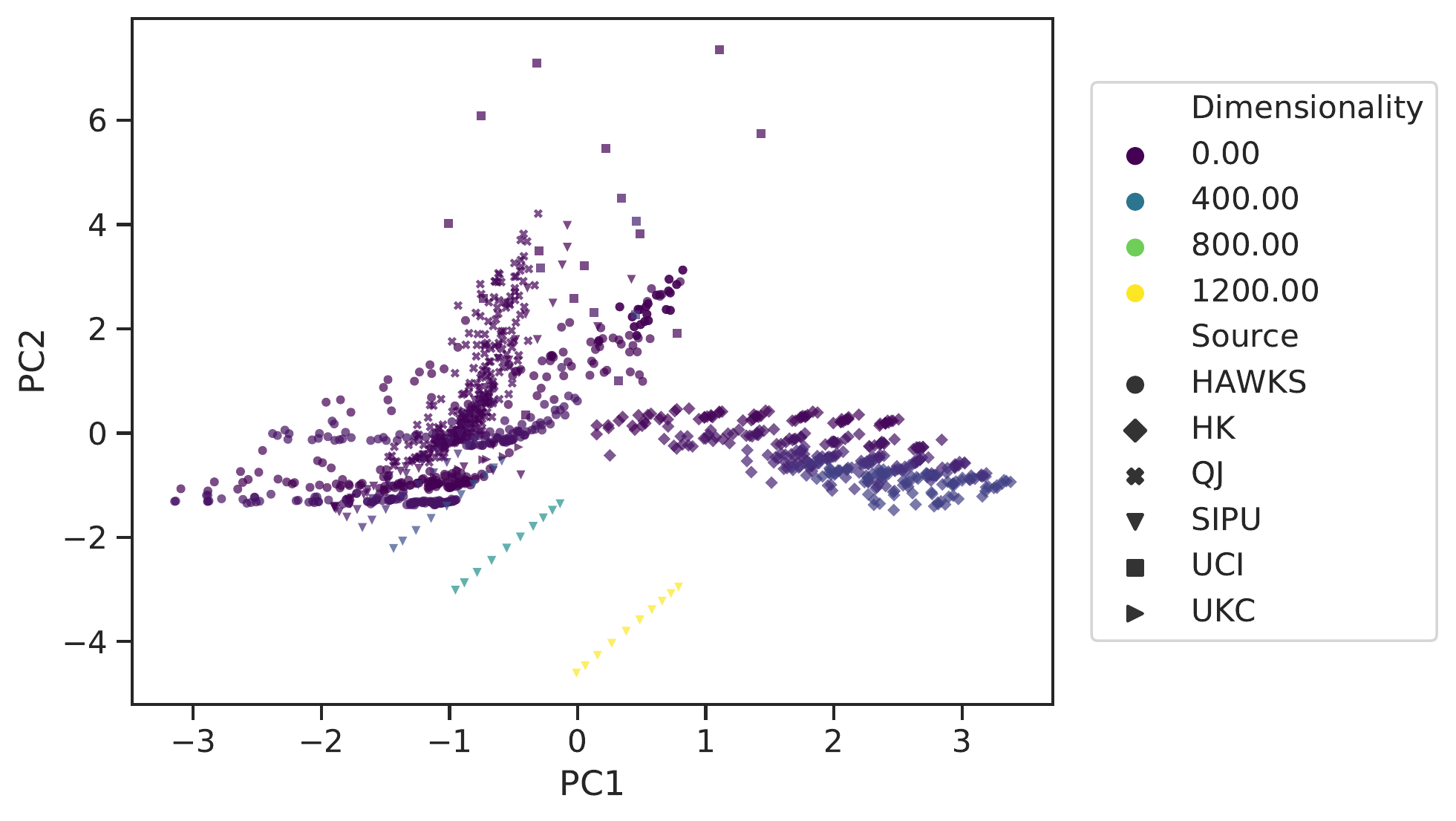}}
    \hfill
    \subfloat[Average eccentricity\label{fig:benchmark_eccen}]{\includegraphics[width=0.46\textwidth]{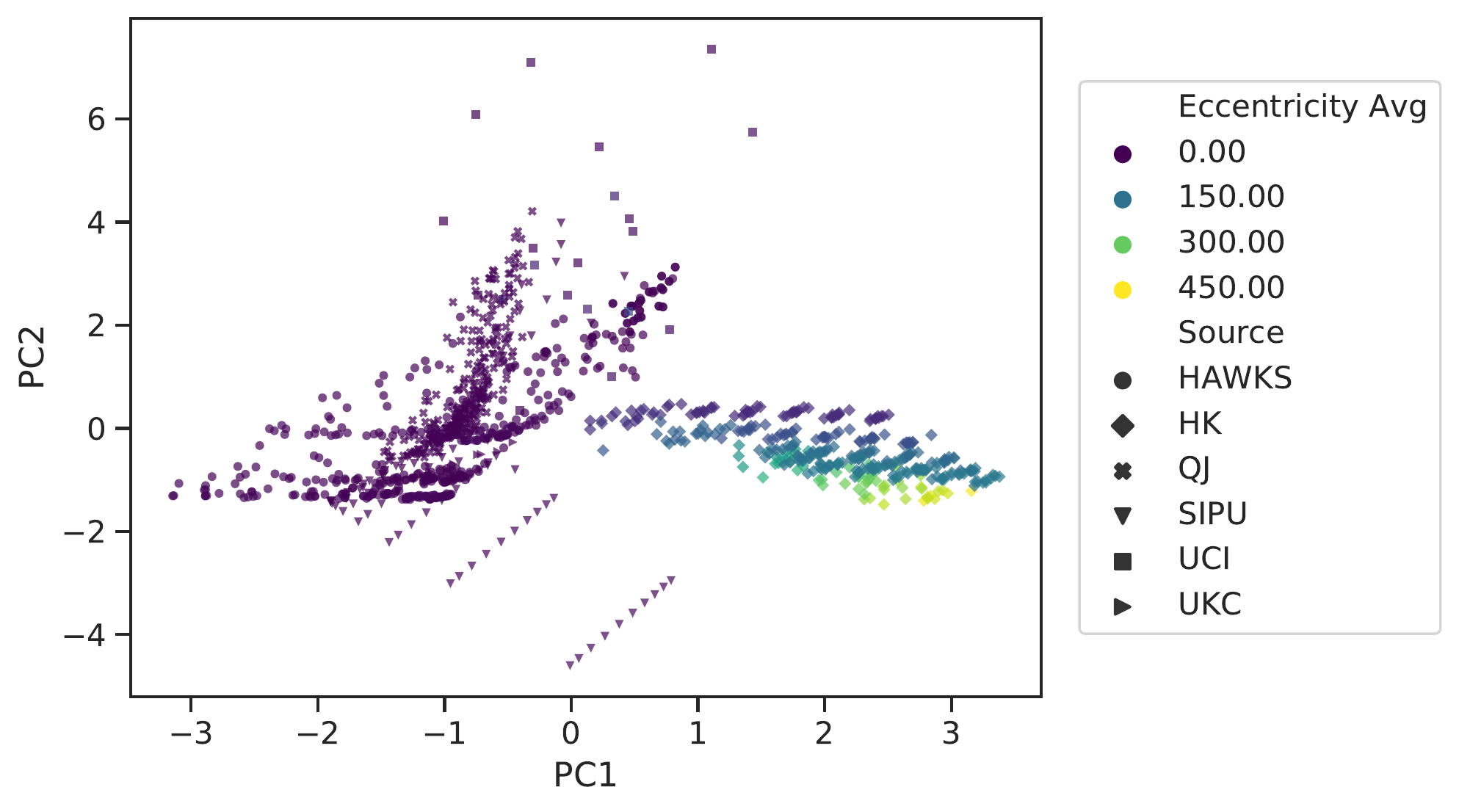}}
    \bigskip

    \subfloat[Entropy of cluster sizes\label{fig:benchmark_entropy}]{\includegraphics[width=0.46\textwidth]{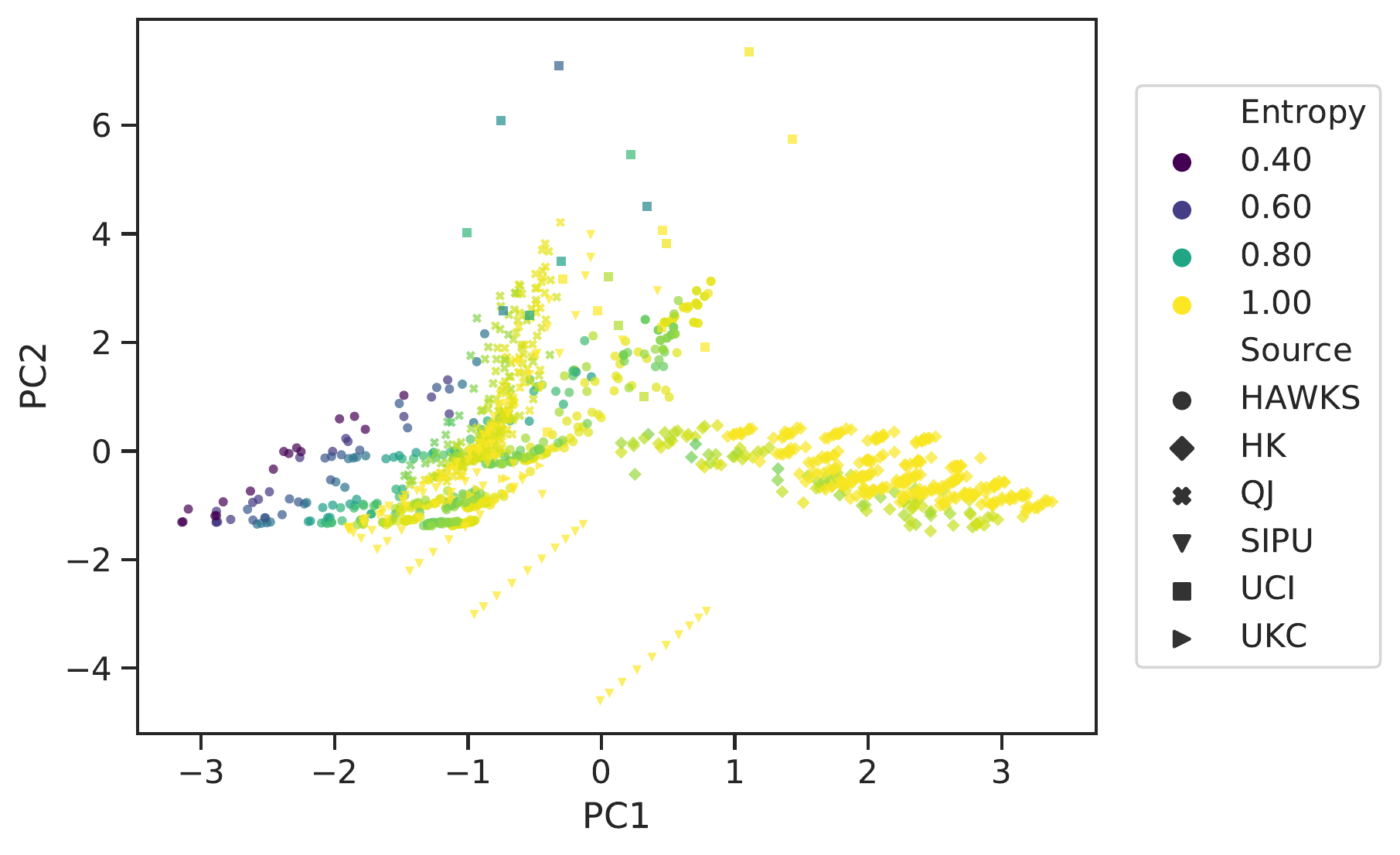}}
    \hfill
    \subfloat[Number of clusters\label{fig:benchmark_k}]{\includegraphics[width=0.46\textwidth]{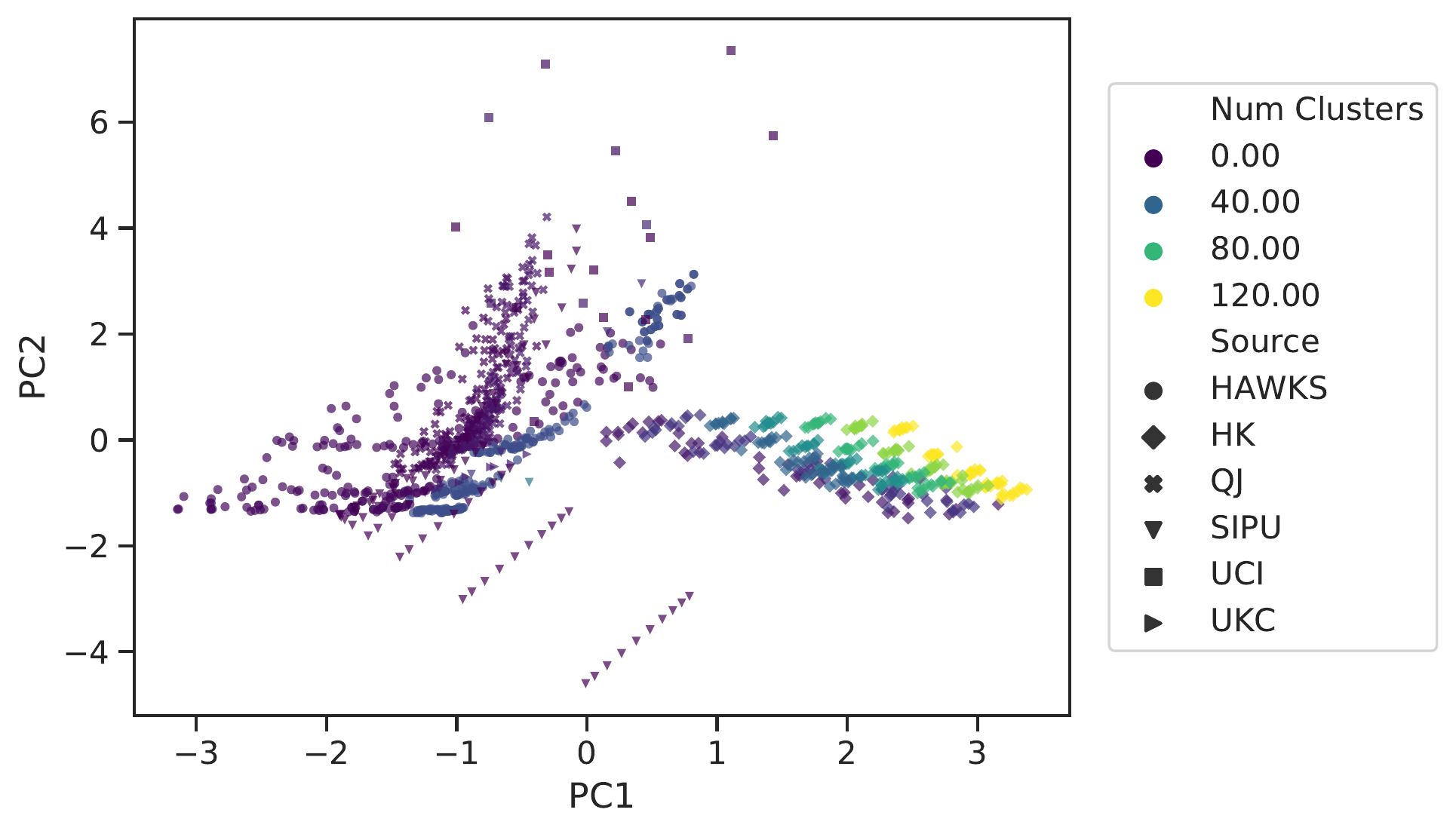}}
    \bigskip

    \subfloat[Silhouette width (average)\label{fig:benchmark_sw}]{\includegraphics[width=0.46\textwidth]{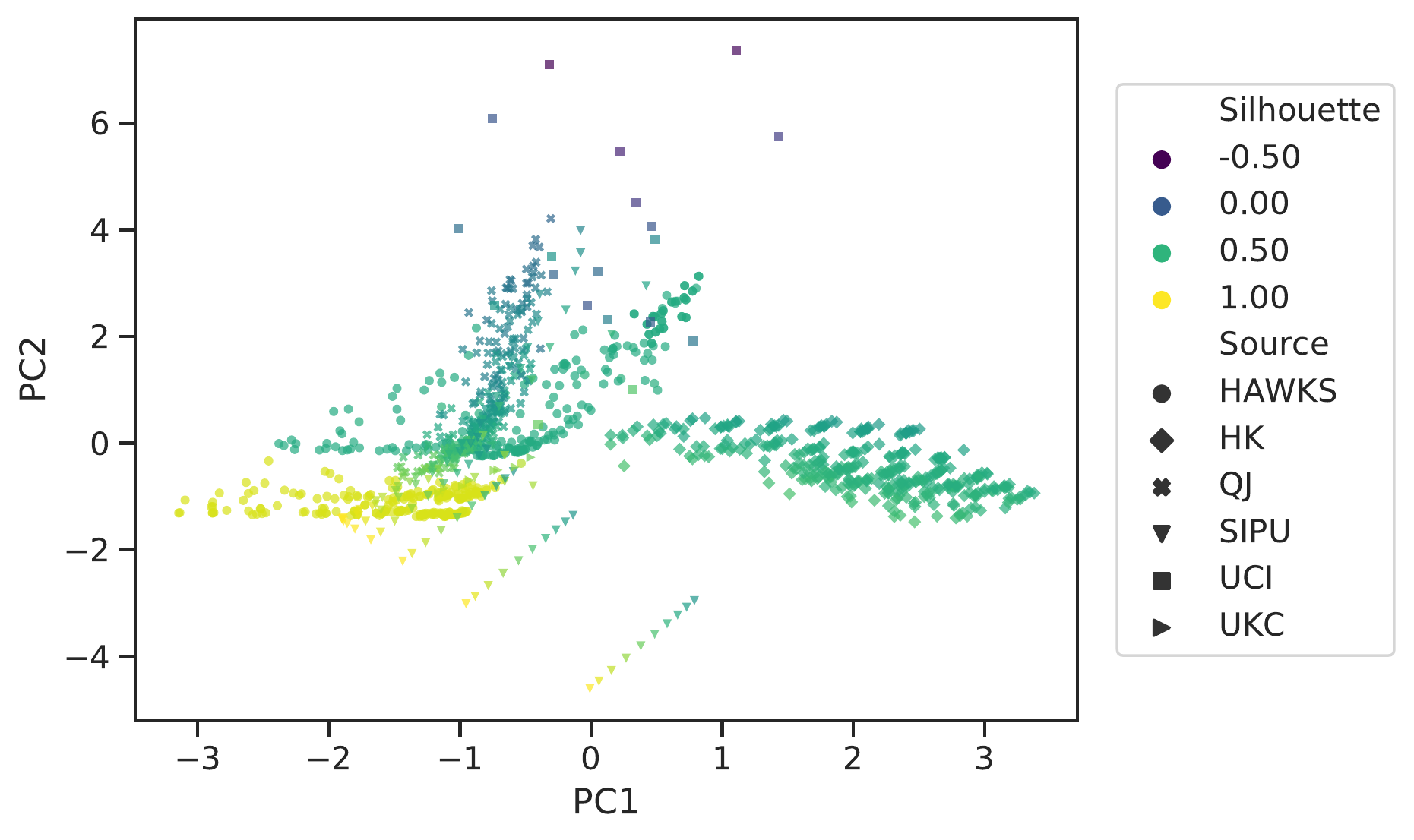}}
    \hfill
    \subfloat[Silhouette width (standard deviation)\label{fig:benchmark_sw-std}]{\includegraphics[width=0.46\textwidth]{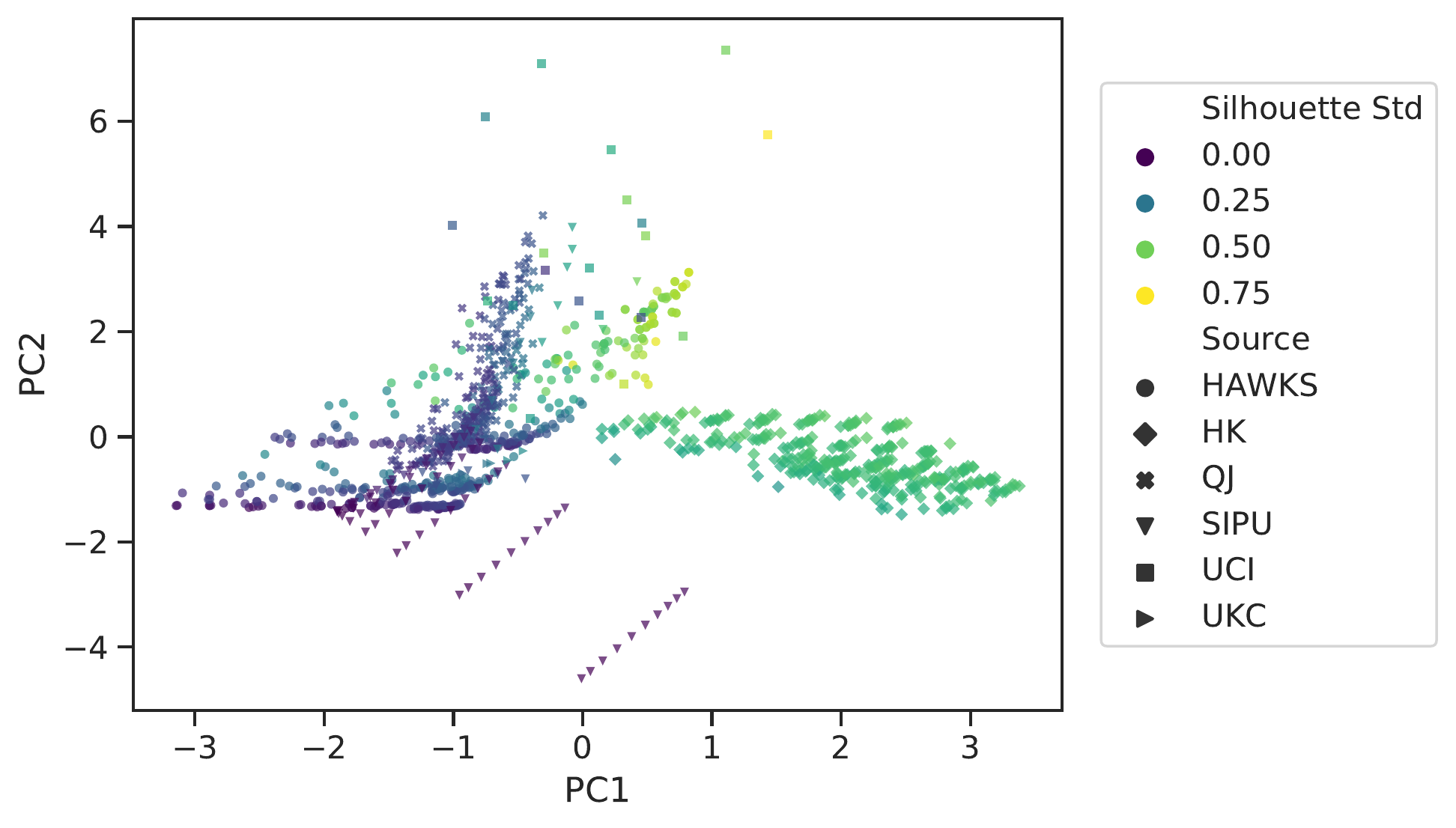}}
    \caption{The instance space created in Section~\zref{main-ssec:benchmark} showing how the values of each of the 7 problem features and the best ARI found by any algorithm vary across the space.}
    \label{fig:benchmark_features}
\end{figure*}

As we can see, for each problem feature there is a clear gradient across the space, highlighting the utility of using PCA to construct the instance space in terms of understanding how the problem features vary across the space. Interestingly, there is also a gradient for the best ARI achieved, for which lower values are generally associated with either a lower silhouette width, higher standard deviation of the silhouette width, and higher average cluster eccentricity. Overall, this highlights that using problem features specific to clustering help to create an instance space that correlates to algorithmic performance. As shown in Fig.~\zref{main-fig:instance_source-alg} however, through either the projection method or an incomplete set of problem features (as discussed in Section~\zref{main-ssec:problem_features}, it is difficult to get a complete set), there is not a complete separation of regions where a single algorithm is superior.

\input{supp_tables/supp-table2.tex}

Table~S-\ref{tab:problem_features} shows the mean and standard deviation of each of the problem features for every collection of datasets studied in this paper, clarifying the observations seen for the instance space visualizations (Fig.~S-\ref{fig:benchmark_features}). The \emph{UCI} datasets show a much higher connectivity, indicating that the nearest neighbours of data points are typically in a different cluster, and thus have poor cluster structure. This is complemented by Fig.~S-\ref{fig:benchmark_features_bplot}, which shows boxplots of these problem features for a direct visual comparison. The greater variation in the standard deviation of the silhouette width is more apparent here, which is reflected in the spread of connectivity values.

\begin{figure}
    \centering
    \includegraphics[width=0.95\columnwidth]{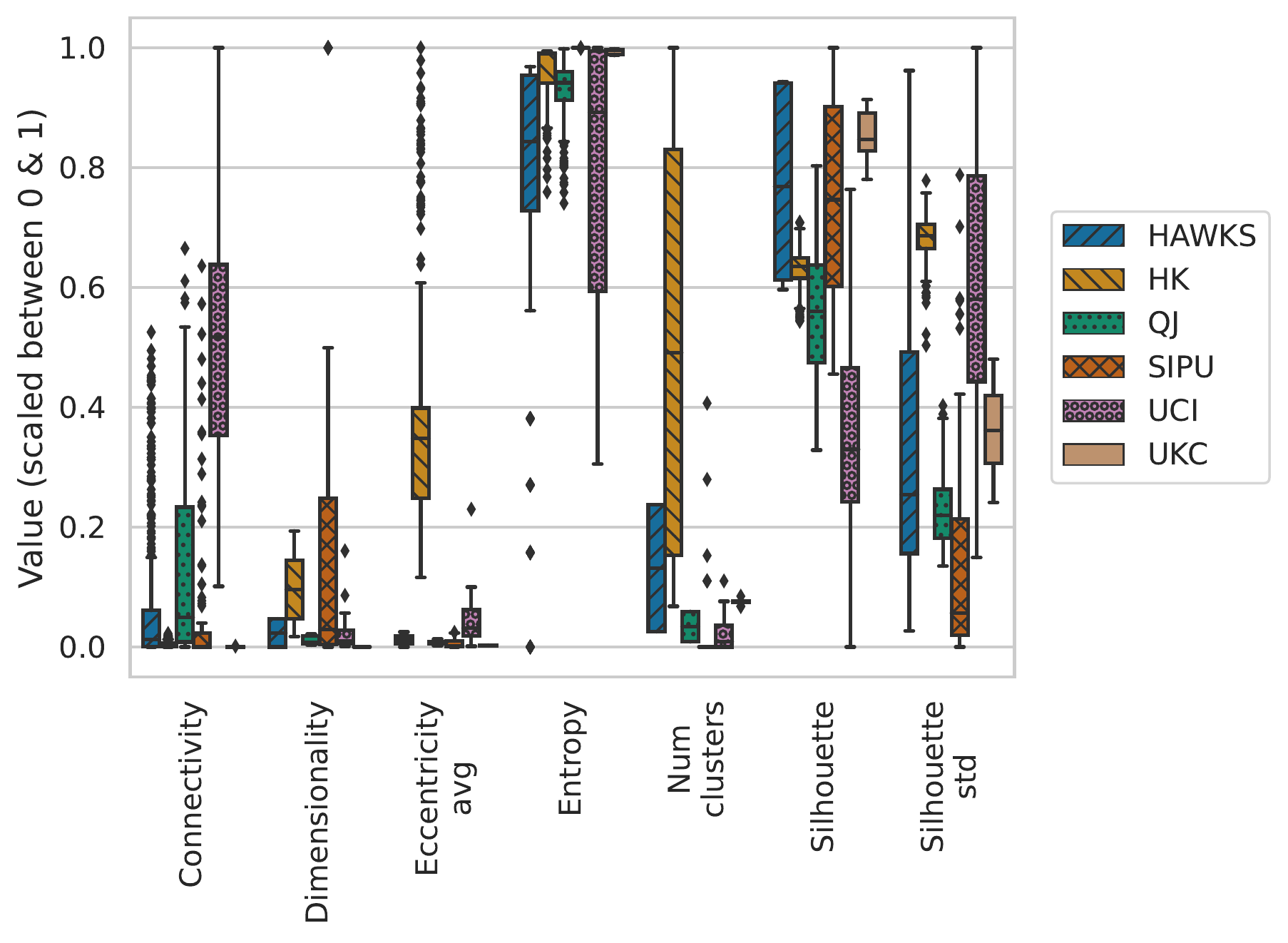}
    \caption{Boxplots of the problem features for each dataset collection.}
    \label{fig:benchmark_features_bplot}
\end{figure}

We can compare the mean values for the problem features for the \emph{Index} and \emph{Versus} modes to see if the different optimization settings had a particular bias towards e.g.\ a low connectivity. The main differences between these datasets, according to the problem features, is that the \emph{Versus} datasets have a slightly higher average connectivity, much lower eccentricity, and higher variance in the silhouette width. The lower eccentricity can be explained through both the initialization used and low dimensionality. As shown in our examples, where highly eccentric clusters were seen, this did not diminish HAWKS' capability of evolving such clusters when needed.

\subsection{Comparing performance diversity via critical difference diagrams} \label{ssec:cd_diagrams}

In the main paper, we showed the critical difference (CD) diagrams for datasets from the HAWKS and \emph{HK} generators. For completeness, here in Fig.~\ref{fig:cd_extras} we show the CD diagrams for the other dataset collections.

\begin{figure*}
	\centering
	\subfloat[\emph{QJ}\label{fig:cd_qj}]{\includegraphics[width=0.46\textwidth]{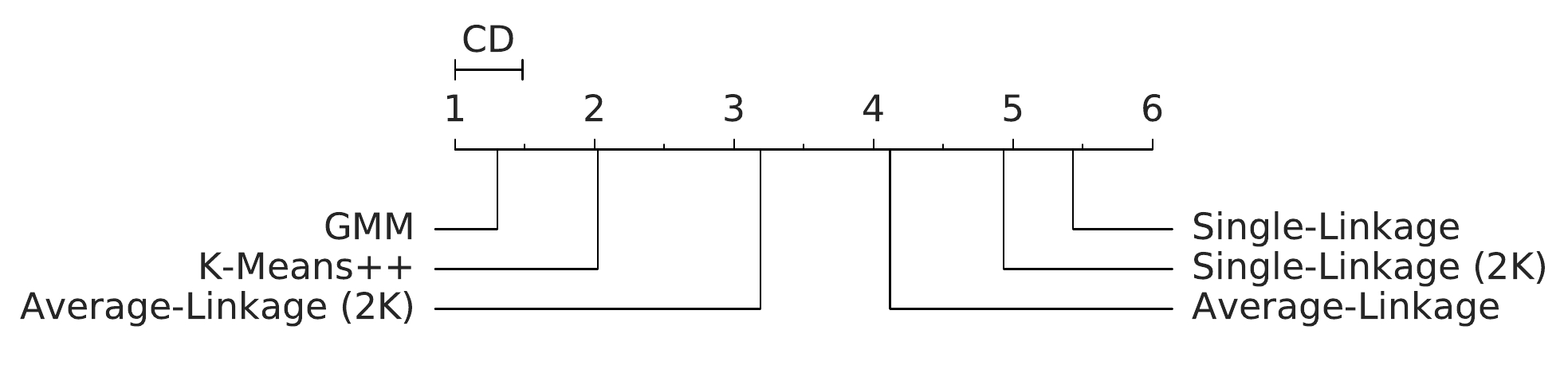}}\hfill
	\subfloat[SIPU\label{fig:cd_sipu}]{\includegraphics[width=0.46\textwidth]{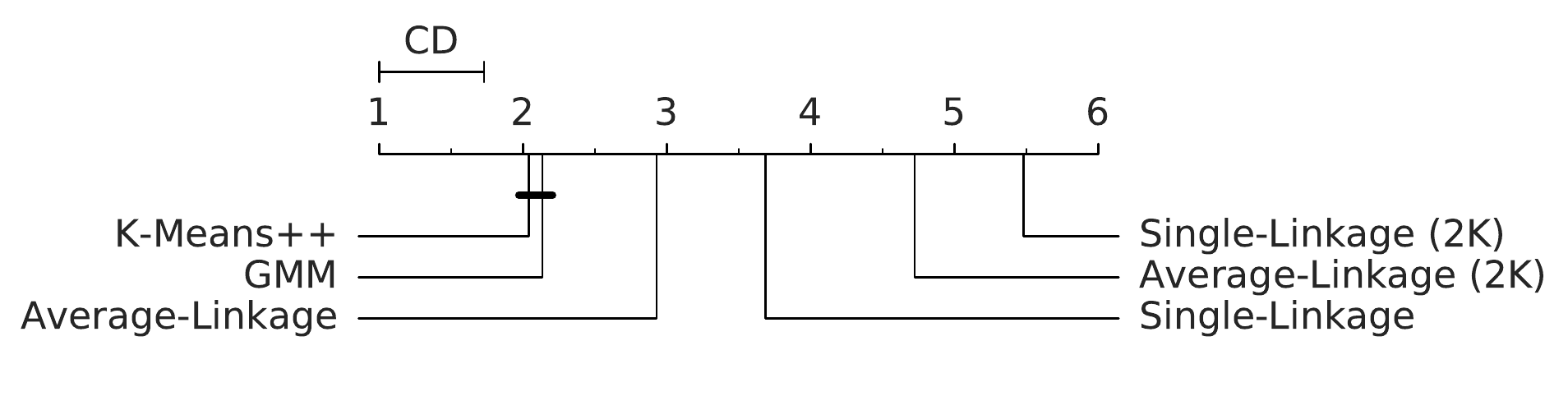}}\bigskip
	
	\subfloat[UCI\label{fig:cd_uci}]{\includegraphics[width=0.46\textwidth]{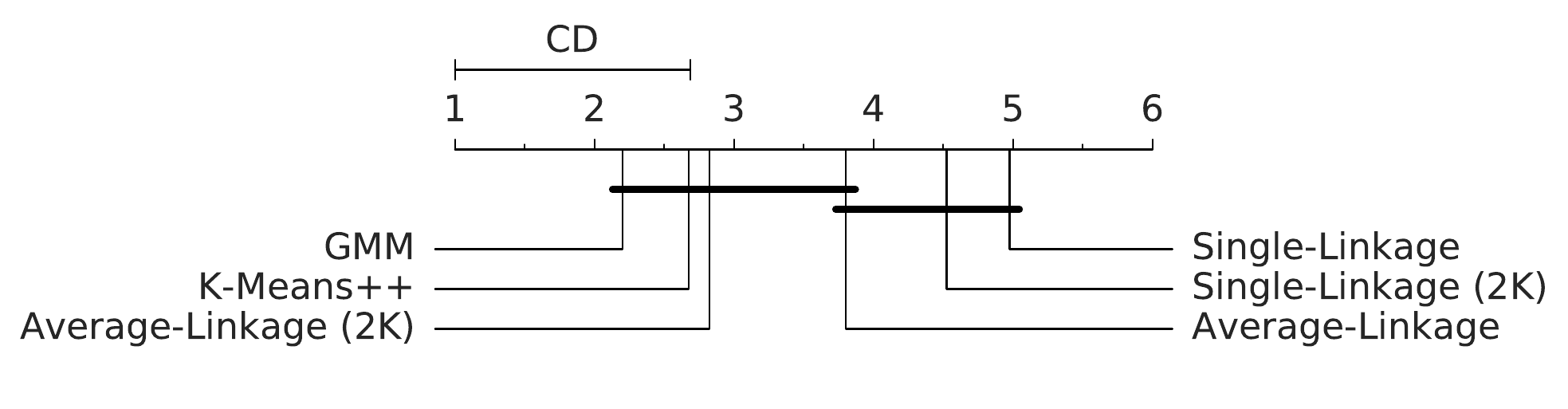}}\hfill
	\subfloat[UKC\label{fig:cd_ukc}]{\includegraphics[width=0.46\textwidth]{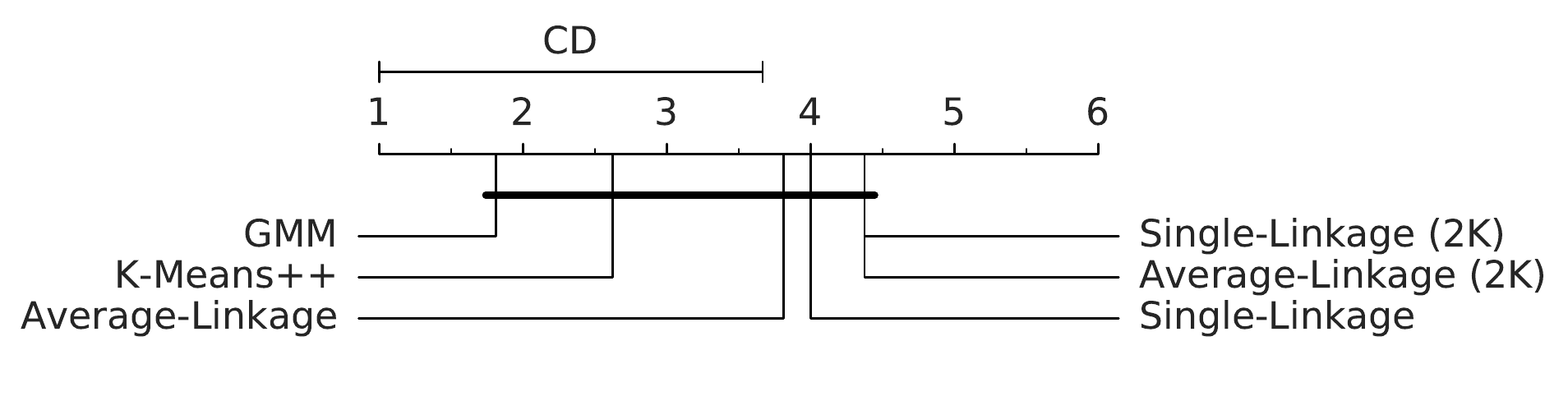}}\bigskip
	\caption{Critical difference (CD) diagrams, showing the mean rank (in terms of ARI) for the remaining dataset collections, for each algorithm. Algorithms connected by solid lines are not significantly different according to a two-tailed Nemenyi test.}
	\label{fig:cd_extras}
  \end{figure*}
  
For the \emph{QJ} datasets (Fig.~\ref{fig:cd_qj}), the compactness-based algorithms (GMM and \kmeanplus) are clearly the best performing algorithms, with GMM nearly superior across all datasets, highlighting a lack of diversity. Similarly, for the SIPU datasets (Fig.~\ref{fig:cd_sipu}) GMM and \kmeanplus~are more equally the best performing, where the lower connectivity (i.e.\ lack of overlap between clusters) compared to \emph{QJ} datasets does not distinguish performance further between these two algorithms. The UCI datasets (Fig.~\ref{fig:cd_uci}) show little significant difference between the clustering algorithms which, when combined with the low performance shown in Fig.~\zref{main-fig:suite_bplot}, indicates that there is a high variance of performance, but that this diversity is due to low cluster structure (rather than varied cluster structures). Finally, the UKC datasets (Fig.~\ref{fig:cd_ukc}) are too few in number to identify significant differences, though the the higher rank of the compactness-based algorithms fits with visualization of the clusters \cite{Garza-Fabre2017}.

\section{Generating datasets that challenge specific algorithms (Versus mode)} \label{sec:versus}
Section~\zref{main-ssec:versus} presented results for the reformulation of the fitness function to evolve datasets that directly maximize the performance difference between two algorithms. Here, we show how algorithm initialization affects the results, and provide further visual examples of the different head-to-head scenarios to illustrate the properties which HAWKS found differentiated the algorithms (in particular for the head-to-head between average- and single-linkage).

\subsection{Sensitivity to algorithm initialization}
With commentary in the main text, Fig.~S-\ref{fig:versus_grid_2} shows the grid plot of each head-to-head with the stochastic algorithms (GMM and \kmeanplus) run again using a different seed to ascertain to what extent the performance is a result of poor initialization.

\begin{figure}[th]
	\centering
	\includegraphics[width=\columnwidth]{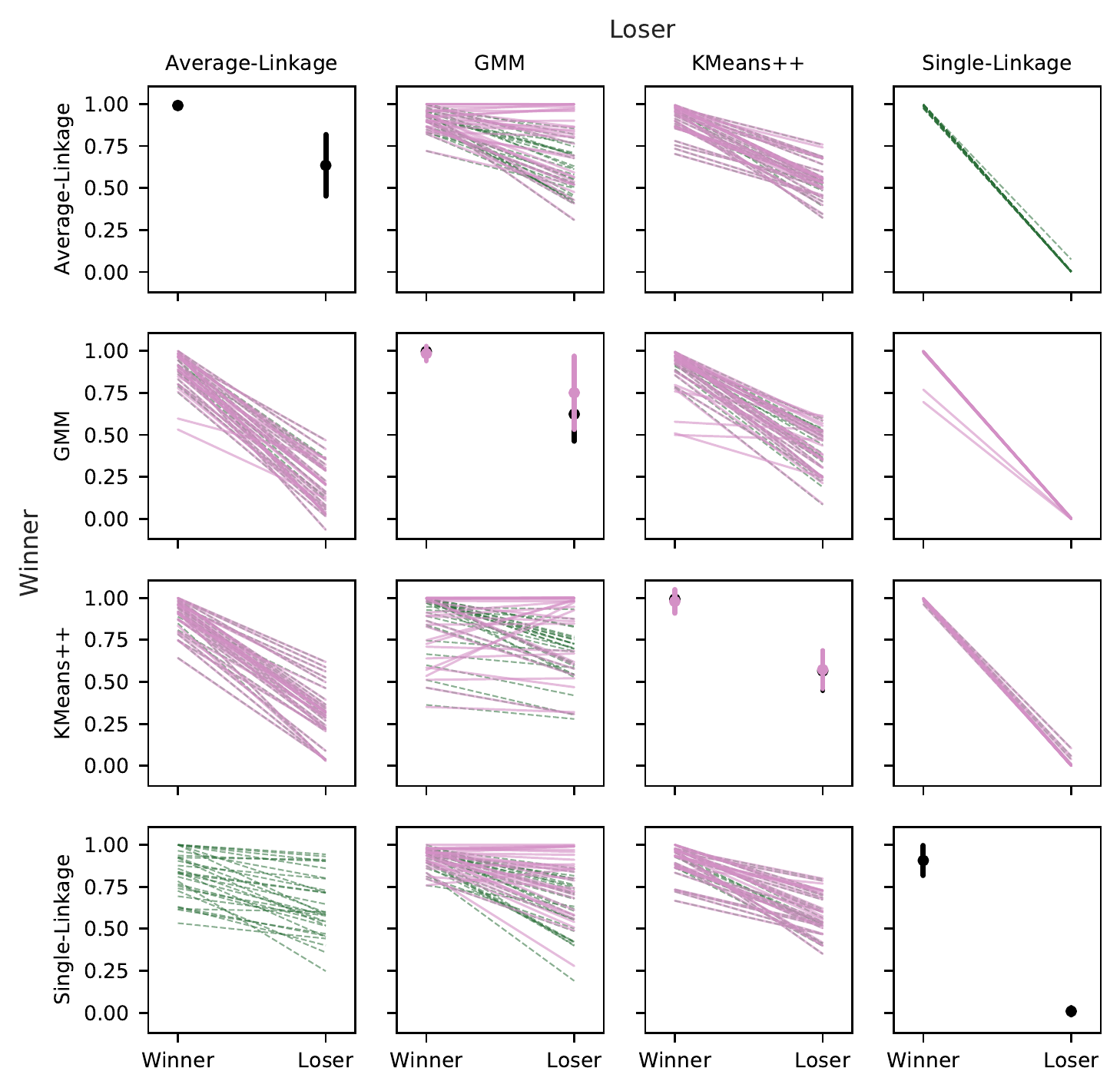}
	\caption{Performance (ARI) for each algorithm as the winner (row) and loser (column). Each line represents the best individual from a single run, connecting the ARI obtained for \winneralg~and \loseralg. On the diagonal, the average (and standard deviation) performance is aggregated for that individual as the winner and loser, indicating the overall capability of HAWKS to produce datasets that are simple and difficult for that algorithm. The stochastic algorithms (GMM and \kmeanplus) are re-run with a different initialization (as indicated by the solid lines, with the original results as dashed lines).}
	\label{fig:versus_grid_2}
\end{figure}

\subsection{Additional scenarios}

\begin{figure*}
	\centering
	\subfloat[\algversus{Single-linkage}{average-linkage}\label{fig:versus_slink-alink}]{\includegraphics[width=0.75\textwidth]{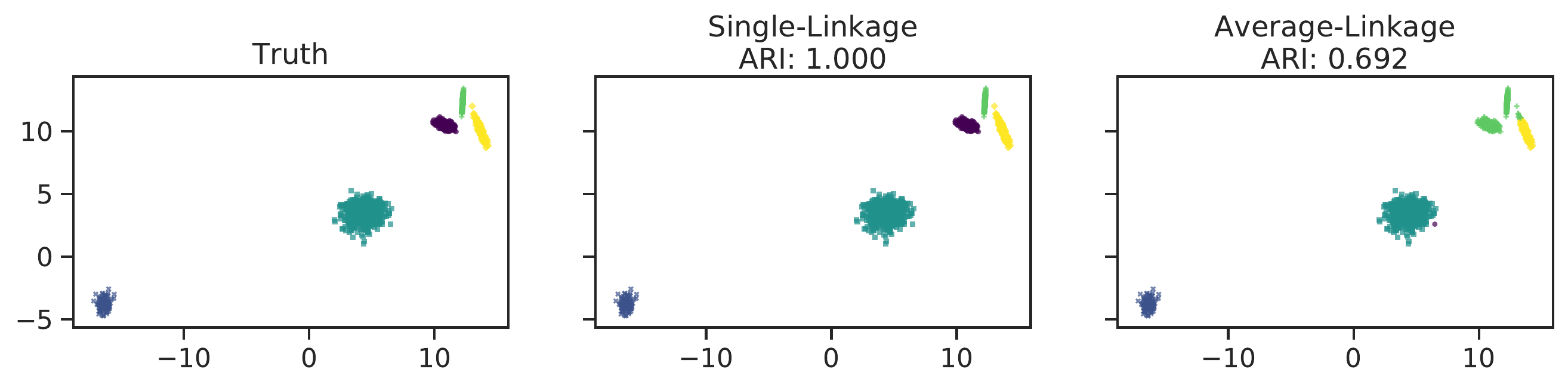}}\hfill
	\subfloat[\algversus{Average-linkage}{single-linkage}\label{fig:versus_alink-slink}]{\includegraphics[width=0.75\textwidth]{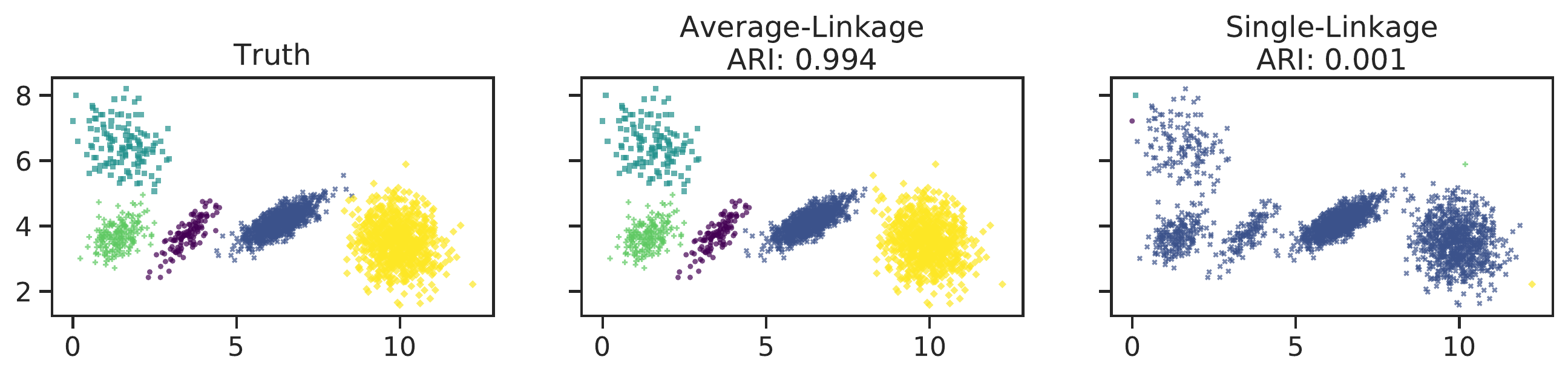}}\hfill
	\subfloat[\algversus{GMM}{\kmeanplus}\label{fig:versus_gmm-kmeans}]{\includegraphics[width=0.75\textwidth]{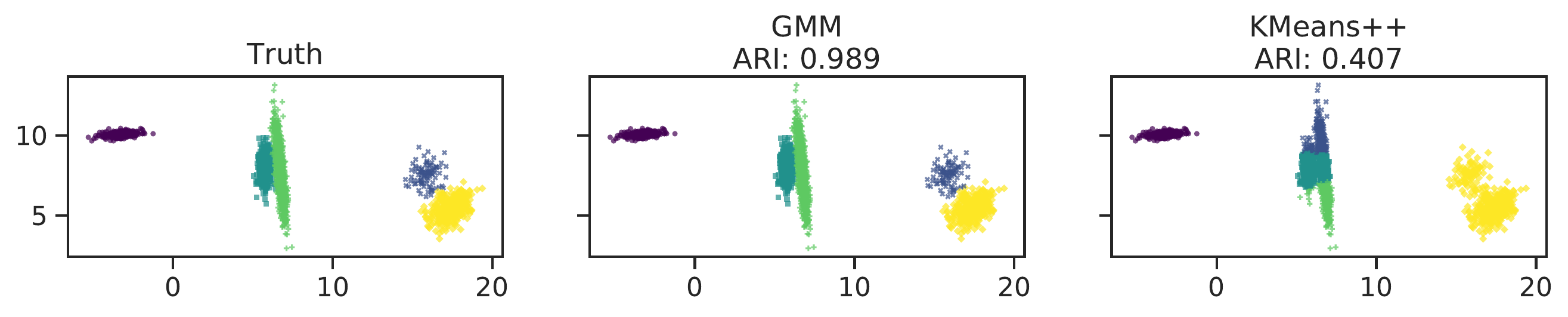}}
	\caption{Examples of datasets for the listed head-to-heads. Each figure shows the ground truth, and the cluster assignment for each algorithm with the associated ARI.}
	\label{fig:versus_extra}
\end{figure*}

\subsubsection{\algversus{Single-linkage}{Average-linkage}}
Fig.~\zref{main-fig:versus_grid} showed that there was a high spread in the ARI for both algorithms, and high variance in the maximum difference found between runs, highlighting the difficulty for this scenario. There are some datasets that correlate both high and low performance between the two, indicating that the algorithms do share some similarities. Fig.~S-\ref{fig:versus_slink-alink} shows an example, where the structure is common among the datasets that did find a performance difference. The clusters are in general well-separated, as by placing some clusters much further away and co-locating a few smaller clusters, the averaging criterion (to determine where to split next) used by average-linkage assigns clusters that are closer together the same cluster label. The mixed eccentricities of clusters that HAWKS is able to generate helps facilitate the discovery of this exploit. As the clusters need to be sufficiently far away before a chance in the fitness can be found, there may be insufficient pressure for this exploit to be reliably found, highlighting the high variation in performance for this scenario.

\subsubsection{\algversus{Average-linkage}{Single-linkage}}
As average-linkage uses the average of the distances between groups, rather than the smallest distance, it is not susceptible to the chaining effect~\cite{hubert1974approximate} that single-linkage is. As a result, Fig.~S-\ref{fig:versus_alink-slink} shows that the exploit used in this head-to-head is simply to place clusters close together in order for single-linkage to determine the majority of the points are in a single cluster.

\bibliographystyle{IEEEtran}
\bibliography{IEEEabrv,refs}

%% file: macros.tex
\DeclareDocumentCommand{\datapoint}{ o o m }{%
    \ensuremath{\mathbf{#3}
    \IfValueT{#2}{_{#2}}
    \IfValueT{#1}{^{#1}}}%
}

\DeclareDocumentCommand{\cluster}{ O{C} m }{%
    \ensuremath{#1_{#2}}%
}
\DeclareDocumentCommand{\clustersize}{ O{C} m }{%
    \ensuremath{|#1_{#2}|}%
}

\DeclareDocumentCommand{\hawksmean}{ o o }{%
    \ensuremath{\bm{\mu}
    \IfValueT{#2}{_{#2}}
    \IfValueT{#1}{^{#1}}}%
}

\newcommand{\hawksmeanarg}[1]{\hawksmean[(#1)]}
\newcommand{\hawksmeanargnew}[1]{\hawksmean[\prime(#1)]}
\newcommand{\hawksmeanglobal}{\ensuremath{\bar{\hawksmean}}}

\DeclareDocumentCommand{\hawkscov}{ o o }{%
    \ensuremath{\bm{\Sigma}
    \IfValueT{#2}{_{#2}}
    \IfValueT{#1}{^{#1}}}%
}

\DeclareDocumentCommand{\hawkscovaxisarg}{ o o }{%
    \ensuremath{\widetilde{\hawkscov}
    \IfValueT{#2}{_{#2}}
    \IfValueT{#1}{^{#1}}}%
}

\newcommand{\hawkscovaxis}{\ensuremath{\hawkscovaxisarg}}
\newcommand{\hawksrotation}{\ensuremath{\bm{R}}}
\newcommand{\hawksrotationarg}[1]{\ensuremath{\hawksrotation^{(#1)}}}
\newcommand{\hawksscaling}{\ensuremath{\bm{S}}}
\newcommand{\hawksscalingarg}[1]{\ensuremath{\hawksscaling^{(#1)}}}

\newcommand{\overlap}{\ensuremath{\mathit{overlap}}}
\newcommand{\silhall}{\ensuremath{s_{\mathit{all}}}}
\newcommand{\silhtarget}{\ensuremath{s_{t}}}

\newcommand{\winneralg}{\ensuremath{\mathcal{A}_{w}}}
\newcommand{\loseralg}{\ensuremath{\mathcal{A}_{l}}}
\newcommand{\kmeanplus}{K-Means\texttt{++}}
\newcommand{\algversus}[2]{#1 vs.\ #2}

\newcommand{\stochranking}{\ensuremath{P_{f}}}
\newcommand{\eigratio}{\ensuremath{\lambda^{\mathit{ratio}}}}

\DeclareMathOperator{\diag}{diag}

\DeclareDocumentCommand{\neighbour}{ o O{x} o o m}{%
    \ensuremath{n
    \IfValueT{#2}{_{\datapoint[#3][#4]{#2}}}
    \IfValueT{#1}{^{#1}}%
    (#5)}%
}

%% file: tables/table1.tex
\begin{table}[t]
    \caption{Number of times (as a percentage of the total number of datasets for each source) each algorithm achieved the highest ARI for a given dataset. The best performing algorithm on each data source is highlighted in bold.}
    \label{tab:results_best-alg}
	% \centering
	% \addtolength{\tabcolsep}{0.1cm} 
    \begin{tabularx}{\columnwidth}{@{} p{0.85cm} *{7}{>{\centering\arraybackslash}X} @{}}
		\toprule
		& Average- & Average- & & K- & Single- & Single- & \\[-0.3cm]
		Source & Linkage & Linkage & GMM & Means & Linkage & Linkage & Tied\\
		& & ($2K$) & & \texttt{++} & &($2K$) & \\
		\midrule 
		% HAWKS & 117 & 25 & 107 & 25 & 7 & 5 & 162 \\
		HAWKS & 0.261 & 0.056 & 0.239 & 0.056 & 0.016 & 0.011 & \textbf{0.362}\\
		% \emph{HK} & -- & 149 & 199 & 2 & -- & -- & -- \\
		\emph{HK}~\cite{handl2005generator} & -- & 0.426 & \textbf{0.569} & 0.006 & -- & -- & --\\
		% \emph{QJ} & 2 & 5 & 157 & 20 & -- & 1 & 58 \\
		\emph{QJ}~\cite{qiu2006generation} & 0.008 & 0.021 & \textbf{0.646} & 0.082 & -- & 0.004 & 0.239\\
		% \emph{SIPU} & -- & -- & 11 & 21 & -- & -- & 75\\
		\emph{SIPU}~\cite{franti2018k} & -- & -- & 0.106 & 0.202 & -- & -- & \textbf{0.692}\\
		% \emph{UCI} & 2 & 3 & 10 & 4 & -- & -- & 1\\
		\emph{UCI}~\cite{Dua2017} & 0.100 & 0.150 & \textbf{0.500} & 0.200 & -- & -- & 0.050\\
		% \emph{UKC} & -- & -- & 5 & 1 & -- & -- & 2\\
		\emph{UKC}~\cite{Garza-Fabre2017} & -- & -- & \textbf{0.625} & 0.124 & -- & -- & 0.250\\
		\bottomrule
	\end{tabularx}
\end{table}

%% file: supp_tables/supp-table1.tex
\begin{table}[tbp]
\begin{threeparttable}[tbp]
    \renewcommand{\arraystretch}{1.2}
    \begin{center}
        \caption{The number of data points ($N$), clusters ($K$), dimensions ($D$), and datasets for each collection of datasets.}
        \label{tab:dataset_params}
        % \begin{tabular}{@{} llp{2.5cm}p{3cm}l @{}}
        \begin{tabularx}{\columnwidth}{@{} p{0.75cm} *{3}{>{\centering\arraybackslash}X} c @{}}
            \toprule
            Source & $N$\tnote{1} & $K$ & $D$ & \# Datasets\\
            \midrule
            HAWKS & 500, 2500 & 5, 30 & 2, 50 & 448\\
            \emph{HK}~\cite{handl2005generator} & $5474 \pm 2137$ & 10, 20, 40, 60, 80, 100, 120 & 20, 50, 100, 150, 200 & 350\\
            \emph{QJ}~\cite{qiu2006generation} & $599 \pm 289$ & 3, 6, 9 & 5--24 & 243\\
            \emph{SIPU}~\cite{franti2018k} & $2248 \pm 817$ & 2, 15, 20, 35, 50 & $2^{1},2^{2},\ldots,2^{10}$ & 107\\
            \emph{UCI}~\cite{Dua2017} & $600 \pm 571$ & 2, 3, 4, 6, 7, 8, 10, 11, 15 & 3, 4, 6--11, 13, 18, 19, 22, 30, 34, 44, 60, 90, 166 & 20\\
            \emph{UKC}~\cite{Garza-Fabre2017} & $30685 \pm 2178$ & 10, 11, 12 & 2 & 8\\
            \bottomrule
        \end{tabularx}
        \begin{tablenotes}
            \item [1] The average (and standard deviation) are given, except for HAWKS for which the values are targetted.
        \end{tablenotes}
    \end{center}
\end{threeparttable}
\end{table}

%% file: supp_tables/supp-table2.tex
\begin{table*}[tbp]
    \begin{center}
    \addtolength{\tabcolsep}{-2pt}
    \begin{threeparttable}[tbp]
        \renewcommand{\arraystretch}{1.15}
            \caption{The mean and standard deviation (SD) of the problem feature values for every collection of datasets, including both the Index and Versus modes of HAWKS.}
            \label{tab:problem_features}
            \begin{tabular}{@{} 
                p{2.1cm}
                *{2}{S[table-format=1.3]}
                *{2}{S[table-format=3.3]}
                S[table-format=3.3]S[table-format=2.3]
                *{2}{S[table-format=1.3]}
                *{2}{S[table-format=2.3]}
                *{2}{S[table-format=1.3]}
                *{2}{S[table-format=1.3]}
                @{}}
                \toprule
                & & & & & & & & & & & & & \multicolumn{2}{c}{Silh.~width}\\
                & & & \multicolumn{2}{c}{Number of} & \multicolumn{2}{c}{Average} & \multicolumn{2}{c}{Cluster Size} & \multicolumn{2}{c}{Number of} & \multicolumn{2}{c}{Silh.~width} & \multicolumn{2}{c}{(standard}\\
                & \multicolumn{2}{c}{Connectivity} & \multicolumn{2}{c}{Dimensions} & \multicolumn{2}{c}{Eccentricity} & \multicolumn{2}{c}{Entropy} & \multicolumn{2}{c}{Clusters} & \multicolumn{2}{c}{(average)} & \multicolumn{2}{c}{deviation)}\\
                \cmidrule(lr){2-3} \cmidrule(lr){4-5} \cmidrule(lr){6-7} \cmidrule(lr){8-9} \cmidrule(lr){10-11} \cmidrule(lr){12-13} \cmidrule(lr){14-15}
                \multicolumn{1}{c}{Collection} & {Mean} & {SD} & {Mean} & {SD} & {Mean} & {SD} & {Mean} & {SD} & {Mean} & {SD} & {Mean} & {SD} & {Mean} & {SD}\\
                \midrule
                HAWKS (\emph{Index})\tnote{1} & 0.114459 & 0.198263 & 26.000000 & 24.026831 & 5.692003 & 2.568014 & 0.884344 & 0.125768 & 17.500000 & 12.513974 & 0.675197 & 0.224665 & 0.226005 & 0.165707\\
                HAWKS (\emph{Versus})\tnote{2} & 0.144148 & 0.174804 & 2.000000 & 0.000000 & 2.550926 & 1.308774 & 0.825312 & 0.074083 & 5.000000 & 0.000000 & 0.711612 & 0.187714 & 0.299956 & 0.137632\\
                \emph{HK}~\cite{handl2005generator}& 0.008082 & 0.007985 & 104.000000 & 65.392796 & 145.202307 & 79.847454 & 0.983670 & 0.021138 & 61.428571 & 38.012429 & 0.474222 & 0.043350 & 0.436918 & 0.022631 \\
                \emph{QJ}~\cite{qiu2006generation}& 0.206903 & 0.247219 & 13.000000 & 6.926414 & 3.615214 & 0.866847 & 0.965749 & 0.023316 & 6.000000 & 2.454545 & 0.380339 & 0.153598 & 0.145065 & 0.035949 \\
                \emph{SIPU}~\cite{franti2018k}& 0.091199 & 0.208818 & 191.345794 & 308.119895 & 3.410572 & 3.007865 & 0.999985 & 0.000072 & 3.411215 & 6.279576 & 0.634213 & 0.225229 & 0.090581 & 0.113241 \\
                \emph{UCI}~\cite{Dua2017}& 0.789089 & 0.409431 & 28.350000 & 39.125002 & 20.715856 & 20.705831 & 0.897355 & 0.115664 & 4.900000 & 3.712000 & 0.083964 & 0.279152 & 0.375244 & 0.150354 \\
                \emph{UKC}~\cite{Garza-Fabre2017}& 0.000539 & 0.000791 & 2.000000 & 0.000000 & 1.938094 & 0.177446 & 0.996914 & 0.002256 & 10.875000 & 0.640870 & 0.780285 & 0.061844 & 0.231887 & 0.057724 \\
                \bottomrule
            \end{tabular}
            \begin{tablenotes}
                \item [1] The 448 datasets generated in Section~\zref{main-ssec:benchmark}.
                \item [2] The 360 datasets generated in Section~\zref{main-ssec:versus}.
            \end{tablenotes}
        \end{threeparttable}
    \addtolength{\tabcolsep}{2pt}
    \end{center}
\end{table*}